\documentclass{article} % For LaTeX2e
\usepackage{iclr2026_conference,times}

\usepackage{amsmath,amsfonts,bm}

\def\eqref#1{equation~\ref{#1}}
\def\1{\bm{1}}

\DeclareMathAlphabet{\mathsfit}{\encodingdefault}{\sfdefault}{m}{sl}
\SetMathAlphabet{\mathsfit}{bold}{\encodingdefault}{\sfdefault}{bx}{n}

\usepackage{hyperref}
\usepackage{url}
\usepackage{minted}
\usepackage{booktabs}
\usepackage{graphicx}
\usepackage{colortbl}
\usepackage{natbib}
\usepackage{xspace}
\usepackage[most]{tcolorbox}
\usepackage[dvipsnames]{xcolor}
\usepackage{float}
\usepackage{fancyvrb}
\usepackage{fvextra}
\usepackage[T1]{fontenc}
\usepackage{caption}
\usepackage{enumitem}
\usepackage{subcaption}
\usepackage{booktabs,multirow}
\usepackage{siunitx}
\usepackage{booktabs,tabularx}
\usepackage{amsmath,amssymb}
\usepackage[ruled,vlined,noend]{algorithm2e}

\usepackage{makecell}
\usepackage{siunitx}
\usepackage{listings}
\usepackage{textcomp}
\usepackage{xcolor}
\usepackage[scaled=0.95]{inconsolata}
\newcommand{\ours}[0]{\textsc{Agent2World}\xspace}
\newcommand{\single}[0]{\textsc{Agent2World}$_{\textit{\text{Single}}}$\xspace}
\newcommand{\multi}[0]{\textsc{Agent2World}$_{\textit{\text{Multi}}}$\xspace}

\newcommand{\fonepred}[0]{\textsc{F1\textsubscript{pred}}\xspace}
\newcommand{\foneparam}[0]{\textsc{F1\textsubscript{param}}\xspace}
\newcommand{\foneprecond}[0]{\textsc{F1\textsubscript{precond}}\xspace}
\newcommand{\foneeff}[0]{\textsc{F1\textsubscript{eff}}\xspace}

\newcommand{\ourgray}[0]{\rowcolor{gray!15}}
\newcommand{\ourblue}[0]{\rowcolor{blue!10}}

\newif\ifshowrevisions
\newenvironment{revblock}
  {\par\begingroup
   \ifshowrevisions \color{red}\fi
  }
  {\par\endgroup}

\newcommand{\revinline}[1]{{\ifshowrevisions \color{red} #1 \else #1 \fi}}

\title{\ours: Learning to Generate Symbolic World Models via Adaptive Multi-Agent Feedback}

\author{
        \textbf{Mengkang Hu}$^{\spadesuit\heartsuit}$\thanks{
        Equal contribution. Corresponding to mkhu@connect.hku.hk, pluo.lhi@gmail.com.} \ \ 
        \textbf{Bowei Xia}$^{\spadesuit\diamondsuit}$ $^*$ \ \ 
        \textbf{Yuran Wu}$^\spadesuit$ \ \ 
        \textbf{Ailing Yu}$^\spadesuit$\ \ 
        \textbf{Yude Zou}$^{\spadesuit}$\ \
        \textbf{Qiguang Chen}$^{\clubsuit}$ \\
        ~\textbf{Shijian Wang}$^\heartsuit$\ \
        \textbf{Jiarui Jin}$^{\heartsuit}$\ \
        \textbf{Kexin Li}$^{\diamondsuit}$\ \
        \textbf{Wenxiang Jiao}$^{\heartsuit}$\ \
        \textbf{Yuan Lu}$^{\heartsuit}$\ \
        \textbf{Ping Luo}$^\spadesuit~^\dagger$ 
       \\ 
        ~$^\spadesuit$ The University of Hong Kong 
        \quad
        $^\heartsuit$ Xiaohongshu Inc. 
        \\
        ~$^\diamondsuit$ UESTC
        \quad
        $^\clubsuit$ Harbin Institute of Technology
}

\iclrfinalcopy % Uncomment for camera-ready version, but NOT for submission.
\usepackage{float}
\begin{document}

\maketitle

\begin{abstract}
Symbolic world models (e.g., PDDL domains or executable simulators) are central to model-based planning, but training LLMs to generate such world models is limited by the lack of large-scale verifiable supervision.
Current approaches rely primarily on static validation methods that fail to catch behavior-level errors arising from interactive execution.
In this paper, we propose \ours, a tool-augmented multi-agent framework that achieves strong inference-time world-model generation and also serves as a data engine for supervised fine-tuning, by grounding generation in multi-agent feedback.
\ours follows a three-stage pipeline: 
(\textit{i}) A \emph{Deep Researcher} agent performs knowledge synthesis by web searching to address specification gaps;
(\textit{ii}) A \emph{Model Developer} agent implements executable world models; 
And (\textit{iii}) a specialized \emph{Testing Team} conducts adaptive unit testing and simulation-based validation. 
\ours demonstrates superior inference-time performance across three benchmarks spanning both Planning Domain Definition Language(PDDL) and executable code representations, achieving consistent state-of-the-art results. 
Beyond inference, \emph{Testing Team} serves as an interactive environment for the Model Developer, providing behavior-aware adaptive feedback that yields multi-turn training trajectories.
The model fine-tuned on these trajectories substantially improves world-model generation, yielding an average relative gain of 30.95\% over the same model before training.
Project page: \href{https://agent2world.github.io}{agent2world.github.io}.
\end{abstract}

\vspace{-10pt}
\section{Introduction}
\label{sec:intro}

% backgroundyu
% The ability to model the world is widely regarded as a prerequisite for artificial superintelligence (ASI)~\citep{lecun2022path,craik1967nature}. 
In recent years, researchers have explored \emph{symbolic world models}, a formal representation of an environment’s dynamics and constraints, which is widely used in model-based planning~\citep{guan2023leveraging,lecun2022path,craik1967nature}.
The task of \emph{symbolic world-model generation} involves automatically synthesizing these models from natural language descriptions, eliminating the need for domain experts to manually design and specify complex rules and dynamics.
Large language models (LLMs)~\citep{guo2025deepseek,zhao2023survey,bai2023qwen} have made this automation possible by combining two key capabilities: commonsense knowledge about how the world works, and code generation abilities that formalize this knowledge into executable representations~\citep{chen2025towards}.
However, learning to generate such models from natural language remains difficult: correctness is behavioral and execution-dependent, while large-scale, verifiable supervision is scarce.

\begin{figure}[t]
    \centering
    \includegraphics[width=0.99\linewidth]{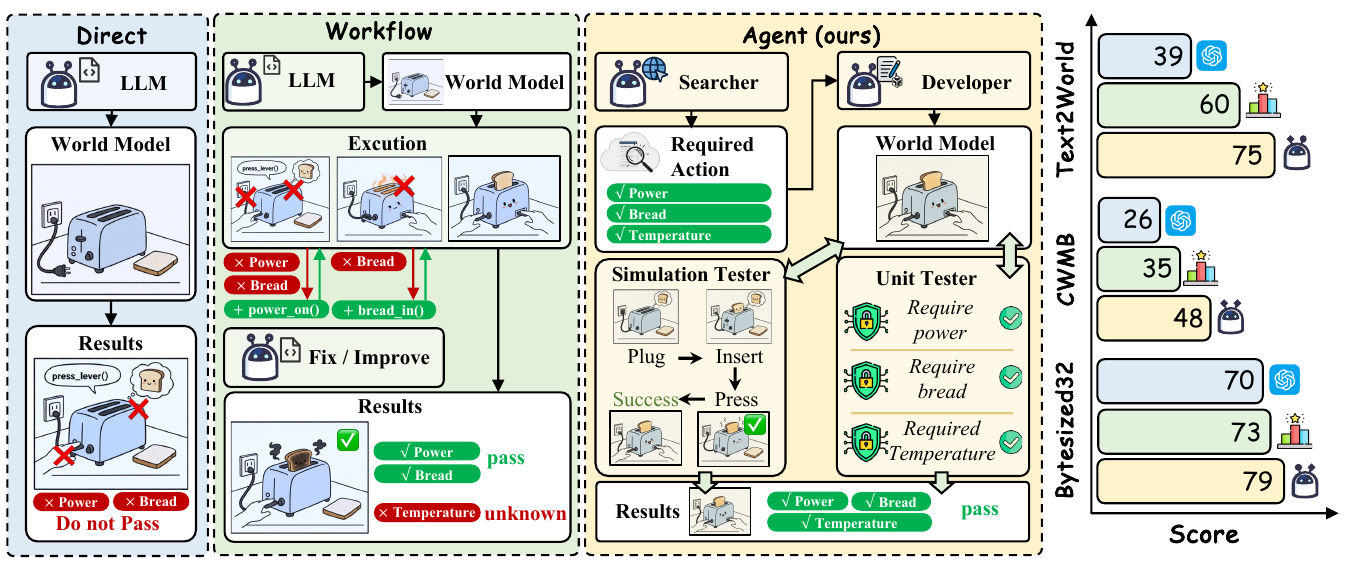}
    \caption{Comparison of \ours and previous world-model generation paradigms.}
    \label{fig:teaser}
    \vspace{-15pt}
\end{figure}

% Prior work landscape
As illustrated in Figure~\ref{fig:teaser}, prior work in this domain largely follows two paradigms: (i) \emph{direct generation} of symbolic world models, and (ii) \emph{scripted workflows} that couple generation with iterative verification and repair. 
Across both PDDL-style domains~\citep{guan2023leveraging,hu2025text2world} and executable code world models~\citep{dainese2024cwmb}, the second paradigm typically couples generation with a pre-specified verification interface (e.g., parsers/planners/validators, fixed sets of evaluation trajectories).
While such static validation improves syntactic validity, it misses behavior-level errors that only appear under interactive execution (e.g., inconsistent state updates or unreachable goals).
Furthermore, existing studies on generating symbolic world models with LLMs have primarily focused on training-free methods for one particular type of world models~\citep{yu2025generating,kong20253d,zhang2025autoenv}, rather than fundamentally enhancing the world modeling capabilities of the LLMs themselves.

% We propose \ours, a new paradigm for symbolic world-model generation that leverages tool-augmented, autonomous LLM-based agents, which plan and call tools adaptively.
% The key advantages lie in:
% (i) \emph{Proactive and adaptive execution.} Rather than passively fixing errors through rigid sequences, \ours proactively gathers information and dynamically adjusts strategies based on intermediate feedback, automatically deciding when to terminate and which tools to use for maximum efficiency.
% (ii) \emph{Scalable external knowledge integration.} Unlike knowledge-isolated approaches or labor-intensive human-in-the-loop methods, \ours incorporate web search as first-class tools to automatically fill specification gaps and enforce commonsense regularities, minimizing LLM hallucination~\citep{huang2025survey}.
% (iii) \emph{Unified cross-representation framework.} While prior systems suffer from representation fragmentation, \ours seamlessly handles both PDDL-based and code-based models through lightweight tool adapters, enabling the same agentic framework to work across different symbolic representations.

In this paper, we propose \ours, a tool-augmented multi-agent framework that evaluates and improves world models through interactive execution.
Given a natural-language description, \ours coordinates multiple LLM-based agents with access to external tools (e.g., web retrieval and code execution) to iteratively produce an executable world model.
At a high level, \ours consists of three stages (Figure~\ref{fig:method}): a \emph{Deep Researcher} resolves underspecified details by gathering missing background knowledge, a \emph{Model Developer} implements the world model in the target representation, and a \emph{Testing Team} evaluates the resulting artifact under adaptive execution (unit tests and simulation-style evaluation) and returns structured feedback for repair.
% This execution-grounded feedback targets behavior-level failures that fixed validators or predetermined test suites can miss.
Unlike prior pipelines that rely on static validators or fixed test suites, \ours conditions evaluation on observed execution behavior and iteratively produces targeted checks, which expose behavior-level failures that predetermined checks often miss.

Most importantly, the same developer-tester interaction loop can be viewed as an interactive environment for the \emph{Model Developer}, naturally producing multi-turn trajectories that capture how world models are revised under feedback and providing verifiable reward.
We use this property to turn inference-time feedback into training data for improving the Model Developer policy via verifier-guided rejection sampling (Section~\ref{sec:training}) and construct a dataset consisting of 1526 high-quality verified trajectories covering four distinct types of world models across diverse domains.

We conducted experiments on three benchmark datasets to evaluate the performance of \ours: (i) Text2World~\citep{hu2025text2world} for PDDL-based domain generation, (ii) Code-based World Models Benchmark (CWMB)~\citep{dainese2024cwmb} for MuJoCo-style environment generation, and (iii) ByteSized32~\citep{bytesized32} for reasoning-heavy text game generation.
First, we validated the inference-time performance of \ours with two different models: GPT-4.1-mini~\citep{gpt4} and Llama-3.1-8b~\citep{llama3}. Our results demonstrate that \ours consistently achieves state-of-the-art performance across all three benchmarks with both models. This highlights its robust capabilities in symbolic world model generation, regardless of the underlying model.
Furthermore, we also validated the effectiveness of \ours through training experiments. We performed supervised fine-tuning on the same Llama-3.1-8b model. Our results show consistent improvements across all benchmarks, indicating that the model is able to refine its world-model generation process effectively through iterative multi-agent feedback. 

\section{Problem Definition}

% \subsection{Problem Definition}
% We investigate the problem of \emph{symbolic world-model generation} from natural language. 
% Given a textual description $x$, the objective is to synthesize an \emph{executable} program $\mathcal{M}_F$ that faithfully captures the dynamics and constraints of the environment. 
% Such a program may take various forms, for instance, a specification in the Planning Domain Definition Language (PDDL)~\citep{hu2025text2world,aeronautiques1998pddl} or an implementation in Python~\citep{dainese2024cwmb,bytesized32}.
% Formally, an environment is defined by a set of predicates $\mathcal{P}$, a set of actions $\mathcal{A}$, and a transition function $T: S \times \mathcal{A} \rightarrow S$, where $S$ denotes the set of possible states. Semantically, $\mathcal{M}_F$ encodes these components to represent the environment in an executable manner.
% We therefore define the task as a mapping $P(x) \;=\; \mathcal{M}_F$, where $\mathcal{M}_F = \langle \mathcal{P}, \mathcal{A}, T \rangle$,
% and $P$ is a synthesis procedure that generates the world-model program from natural language input $x$.

We investigate the problem of \emph{symbolic world-model generation} from natural language. 
Given a textual description $x$, the objective is to synthesize an \emph{executable} program $\mathrm{WM}$ that faithfully captures the dynamics and constraints of the environment. Such a program may take various forms, for instance, a specification in the Planning Domain Definition Language (PDDL)~\citep{hu2025text2world,aeronautiques1998pddl} or an implementation in Python~\citep{dainese2024cwmb,bytesized32}.
Formally, an environment is defined by a set of predicates $\mathcal{P}_{\mathrm{env}}$, a set of actions $\mathcal{A}_{\mathrm{env}}$, and a transition function $T_{\mathrm{env}}: S_{\mathrm{env}} \times \mathcal{A}_{\mathrm{env}} \rightarrow S_{\mathrm{env}}$, where $S_{\mathrm{env}}$ denotes the set of possible states. Semantically, $\mathrm{WM}$ encodes these components to represent the environment in an executable manner.
We therefore define the task as a mapping $F(x) \;=\; \mathrm{WM} \quad \text{where } \mathrm{WM} = \langle \mathcal{P}_{\mathrm{env}}, \mathcal{A}_{\mathrm{env}}, T_{\mathrm{env}} \rangle,$
and $F$ is a synthesis procedure that generates the world-model program from natural language input $x$.

% \subsection{Autonomous Agent}
% \label{sec:autoagent}

% We consider an \emph{autonomous agent} that \emph{interleaves reasoning with tool use}~\citep{qin2023toolllm,yao2022react}. Concretely, the agent firstly produces reasoning traces and issues tool calls in an alternating loop.
% Let $\mathcal{T}=\{t_1,\dots,t_m\}$ denote a set of callable tools. At discrete time $k$, the agent maintains a history
% $h_k = (x, o_{\le k}, a_{<k}, r_{<k})$ consisting of the task description $x$, observations $o$, past actions $a$, and internal reasoning traces $r$.
% A \emph{policy} $\pi_\theta$ maps histories to the next reasoning trace and action: $(r_k, a_k) \sim \pi_\theta(\cdot \mid h_k)$, where $a_k \in \mathcal{T}$.
% In practice, $\pi_\theta$ is implemented by an LLM.

\section{Methodology}

\subsection{Architectural Design}
\label{sec:arch}

\begin{figure}[tb]
    \centering
    \includegraphics[width=0.99\linewidth]{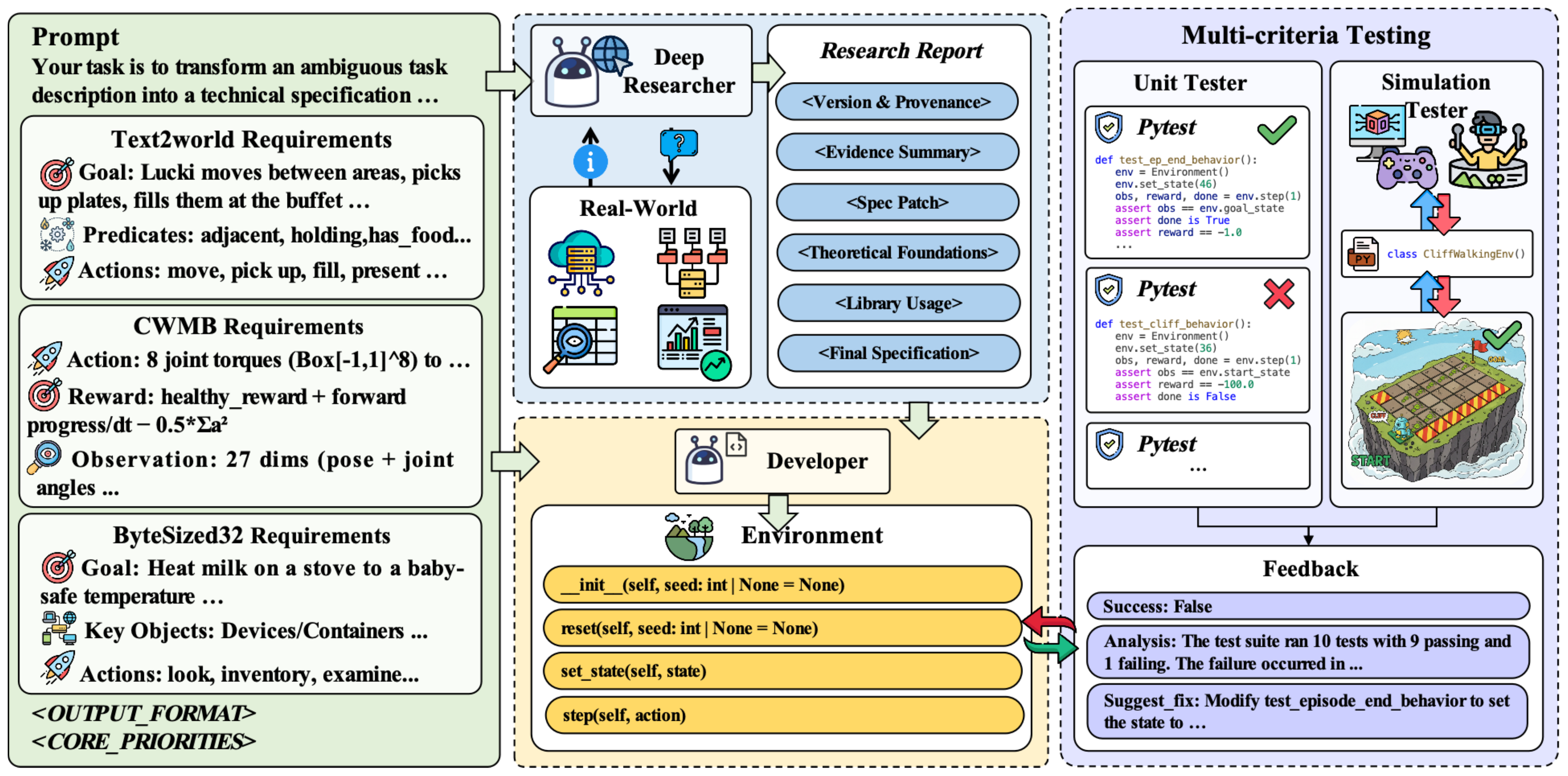}
    \caption{Overall Pipeline of \ours.}
    \label{fig:method}
\end{figure}

As is shown in Figure~\ref{fig:method}, \multi unfolds in three stages:
\textit{(i)} \textbf{Knowledge Synthesis} (\S\ref{sec:ks}): As outlined in Section~\ref{sec:intro}, a key challenge in symbolic world-model construction arises from incomplete descriptions. For example, commonsense knowledge may be missing both from LLMs and from the given specifications. To address this limitation, we employ a \emph{deep researcher agent} that interacts with external resources such as the internet or structured databases, thereby enriching the specification and producing an intermediate representation.  
\textit{(ii)} \textbf{World Model Generation} (\S\ref{sec:wmg}): At this stage, a \emph{developer agent} equipped with a code execution tool constructs the symbolic world model. The process is iteratively refined based on execution feedback, ensuring both correctness and executability.  
\textit{(iii)} \textbf{Evaluation-Driven Refinement} (\S\ref{sec:edr}): We enhance the semantic fidelity by designing two complementary \emph{test agents}: one that generates unit tests to validate functional behavior, and another that simulates downstream usage to evaluate performance through trajectory-based testing.  
We also provide a pseudo code in Algorithm~\ref{alg:method}.

\subsubsection{Stage I: Knowledge Synthesis}
\label{sec:ks}

We introduce a \emph{Deep Researcher} agent designed to gather background knowledge and fill in missing details that are not explicitly provided in the world model description. 
By leveraging external information sources, this agent not only compensates for potential knowledge gaps inherent in large language models but also enhances the factual reliability of world model descriptions. 
Equipped with web search and retrieval tools, it iteratively retrieves the knowledge required for world model construction from the internet and ultimately outputs a structured intermediate representation with the missing information completed.

\subsubsection{Stage II: World Model Generation}
\label{sec:wmg}

After obtaining the comprehensive world-model description from the previous stage, the \emph{Model Developer} takes this as input and generates a concrete implementation of the world model in the required formalism (e.g., PDDL or executable code). 
To support iterative refinement, the \emph{Model Developer} is equipped with a sandboxed code-execution tool, enabling it to test and debug implementations in multiple rounds until the code is functional and consistent with the specification.
% The MD not only synthesizes executable code but also updates the implementation in response to feedback from downstream evaluators. 
% This agent bridges the gap between abstract world descriptions and runnable system artifacts.

\subsubsection{Stage III: Evaluation-Driven Refinement}
\label{sec:edr}

A key component of our approach is the refinement of a bug-free, code-based world model. Unlike prior works that rely on annotated gold trajectories~\citep{dainese2024cwmb} or human feedback~\citep{guan2023leveraging}, our method is fully autonomous and does not require manual labels. 
More specifically, we introduce a two-agent \emph{Testing Team} to evaluate and diagnose the generated models:
\textit{(i)} The \emph{Unit Tester} conducts systematic, programmatic verification to validate the basic functionality of generated world model. It automatically generates Pytest-style unit tests targeting the predicates, actions, and invariants specified in the world-model descriptions. 
\textit{(ii)} The \emph{Simulation Tester} evaluates the world model in a play-testing manner by attempting to perform tasks, explore actions, and issue queries within the environment. Specifically, it interacts with the environment in a ReAct-style~\citep{yao2022react} loop to collect trajectories for subsequent behavior and reward analysis, which uncovers execution-time failures such as unreachable goals, missing preconditions, or inconsistent state updates.
% \textcolor{red}{Unit Tester check the basic functionality -> Simulation Tester check the validity in end-to-end tasks.}
Together, these agents produce a detailed \emph{test report} that assesses the quality of the generated world model and provides fine-grained diagnostic signals on correctness, coverage, logical consistency, and compliance with physical requirements. 
Unlike prior methods such as Text2World~\citep{hu2025text2world}, which uses PDDL validators, or GIF-MCTS~\citep{dainese2024cwmb}, which relies on a fixed set of offline agent trajectories, our Testing Team \emph{dynamically} synthesizes test cases based on the specific errors exhibited by each world model, enabling precision-guided debugging rather than generic checks.
This \emph{adaptive feedback} is propagated back to the \emph{Model Developer}; if inconsistencies or failures are detected, the Model Developer revises the implementation, triggering another evaluation round. This loop continues until all checks are satisfied or a predefined convergence criterion is reached.
% As we demonstrate in \S\ref{sec:training}, this adaptive mechanism is crucial not only for effective refinement but also for generating high-quality preference signals that enable downstream training.

% recorded in a \emph{Playtest Report}, which highlights the empirical shortcomings of the implementation.
% Execution of these tests produces a \emph{Test Report}, providing fine-grained diagnostic signals on correctness, coverage, and logical consistency.
% The Playtest and Test Reports are merged into a unified feedback signal, which is fed back to the Model Developer. 
% In this way, \ours achieves multi-round refinement, enabling both robustness and generalization in world-model generation.

\subsection{Training Paradigm}
\label{sec:training}

So far, \multi has been described as an inference-time multi-agent workflow in which a frozen backbone LLM plays all the agent roles.
However, the interaction between the Model Developer and the Testing Team naturally defines a learning environment that can be used to train more effective world-model agents.

\subsubsection{Agent-in-the-Loop Formalization}

We view each role in \multi as a tool-augmented LLM agent that interleaves generation with tool use.
Let $\mathcal{T}=\{t_1,\dots,t_m\}$ be the available tools (e.g., code execution, testing, retrieval).
At step $t$, the agent maintains a history
$h_t=(x, o_{\le t}, a_{<t})$,
where $x$ is the task specification, $o$ are tool observations (e.g., execution logs and test reports), and $a$ are past tool calls or edits.
A policy $\pi_\theta$ (implemented by an LLM) selects the next action: $a_t \sim \pi_\theta(\cdot \mid h_t),~ a_t \in \mathcal{T}.$

To connect this view to training, we model the \textbf{Model Developer} as an agent acting in an induced Markov Decision Process (MDP):
$\mathcal{M}_{\text{MD}}=(\mathcal{S},\mathcal{A},\mathcal{P},\mathcal{R},\gamma)$.
Here, the state $s_t\in\mathcal{S}$ concatenates the world-model specification with diagnostics produced by the Testing Team;
the action $a_t\in\mathcal{A}$ is a new implementation or a patch to the current world model.
The transition $\mathcal{P}$ is realized by executing the candidate model in a sandbox and re-running the Testing Team to obtain updated diagnostics.
The reward $\mathcal{R}(s_t,a_t)$ aggregates testing outcomes (e.g., unit-test pass rates and simulation performance), capturing both local correctness and downstream utility.
This formalization lets us treat world-model synthesis as sequential decision making and naturally yields multi-turn interaction trajectories for training.

\subsubsection{Verifier-Guided Rejection Sampling}
\label{sec:vgrs}

The agent-in-the-loop view naturally yields multi-turn interaction trajectories between the Model Developer and the Testing Team.
We leverage this interaction as a data engine to construct training trajectories without manual labels via \textbf{verifier-guided rejection sampling}. Given a world-model specification $x$, we run \multi to produce a sequence of developer proposals and feedback,
$\tau = \{(s_t, a_t, o_{t+1})\}_{t=0}^{T-1}$, where $o_{t+1}$ contains execution logs and testing diagnostics.
We define a verifier outcome $V(\tau)\in\{0,1\}$ based on the final candidate model produced in $\tau$.
Concretely, $V(\tau)=1$ if the final world model (i) executes successfully in the sandbox and (ii) satisfies the Testing Team's evaluation, i.e., it passes the synthesized unit tests and meets simulation-based checks (when applicable); otherwise $V(\tau)=0$.
Rejection sampling keeps only accepted trajectories: $\mathcal{D}_{\text{RS}} = \{(x,\tau)\;|\; V(\tau)=1\}.$
Intuitively, the Testing Team acts as a verifier that filters for executable and behaviorally consistent solutions, while preserving the intermediate repair steps.
As a result, $\mathcal{D}_{\text{RS}}$ contains multi-turn training traces that teach a Model Developer policy to iteratively revise world models under execution-grounded feedback, rather than producing a single-shot solution.
% In practice, we apply the same acceptance criterion across representations by delegating representation-specific checks (e.g., PDDL parsing/validation or Python execution) to lightweight tool adapters, while keeping the overall rejection-sampling procedure unchanged.

\section{Experiments}

In this section, we first describe the training dataset (\S~\ref{sec:data}), baselines (\S~\ref{sec:baseline}), and implementation details (\S~\ref{sec:implementation}), and then present experiments on three benchmarks:  
\textit{(i)} \textbf{\emph{Text2World}} \citep{hu2025text2world} (\S~\ref{sec:t2w}): A PDDL-centric benchmark for text-to-symbolic world modelling.  
\textit{(ii)} \textbf{\emph{Code World Models Benchmark (CWMB)}} \citep{dainese2024cwmb} (\S~\ref{sec:cwmb}):
A code-based world-model benchmark comprising MuJoCo-style environments, designed to assess \emph{predictive correctness} and \emph{downstream control utility} under both discrete and continuous action settings.  
\textit{(iii)} \textbf{\emph{ByteSized32}} \citep{bytesized32} (\S~\ref{sec:bytesized32}): 
A suite of reasoning-heavy text games requiring executable Python environments.
% A side-by-side comparison is shown in Table~\ref{tab:benchmark_overview}.
A side-by-side comparison and a detailed metric explanation are shown in Appendix~\ref{sec:benchmark_detail}.

% We evaluate two families of baselines: (i) PDDL-based, including Desc2Domain \citep{hu2025text2world} ( text-to-PDDL pipeline) and SingleAgent (multi-step tool-calling baseline); and (ii) Python-based, including WorldCoder \citep{tang2024worldcoder} (plan--code--execute--repair loop), GIF-MCTS \citep{dainese2024cwmb} (the MCTS-guided Generate/Improve/Fix method introduced with CWMB), the ByteSized32 official checker baseline \citep{bytesized32}, and SingleAgent. \textit{Leakage control:} ByteSized32 provides automated checks (e.g., runnability, winnability, reward) that some pipelines may use as generation-time feedback; to prevent metric leakage, we do not use these signals during generation. All re-run methods use the same backbone model and identical tool permissions and budget/step limits.

\subsection{Dataset Construction}
\label{sec:data}

Our data construction follows a staged pipeline in which we 
(\textit{i}) synthesize diverse world-model specifications spanning multiple representations (PDDL, Mujoco-style environments, text games, MCP-style tool environments). 
% We also  the domains (physics simulation, game logic, planning, \emph{etc.}); 
(\textit{ii}) run \ours with \texttt{gpt-4.1-mini} to generate executable world models and record the full multi-agent interaction traces (including iterative repairs under execution feedback); \textit{iii} apply verification mechanisms, such as legality, executability, and compliance checks, while combining semantic quality assessments regarding completeness, rationality, behavioral diversity, and execution consistency. Only retain the trajectories that meet these criteria.More detail show in Appendix~\ref{app:train_data_construction}. 
% and (\textit{iii}) apply verifier-guided filtering, retaining only trajectories whose final artifacts pass both unit tests and execution-/simulation-based validation.
% In total, we generate and log 2{,}400 candidate interaction trajectories (500 CWMB, 500 ByteSized32, 600 PDDL, and 800 MCP). 
% After filtering, 1{,}526 trajectories remain, comprising 266 CWMB, 277 ByteSized32, 483 PDDL, and 500 MCP trajectories.
To reduce the risk of training--evaluation leakage, we avoid using any benchmark instances as prompts or templates for rewriting/augmentation. Instead, we follow the environment synthesis method in AgentGen~\citep{hu2025agentgen} and using LIMA~\citep{zhou2023limaalignment} dataset as the inspiration corpus.

% This design minimizes direct overlap with existing benchmarks and mitigates reverse-inference risks.

% \emph{(ii)} for each specification, we deploy \multi to generate world models, recording the full interaction trace between the Model Developer and Testing Team agents;
% \emph{(iii)} we apply the Testing Team's aggregated reward signal $\mathcal{R}(s_t, a_t)$ as a quality filter, retaining only trajectories where the final world model passes both unit tests and simulation-based validation; and
% \emph{(iv)} for specifications with multiple candidate solutions, we construct preference pairs by comparing the Testing Team's feedback signals, creating data suitable for preference-based optimization. \\
% Additionally, to reduce the risk of information leakage between training data and evaluation benchmarks, we avoid using any benchmark instances as prompts or templates for rewriting or augmentation. 
% % to guide the generation of task specifications with similar structural constraints, while keeping concrete instances, parameters, and detailed configurations independent. 
% This design minimizes direct overlap with existing benchmarks and reduces the possibility of reverse inference.

\subsection{Baselines}
\label{sec:baseline}

We compare \multi against the following methods:

\noindent\textit{(i) \textbf{Direct Generation (Direct)}}: Single-shot generation of the symbolic world model without tool use, external retrieval, or feedback.  
\noindent\textit{(ii) \textbf{Agent2World${_\text{single}}$}}:
A single agent closes the loop by invoking code execution/validators/web search tools for self-repair and information synthesis, without multi-agent specialization.  
\noindent\textit{(iii) \textbf{Text2World (EC=k)~\citep{hu2025text2world}}}: directly using large language models to generate PDDL-based world model and iteratively repairing with planner/validator signals, where EC denotes the error-correction budget.  
\noindent\textit{(iv) \textbf{WorldCoder~\citep{tang2024worldcoder}}}:
A plan–code–execute–repair search that scores and iteratively improves candidate programs using simulator/planner signals to select runnable hypotheses.  
\noindent\textit{(v) \textbf{GIF\text{-}MCTS~\citep{dainese2024cwmb}}}:
A macro-action MCTS that orchestrates Generate/Improve/Fix steps, guided by unit tests and trajectory-based feedback for code world-model synthesis. \revinline{We also introduced an enhanced version of GIF-MCTS where a Deep Researcher agent gathers additional research data.}
\noindent\textit{(vi) \textbf{ByteSized32 baseline~\citep{bytesized32}}}.
The reference pipeline introduced by~\citet{bytesized32}. In order to avoid metric leakage, we do not use the official checker’s evaluation signals.
\revinline{\noindent\textit{(vii) \textbf{Best-of-N~\citep{stiennon2020learning,yu2025generating}}}: A method that performs reasoning over multiple samples and selects the best result.}
\revinline{\noindent\textit{(viii) \textbf{Self-Consistency~\citep{wang2022self}}}: A multi-sample reasoning method that votes over the results to improve consistency in decision-making.}

\subsection{Implementation Details}
\label{sec:implementation}
We employ the OpenAI GPT-4.1-mini model via~\href{https://platform.openai.com/docs/overview}{the official API} \revinline{and Llama-3.1-8b-instruct via the official Huggingface repo. We set the decoding temperature to 0 and top\_p to 1 for deterministic reproducibility.}
All agents operate within a ReAct~\citep{yao2022react} framework, following a "think $\rightarrow$ act (tool) $\rightarrow$ observe" loop for a maximum of 10 steps. 
% Each agent (e.g., \emph{Model Developer}, \emph{Simulation Tester}, \emph{Unit Tester}) is guided by role-defining prompts\ref{app:prompt_examples} that shape its behavior and workflow. 
The \emph{Deep Researcher} agent utilizes the \href{https://serper.dev/}{Serper API} for web searching. 
We blocked some websites to ensure experimental integrity and prevent information leakage~\footnote{For example, the original \href{https://huggingface.co/datasets/xdzouyd/text2world}{Text2World} and \href{https://huggingface.co/datasets/thuml/bytesized32-world-model-cot}{ByteSized32} huggingface pages, \href{https://github.com/nicoladainese96/code-world-models}{CWMB} source code, \href{https://github.com/openai/gym}{OpenAI-Gym} code repository are blocked.}
% , we implement a case-insensitive URL denylist during the retrieval phase, filtering out domains and pages that might contain solutions or detailed annotations; 
Regarding the configuration of refinement turns, we set Text2World and ByteSized32 to 2 iterations and CWMB to 3 iterations based on the complexity of environments.
For automated evaluation on the ByteSized32 benchmark, we leverage GPT-4o~\citep{gpt4o} as the LLM evaluator. 
\revinline{All experiments with gpt-4.1-mini are conducted on a CPU server without GPU acceleration. The experiments with llama-3.1-8b-instruct (including training and inference) are conducted with an 8xA100 server.}
The prompt examples could be found at Appendix~\ref{app:prompt_examples}.
As for the training experiments, we leverage the LlamaFactory~\citep{zheng2024llamafactory} to manage and execute the training procedure. 
We perform supervised fine-tuning (SFT) on the \texttt{llama-3.1-8b} models, and truncate input sequences to a maximum length of 30,000 tokens. 
We train the model for 5 epochs with learning rate of $1\times10^{-6}$.

\subsection{Text2World}
\label{sec:t2w}

\begin{table}[t!]
    \centering
    \caption{
    Benchmark results on Text2World~\citep{hu2025text2world}.
    Following the reporting convention in Text2World, all metrics are presented as percentage scores (\%).
    }
    \label{tab:text2world_metrics}
    \setlength{\tabcolsep}{6pt}
    \resizebox{\textwidth}{!}{%
    \begin{tabular}{llllllll}
        \toprule
        \multirow{2}{*}{\textbf{Methods}} & \multirow{2}{*}{\textbf{Executability} (\(\uparrow\))} & \multirow{2}{*}{\textbf{Similarity} (\(\uparrow\))} & \multicolumn{4}{c}{\textbf{Component-wise F1} (\(\uparrow\))} & \multirow{2}{*}{\textbf{F1$_{\textsc{\text{avg}}}$ (\(\uparrow\))}} \\
        \cmidrule(lr){4-7}
        & & & \fonepred & \foneparam & \foneprecond & \foneeff & \\
        \midrule
        \ourgray \texttt{gpt-4.1-mini} & & & & & & & \\
        Text2World$_{\text{EC=3}}$ & 78.2 & 81.1 & 73.4 & 64.5 & 49.3 & 53.3 & 60.1 \\
        Direct Generation          & 45.5 & 82.8 & 45.0 & 40.3 & 33.9 & 34.9 & 38.5 \\

        GIF-MCTS          & 44.6 & 81.1 & 42.8 & 39.4 & 33.4 & 32.2 & 36.9 \\

        Best-of-N          & 54.5 & 84.0 & 54.1 & 53.5 & 44.9 & 47.0 & 49.9 \\

        Self-Consistency          & 50.5 & \textbf{84.2} & 50.4 & 48.5 & 41.2 & 43.0 & 45.8 \\
        \single                    & 79.2 & 82.5 & 76.5 & 75.8 & 60.1 & 66.0 & 69.6 \\
        % \ourblue \multi            & \textbf{81.2} & 81.2 & \textbf{79.1} & 74.9 & \textbf{60.5} & 65.4 & \textbf{70.0} \\
        \ourblue \multi            & \textbf{93.1} & 81.0 & \textbf{87.2} & \textbf{82.3} & \textbf{63.7} & \textbf{68.2} & \textbf{75.4} \\
        % \ourgray \emph{Single} $\rightarrow$ \emph{Multi} & \textbf{93.1} & 81.0 & \textbf{87.2} & \textbf{82.3} & \textbf{63.7} & \textbf{68.2} & \textbf{75.4} \\
        \midrule
        \ourgray \texttt{llama-3.1-8b-instruct} & & & & & & &  \\
        Text2World$_{\text{EC=3}}$ & 0.0 & 74.9 & 0.0 & 0.0 & 0.0 & 0.0 & 0.0 \\
        Direct Generation          & 5.0 & 79.0 & 4.0 & 3.9 & 2.2 & 1.9 & 3.0 \\
        GIF-MCTS          & 4.0 & 72.9 & 0.0 & 0.0 & 0.0 & 0.0 & 0.0 \\
        Best-of-N          & 4.0 & 79.1 & 3.9 & 3.7  & 1.7 & 2.0 & 2.8 \\
        Self-Consistency          & 2.0 & 79.1 & 1.0 & 0.2 & 0.0 & 0.0 & 0.3 \\
        \single          & 19.8 & 81.9 & 19.5 & 18.7 & 13.8 & 12.8 & 16.2 \\
        \ourblue \multi            & 27.7 & 81.0 & 27.5 & 25.5 & 16.4 & 17.8 & 21.8 \\
        \ourblue \multi (SFT) &
        \textbf{44.6} {\footnotesize\color{ForestGreen}+16.9} &
        \textbf{81.9} {\footnotesize\color{ForestGreen}+0.9} &
        \textbf{41.4} {\footnotesize\color{ForestGreen}+13.9} &
        \textbf{32.6} {\footnotesize\color{ForestGreen}+7.1} &
        \textbf{17.4} {\footnotesize\color{ForestGreen}+1.0} &
        \textbf{21.5} {\footnotesize\color{ForestGreen}+3.7} &
        \textbf{28.2} {\footnotesize\color{ForestGreen}+6.4} \\

        \bottomrule
    \end{tabular} 
    }
\end{table}

% \begin{table}[h]
%     \centering
%     \caption{Benchmark results on Text2World~\citep{hu2025text2world}.}
%     \label{tab:text2world_metrics}
%     \resizebox{0.98\textwidth}{!}{%
%     \begin{tabular}{lccccccc}
%         \toprule
%         \multirow{2}{*}{\textbf{Methods}} & \multirow{2}{*}{\textbf{Executability} (\(\uparrow\))} & \multirow{2}{*}{\textbf{Similarity} (\(\uparrow\))} & \multicolumn{4}{c}{\textbf{Component-wise F1} (\(\uparrow\))} & \multirow{2}{*}{\textbf{F1$_{\textsc{\text{avg}}}$ (\(\uparrow\))}} \\
%         \cmidrule(lr){4-7}
%         & & & \fonepred & \foneparam & \foneprecond & \foneeff & \\
%         \midrule
%         Text2World$_{\text{EC=3}}$ & 78.2 & 81.1 & 73.4 & 64.5 & 49.3 & 53.3 & 60.1 \\
%         Direct Generation          & 45.5 & \textbf{82.8} & 45.0 & 40.3 & 33.9 & 34.9 & 38.5 \\
%         \single                    & 79.2 & 82.5 & 76.5 & 75.8 & 60.1 & 66.0 & 69.6 \\
%         % \ourblue \multi            & \textbf{81.2} & 81.2 & \textbf{79.1} & 74.9 & \textbf{60.5} & 65.4 & \textbf{70.0} \\
%         \ourblue \multi            & \textbf{92.73±1.42} & 80.60±0.32 & \textbf{87.27±1.50} & \textbf{81.03±2.46} & \textbf{62.13±1.40} & \textbf{67.5±0.92} & \textbf{74.48±1.31} \\
%         \bottomrule
%     \end{tabular} 
%     }
% \end{table}

We evaluate the Planning Domain Definition Language (PDDL)-based world model generation of \ours on Text2World~\cite{hu2025text2world}, which comprises 103 PDDL domains paired with natural language descriptions. 
The evaluation metrics are: 
\textit{(i)} \textbf{Executability}: whether the generated PDDL can be parsed and validated; 
\textit{(ii)} \textbf{Structural Similarity}: the normalized Levenshtein similarity; 
\textit{(iii)} \textbf{Component-wise F1}: the macro-averaged F1 of predicates (\fonepred) and action components, including parameters (\foneparam), preconditions (\foneprecond), and effects (\foneeff).

\paragraph{Results.} 
We can draw several conclusions from Table~\ref{tab:text2world_metrics}:
\textit{(i)} \textit{Direct Generation} attains the highest Similarity (82.8) yet performs poorly on executability (45.5) and all component-wise F1s, underscoring that surface-level textual overlap is a weak proxy for runnable, semantically correct PDDL. 
\textit{(ii)} While agent-based methods achieve executability comparable to the reference solution (e.g., \single with 79.2 vs. Text2World$_{\text{EC=3}}$ with 78.2), they exhibit substantial gaps in F1 scores (\single: 69.6 vs. Text2World$_{\text{EC=3}}$: 60.1). This suggests that while integrating validators for iterative correction can significantly improve syntactic validity, the semantic utility of the generated world models remains limited without comprehensive knowledge synthesis.
\textit{(iii)} \multi achieves both the highest executability (+14.9 points over Text2World$_{\text{EC=3}}$) and superior F1 performance (+15.3 points), demonstrating the synergistic benefits of multi-agent specialization.
These patterns align with our design philosophy: knowledge synthesis combined with evaluation-driven refinement steers the model to recover the correct predicate inventory and logical gating constraints, producing domains that are both \textit{syntactically valid} and \textit{semantically solvable}, even when surface-level representations diverge from reference implementations.
\textit{(iv)} Based on the Llama3.1-8b-instruct model, \multi(SFT) significantly improved executability (+16.9 points over \multi) and optimized F1 performance (+6.4 points over \multi). This demonstrates that our SFT data effectively reduced the model's errors in PDDL code and enhanced its understanding of task descriptions, improving the semantic structure of the generated code. 
In terms of structural similarity, the model scored 81.9, nearly equivalent to the Llama 4.1-mini model (81.0), indicating that the generated PDDL is structurally close to the reference standard.

\subsection{Code World Models Benchmark (CWMB)}
\label{sec:cwmb}

% \begin{table}[ht]
%   \centering
%   \caption{Benchmark results on CWMB~\citep{dainese2024cwmb}. $\dagger$ We adopted the official implementation of GIF-MCTS~\citep{dainese2024cwmb} and their reimplementation of WorldCoder~\citep{tang2024worldcoder}.
%   }
%   \label{tab:cwmb_metrics}
%   \footnotesize
%   \setlength{\tabcolsep}{5pt} 
%   \renewcommand{\arraystretch}{1.1}
%   \resizebox{\textwidth}{!}{%
%   \begin{tabular}{
%       l
%       S[table-format=1.4] S[table-format=1.4]
%       S[table-format=1.4] S[table-format=1.4]
%   }
%     \toprule
%     \multirow{2}{*}{\textbf{Method}} &
%     \multicolumn{2}{c}{\textbf{Discrete Action Space}} &
%     \multicolumn{2}{c}{\textbf{Continuous Action Space}} \\
%     \cmidrule(lr){2-3}\cmidrule(lr){4-5}
%      & 
%      \multicolumn{1}{c}{Accuracy (\(\uparrow\))} &
%      \multicolumn{1}{c}{\(\mathcal{R}\) (\(\uparrow\))} &
%      \multicolumn{1}{c}{Accuracy (\(\uparrow\))} &
%      \multicolumn{1}{c}{\(\mathcal{R}\) (\(\uparrow\))} \\
%     \midrule
%         WorldCoder$^\dagger$        & 0.9045 & 0.5432 & 0.3405 & 0.1899 \\
%         GIF-MCTS$^\dagger$          & 0.9136 & 0.6842 & 0.2748 & 0.1858 \\
%         Direct Generation & 0.7321 & 0.4527 & 0.3038 & 0.1666 \\
%         \single       & 0.6099 & 0.4196 & 0.3023 & 0.1451 \\
%         \ourblue \multi & 0.9174 & 0.8333 & 0.3575 & 0.3050 \\
%     \bottomrule
%   \end{tabular}%
%   }
% \end{table}

\begin{table}[t!]
  \centering
  \caption{Benchmark results on CWMB. $\dagger$ We adopted the official implementation of GIF-MCTS~\citep{dainese2024cwmb} and their reimplementation of WorldCoder~\citep{tang2024worldcoder}.
    \begin{revblock}
    GIF-MCTS$^\star$ denotes the enhanced version that we obtain by connecting the deep research agent with the original GIF-MCTS pipeline.
    \end{revblock}
  }
  \label{tab:cwmb_metrics}
  \footnotesize
  \setlength{\tabcolsep}{6pt} 
  \renewcommand{\arraystretch}{1.1}
  \resizebox{0.96\textwidth}{!}{%
  \begin{tabular}{
      lllllll
    }
    \toprule
    \multirow{2}{*}{\textbf{Method}} &
    \multicolumn{2}{c}{\textbf{Discrete Action Space}} &
    \multicolumn{2}{c}{\textbf{Continuous Action Space}} &
    \multicolumn{2}{c}{\textbf{Overall}} \\
    \cmidrule(lr){2-3}\cmidrule(lr){4-5}\cmidrule(lr){6-7}
     & 
     \multicolumn{1}{l}{Accuracy (\(\uparrow\))} &
     \multicolumn{1}{l}{\(\mathcal{R}\) (\(\uparrow\))} &
     \multicolumn{1}{l}{Accuracy (\(\uparrow\))} &
     \multicolumn{1}{l}{\(\mathcal{R}\) (\(\uparrow\))} &
     \multicolumn{1}{l}{Accuracy (\(\uparrow\))} &
     \multicolumn{1}{l}{\(\mathcal{R}\) (\(\uparrow\))} \\
    \midrule
        \ourgray \texttt{gpt-4.1-mini} & & & & & & \\
        WorldCoder$^\dagger$        & 0.9024 & 0.5399 & 0.3303 & 0.2097 & 0.5210 & 0.3197 \\
        GIF-MCTS$^\dagger$          & 0.9136 & 0.6842 & 0.2748 & 0.1811 & 0.4877 & 0.3488 \\
        GIF-MCTS$^\star$ & 0.8876 & 0.7955 & 0.3030 & 0.1495 & 0.4979 & 0.3649 \\
        Direct Generation & 0.7321 & 0.4527 & 0.3038 & 0.1666 & 0.4466 & 0.2620 \\
        Best-of-N  & 0.7488 & 0.6012 & 0.2642 & 0.1970 & 0.4257 & 0.3317 \\
        Self-Consistency & 0.7870 & 0.4912 & 0.2479 & 0.2158 & 0.4276 & 0.3076 \\
        \single       & 0.7897 & 0.5418 & 0.1917 & 0.2420 & 0.3911 & 0.3419 \\
        \ourblue \multi
          & {\bfseries 0.9174} & {\bfseries 0.8333}
          & {\bfseries 0.3575} & {\bfseries 0.3050}
          & {\bfseries 0.5441} & {\bfseries 0.4811} \\

% \ourred ~~~No \emph{Deep Researcher} & 0.8794 & 0.4407 & 0.3404 & 0.2201 & 0.5201 & 0.2936 \\
% \ourred ~~~No \emph{Simulation Tester} & 0.8920 & 0.5718 & 0.3288 & 0.1577 & 0.5275 & 0.3039 \\
% \ourred ~~~No \emph{Unit Tester} & 0.6166 & 0.3863 & 0.3025 & 0.1704 & 0.4072 & 0.2423 \\
    \midrule
        \ourgray \texttt{llama-3.1-8b-instruct} & & & & & & \\
        WorldCoder$^\dagger$        & 0.5192 & 0.2779 & 0.1173 & 0.1073 & 0.2513 & 0.1642 \\
        GIF-MCTS$^\dagger$          & 0.5983 & 0.3357 & 0.1332 & 0.1427 & 0.2883 & 0.2070 \\
        GIF-MCTS$^\star$ & 0.4691 & 0.3752 & 0.1934 & 0.1457 & 0.2853 & 0.2222 \\
        Best-of-N  & 0.2236 & 0.0004 & 0.1544 & 0.1404 & 0.1775 & 0.0937 \\
        Self-Consistency & 0.4626 & 0.1667 & 0.0000 & 0.0036 & 0.1542 & 0.0580 \\
        \single & 0.1531 & 0.0000 & 0.2690 & 0.1418  & 0.2304 & 0.0945 \\
        \ourblue \multi       & 0.5797 & 0.2998 & 0.1826 & 0.1945 & 0.3150 & 0.2296 \\
        \ourblue \multi(SFT) &
        \textbf{0.6857} {\footnotesize\color{ForestGreen}+0.1060} &
        \textbf{0.5333} {\footnotesize\color{ForestGreen}+0.2335} &
        \textbf{0.2203} {\footnotesize\color{ForestGreen}+0.0377} &
        \textbf{0.2128} {\footnotesize\color{ForestGreen}+0.0183} &
        \textbf{0.3754} {\footnotesize\color{ForestGreen}+0.0604} &
        \textbf{0.3197} {\footnotesize\color{ForestGreen}+0.0901} \\
    \bottomrule
  \end{tabular}%
  }
\end{table}

The CWMB~\citep{dainese2024cwmb} evaluates the ability of generated executable code to serve as faithful world models across 18 MuJoCo-style environments. 
It measures both the predictive accuracy of next-state dynamics and the normalized return ($\mathcal{R}$) when the model is used by a planner, where $\mathcal{R}$ reflects the gap between a random policy and an oracle planner with the true environment. 
This setup ensures CWMB jointly assesses the correctness of the simulation code and its practical utility for downstream control.

\paragraph{Results.}
Table~\ref{tab:cwmb_metrics} reveals several key findings. 
\textit{(i)} All methods demonstrate superior performance in discrete spaces compared to continuous settings, reflecting the inherent difficulty of modeling continuous dynamics.
\textit{(ii)} Workflow-based approaches consistently outperform both \emph{Direct Generation} and \single, indicating that LLMs' native world model generation capabilities are limited and require expert-designed iterative refinement to achieve competitive performance.
\textit{(iii)} \multi establishes new state-of-the-art results, surpassing the previous best method GIF-MCTS by +0.132 $\mathcal{R}$ points in overall normalized return.
Notably, while other methods achieve comparable predictive accuracy (e.g., 0.917 vs 0.914 on discrete settings), our simulation-based testing framework significantly enhances the downstream utility of generated world models, demonstrating that accurate next-state prediction alone is insufficient for effective model-based planning.
\emph{(iv)} \multi(SFT) further improves ByteSized32 by enhancing executability and specification compliance. It significantly improves the "runnable game" score (0.0849 points higher than multi) and greatly improves the compliance of "key actions" and "distractors" (0.2944 points higher than multi), thereby increasing the "winnable game" rate (0.0388 points higher than multi). The training data improves the model's understanding of task and execution-based iterative debugging, thereby reducing missing necessary components and generating more runnable implementations.
% in continuous control it likewise tops both accuracy (0.3575) and return (0.3050), a +0.119 $\mathcal{R}$ gain over GIF-MCTS. 
% Finally, \single underperforms even Direct Generation in discrete mode (accuracy 0.6099 vs.\ 0.7321; $\mathcal{R}$ 0.4196 vs.\ 0.4527), highlighting that unspecialized self-repair can be counterproductive; specialization and tool orchestration in \ourblue \multi are key to translating edits into control-relevant improvements.

\subsection{ByteSized32}
\label{sec:bytesized32}

% \begin{table}[!htbp]
%     \centering
%     \caption{Benchmark results on ByteSized32~\cite{bytesized32}. 
%     % The presented results of ByteSized32 is derived from its official reference solution.
%     }
%     \label{tab:bytesized32_metrics_hier}
%     \resizebox{\textwidth}{!}{%
%     \begin{tabular}{lcccccccc}
%         \toprule
%         \textbf{Method} 
%         & \multicolumn{3}{c}{\textbf{Technical Validity}} 
%         & \multicolumn{3}{c}{\textbf{Specification Compliance}} 
%         & \textbf{Winnability} 
%         & \textbf{Physical Reality Alignment} \\
%         \cmidrule(lr){2-4} \cmidrule(lr){5-7} \cmidrule(lr){8-8} \cmidrule(lr){9-9}
%         & Game Init. & Possible Actions & Runnable Game 
%         & Critical Objects & Critical Actions & Distractors 
%         & Winnable Game 
%         & Alignment Score \\
%         \midrule
%         ByteSized32   & 0.9792 & 0.9375 & 0.7292 & 0.9375 & 0.9375 & 0.8750 & 0.0625 & 0.0600 \\
%         Direct Generation & 0.9271 & 0.8854 & 0.7604 & 1.0000 & 0.9375 & \textbf{0.9375} & 0.1354 & 0.1584\\
%         \single   & 0.9792 & 0.9375 & 0.7708 & 1.0000 & \textbf{1.0000} & 0.8438 & 0.1354 & 0.1920 \\
%         \ourblue \multi   & \textbf{0.9896} & \textbf{0.9583} & \textbf{0.8958} & \textbf{1.0000} & 0.9688 & 0.8750 & \textbf{0.1458} & \textbf{0.4768} \\
%         \bottomrule
%     \end{tabular}
%     }
% \end{table}

% 第一个表格：Technical Validity 和 Physical Reality Alignment
\begin{table}[t!]
    \centering
    \caption{Technical Validity and Physical Reality Alignment (Physical Align.) scores on ByteSized32.
    % -. The presented results of ByteSized32 is derived from its official reference solution.
    }
    \label{tab:bytesized32_technical_physical}
    \setlength{\tabcolsep}{6pt}
    \resizebox{0.96\textwidth}{!}{%
    \begin{tabular}{lllll}
        \toprule
        \textbf{Method} 
        & \multicolumn{3}{c}{\textbf{Technical Validity} (\(\uparrow\))} 
        & \textbf{Physical Align.} (\(\uparrow\)) \\
        \cmidrule(lr){2-4} \cmidrule(lr){5-5}
        & Game Init. & Possible Actions & Runnable Game 
        & Alignment Score \\
        \midrule
        \ourgray \texttt{gpt-4.1-mini} & & & & \\
        ByteSized32$_{\text{0-shot}}$   & 0.9792 & 0.9375 & 0.7292 & 0.0600 \\
        ByteSized32$_{\text{1-shot}}$  & 0.9792 & 0.8958 & 0.7500 & 0.1748 \\

        GIF-MCTS  & 0.8021 & 0.7083 & 0.4375 & 0.0656 \\
        Direct Generation & 0.9271 & 0.8854 & 0.7604 & 0.0738\\

        Best-of-N & 0.9583 & 0.8750 & 0.7604 & 0.2347\\

        Self-Consistency & 0.9688 & 0.8021 & 0.6875 & 0.3562\\
        \single   & 0.9792 & 0.9375 & 0.7708 & 0.1920 \\
        \ourblue \multi   & \textbf{0.9896} & \textbf{0.9583} & \textbf{0.8958} & \textbf{0.4768} \\
        \midrule
        \ourgray \texttt{llama-3.1-8b-instruct} & & & & \\
        ByteSized32$_{\text{0-shot}}$  & 0.5417 & 0.3958 & \textbf{0.3125} & 0.0023 \\
        ByteSized32$_{\text{1-shot}}$  & 0.6458 & 0.5833 & 0.2540 & 0.0287 \\
        Direct Generation & 0.4286 & 0.3333 & 0.1458 & 0.0168\\
        GIF-MCTS & 0.4167 & 0.3021 & 0.125 & 0.0237\\
        Best-of-N & 0.4479 & 0.3854 & 0.1042 & 0.0660\\
        Self-Consistency & 0.5729 & 0.4792 & 0.1354 & 0.0424\\
        \single & 0.5474 & 0.4842 & 0.1579 & 0.0281\\
        \ourblue \multi   & 0.7850 & 0.6355 & 0.1963 & 0.0837 \\
        \ourblue \multi(SFT) &
        \textbf{0.8229} {\footnotesize\color{ForestGreen}+0.0379} &
        \textbf{0.6458} {\footnotesize\color{ForestGreen}+0.0103} &
        0.2812 {\footnotesize\color{ForestGreen}+0.0849} &
        \textbf{0.1040} {\footnotesize\color{ForestGreen}+0.0203} \\
        
        \bottomrule
    \end{tabular}
    }
\end{table}

% 第二个表格：Specification Compliance 和 Winnability
\begin{table}[h]
    \centering
    \caption{Specification Compliance and Winnability scores on ByteSized32.
    % - . The presented results of ByteSized32 is derived from its official reference solution.
    }
    \label{tab:bytesized32_spec_winnability}
    \setlength{\tabcolsep}{6pt}
    \resizebox{0.9\textwidth}{!}{%
    \begin{tabular}{lllll}
        \toprule
        \textbf{Method} 
        & \multicolumn{3}{c}{\textbf{Specification Compliance} (\(\uparrow\))} 
        & \textbf{Winnability} (\(\uparrow\)) \\
        \cmidrule(lr){2-4} \cmidrule(lr){5-5}
        & Critical Objects & Critical Actions & Distractors 
        & Winnable Game \\
        \midrule
        \ourgray \texttt{gpt-4.1-mini} & & & & \\
        ByteSized32$_{\text{0-shot}}$   & 0.9375 & 0.9375 & 0.8750 & 0.0625 \\
        ByteSized32$_{\text{1-shot}}$   & \textbf{1.0000} & 0.9375 & \textbf{0.9375} & 0.1354 \\
        GIF-MCTS & 0.9062 & 0.9375 & \textbf{0.9375} & 0.0729\\
        Direct Generation & \textbf{1.0000} & 0.9375 & 0.9375 & 0.1354\\
        Best-of-N & \textbf{1.0000} & 0.9375 & 0.9062 & 0.1354 \\
        Self-Consistency & \textbf{1.0000} & 0.9375 & 0.8125 & 0.1267\\
        \single   & \textbf{1.0000} & \textbf{1.0000} & 0.8438 & 0.1354 \\
        \ourblue 
        \multi   & \textbf{1.0000} & 0.9688 & 0.8750 & \textbf{0.1458} \\
        \midrule
        \ourgray \texttt{llama-3.1-8b-instruct} & & & & \\
        ByteSized32$_{\text{0-shot}}$  & \textbf{1.0000} & 0.8125 & 0.0938 & 0.0000 \\
        ByteSized32$_{\text{1-shot}}$  & 0.9688 & 0.8125 & 0.0938 & 0.0433 \\
        Direct Generation & 0.9375 & 0.7812 & 0.0312 & 0.0000 \\
        GIF-MCTS & \textbf{1.0000} & 0.7812 & \textbf{0.1875} & 0.0312 \\
        Best-of-N & 0.9688 & 0.8438 & 0.0625 & 0.0417\\
        Self-Consistency & 0.9688 & 0.6250 & 0.0312 & 0.0433\\
        \single   & 0.9688 & 0.6875 & 0.0323 & 0.0433 \\
        \ourblue \multi   & 0.9688 & 0.6744 & 0.0938 & 0.0654 \\
        \ourblue \multi(SFT) &
        0.9688 {\footnotesize\color{ForestGreen}+0.0000} &
        \textbf{0.9688} {\footnotesize\color{ForestGreen}+0.2944} &
        \textbf{0.1875} {\footnotesize\color{ForestGreen}+0.0937} &
        \textbf{0.1042} {\footnotesize\color{ForestGreen}+0.0388} \\

        \bottomrule
    \end{tabular}
    }
\end{table}

The ByteSized32~\citep{bytesized32} benchmark consists of 32 reasoning-heavy text games, each implemented as an executable Python environment. 
Models are required to generate runnable game code that captures task-specific dynamics, objects, and rules, allowing direct interaction and evaluation. 
The benchmark evaluates four dimensions: 
\textbf{Technical Validity} (whether the code runs),
\textbf{Specification Compliance} (whether all required elements are present), 
\textbf{Winnability} (whether the task can be completed), 
and \textbf{Physical Reality Alignment} (whether the environment dynamics are consistent with commonsense constraints). 
This setting emphasizes both logical fidelity and practical executability, making it a stringent testbed for language models as world-model generators.

\paragraph{Results.}
Several conclusions could be drawn from Table~\ref{tab:bytesized32_technical_physical} and~\ref{tab:bytesized32_spec_winnability}:
\emph{(i)} The official reference pipeline outperforms direct generation with in-context learning and shows comparable performance to \single on certain metrics.
\emph{(ii)} The \single baseline shows moderate gains in game solvability, yet its alignment with physical reality is slightly weaker. 
\emph{(iii)} \multi outperforms both baselines across almost all dimensions, especially $+0.2848$ physical alignment score, which stems from \emph{Deep Researcher} agent synthesizing the commonsense knowledge required for reasoning-heavy games.
\emph{(iv)}
\section{Analysis}

\begin{figure}[t]
  \centering
  \begin{minipage}[t]{0.47\linewidth}
    \centering
    \includegraphics[width=\linewidth]{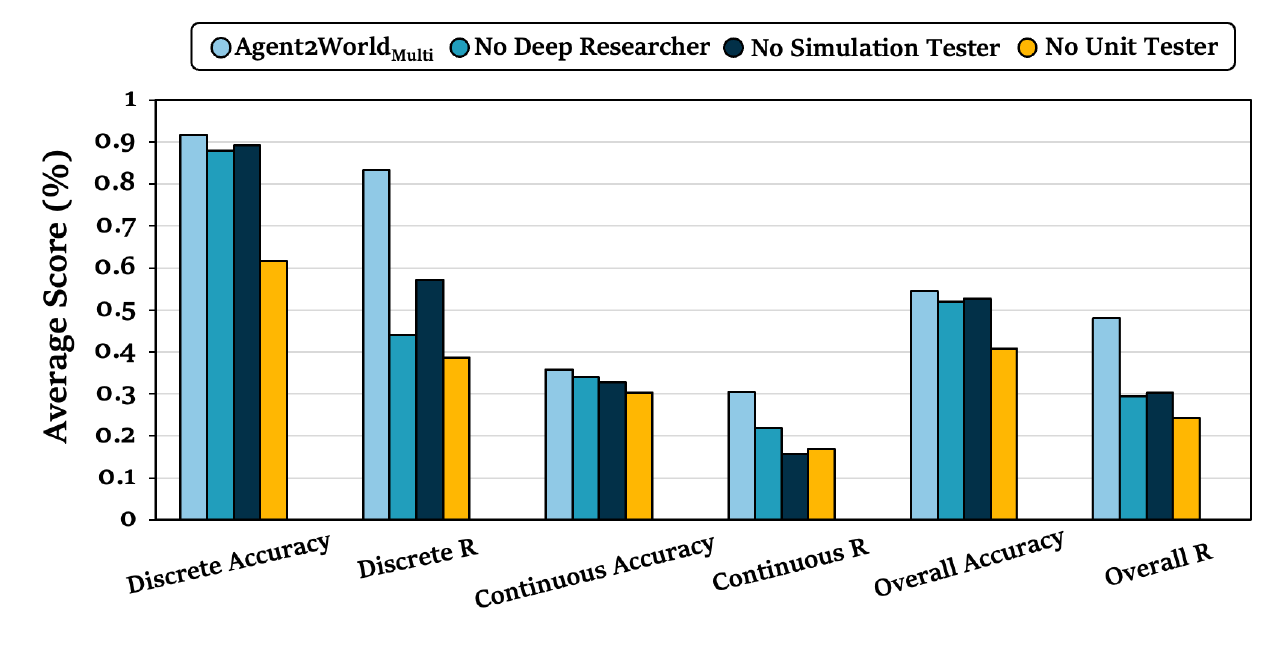}
    \captionof{figure}{Ablation Study on CWMB.
    % ~\citep{dainese2024cwmb}
    }
    \label{fig:gif_mcts_ablation}
  \end{minipage}\hfill
  \begin{minipage}[t]{0.52\linewidth}
    \centering
    \includegraphics[width=\linewidth]{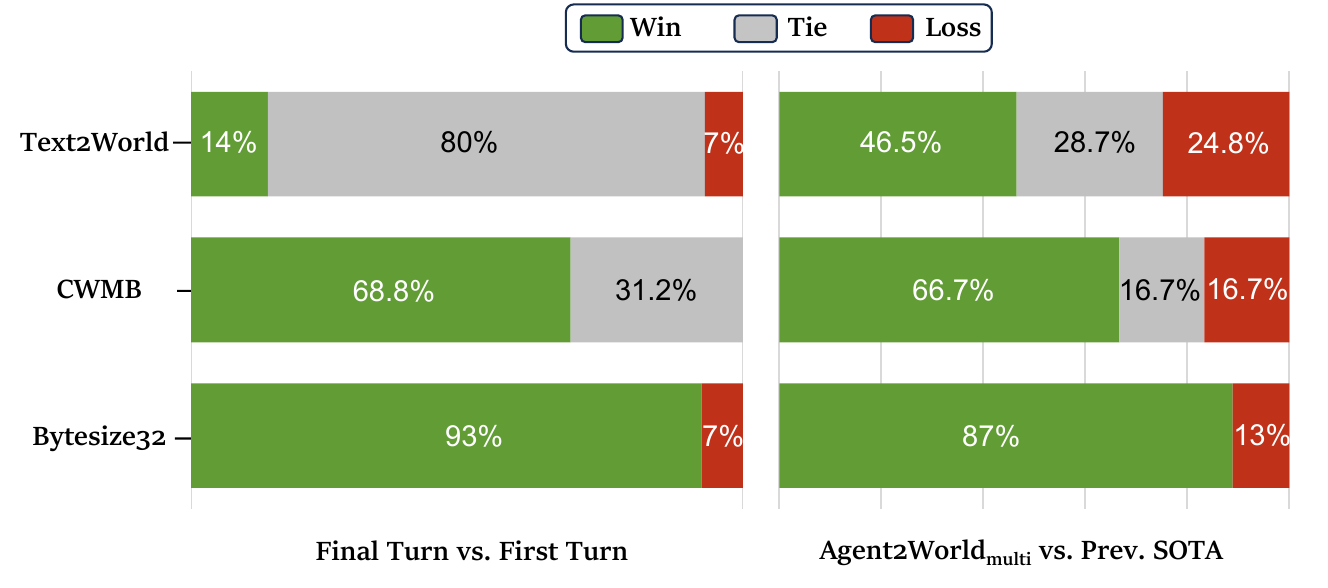}
    \captionof{figure}{
    Pair-wise Win–Tie–Loss analysis. 
    % Left: \multi\ final turn vs.\ first turn (effect of refinement). Right: \multi\ vs.\ previous state-of-the-art methods.
    }
    \label{fig:Win–Tie–Loss}
  \end{minipage}
\end{figure}

\subsection{Ablation Study}

Ablations from Figure~\ref{fig:gif_mcts_ablation} and Table~\ref{tab:gif_mcts_ablation} clarify where the lift comes from:
\emph{(i)} Removing the \emph{Unit Tester} causes the most significant performance drop, with accuracy declining by 0.3008 and reward $\mathcal{R}$ by 0.4470 in discrete action spaces.
\emph{(ii)} The \emph{Deep Researcher} primarily impacts reward $\mathcal{R}$ quality, showing a substantial decrease of 0.3926 for discrete spaces when removed. 
\emph{(iii)} Although the removal of \emph{Simulation Tester} results in the smallest overall performance drops, the reward $\mathcal{R}$ decreases by 0.2615 and 0.1473 for discrete space and continuous space, respectively.
These results collectively validate our design choices and highlight the complementary nature of the three components.

\subsection{Pair-wise Evaluation}
\label{sec:pair_wise}

To quantify the effect of \multi and the refinement procedure, we perform instance-level pairwise comparisons, recording a Win–Tie–Loss (WTL) outcome according to the benchmarks' primary metric: 
\emph{(i)} F1$_{\textsc{\text{avg}}}$ for \emph{Text2World};
\emph{(ii)} $\mathcal{R}$ for \emph{CWMB};
and \emph{(iii)} the mean of all official metrics for \emph{ByteSized32}. As shown in Figure~\ref{fig:Win–Tie–Loss}, the \emph{left} panel contrasts the final-turn model with its first-turn counterpart. Refinement yields consistent gains on \emph{CWMB} and \emph{ByteSized32} (68.8\% and 93\% wins, respectively;  no losses), largely preserves performance on \emph{Text2World} while delivering occasional improvements (14\% wins vs.\ 7\% losses).
% , and produces substantial improvements on \emph{ByteSized32} (93\% wins, 7\% losses). 
The \emph{right} panel compares \multi\ against previous state-of-the-art systems. \multi\ attains clear advantages across all three benchmarks, most notably on \emph{ByteSized32} (87\% wins) and \emph{CWMB} (66.7\% wins).

% To assess the effectiveness of our testing team refinement process, we conduct a systematic comparison between the initial world model and the refined version across three established benchmarks. For each benchmark, we directly compare the performance metrics and categorize the results into three outcomes: \textit{win} (refined model outperforms initial model), \textit{tie} (performance remains equivalent), and \textit{lose} (refined model underperforms compared to initial model).

% \begin{figure}[thb]
%     \centering
%     \includegraphics[width=0.80\linewidth]{images/Win–Tie–Loss.pdf}
%     \caption{Win–Tie–Loss comparison of \ours vs. baseline}
%     \label{fig:Win–Tie–Loss}
% \end{figure}

% These results provide strong empirical evidence that our testing team refinement process effectively enhances world model performance across diverse evaluation scenarios. The consistent improvement pattern across multiple benchmarks suggests that the refinement process captures generalizable insights rather than benchmark-specific optimizations.

\subsection{Feedback Iteration}
\label{subsec:feedback_analysis}

\begin{figure}[t]
\centering
% Include your line plot here
\includegraphics[width=0.99\textwidth]{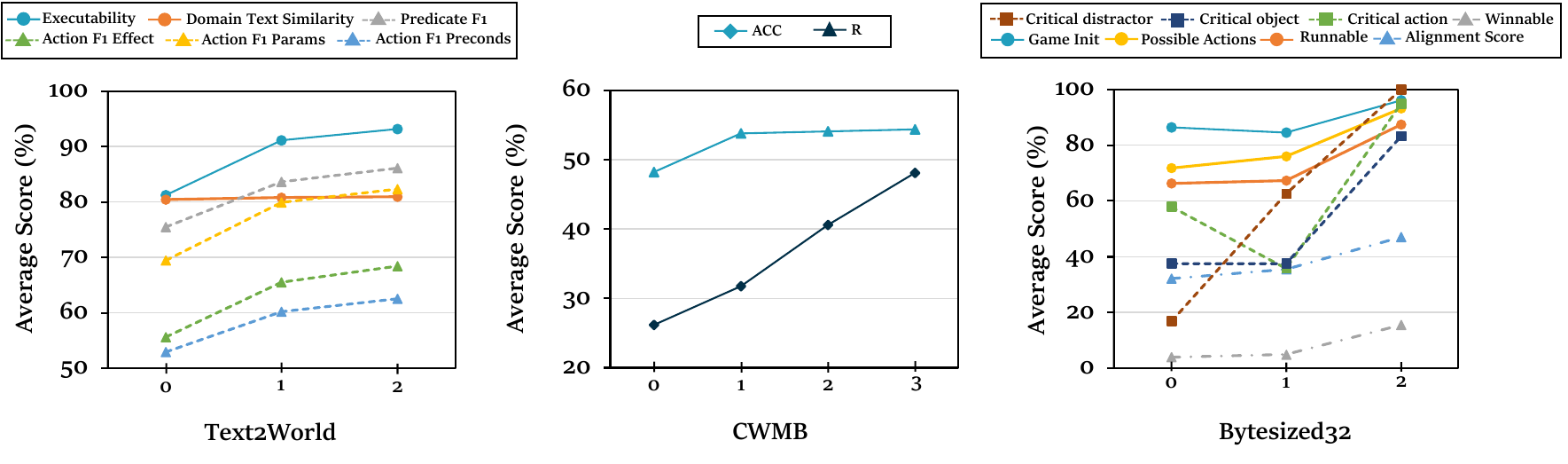}
\caption{Primary performance evolution across successive rounds of testing team feedback (\S~\ref{sec:edr}.)
% as a function of testing team feedback iterations. Each line represents performance on a different benchmark, showing the impact of iterative refinement on model capabilities.
}
\label{fig:feedback_performance}
\end{figure}

To understand the dynamics of performance improvement through iterative feedback, we analyze how model performance evolves as the number of testing team feedback iterations increases. 
Figure~\ref{fig:feedback_performance} illustrates the relationship between feedback iteration count and model performance across the evaluation benchmarks. The results reveal several key patterns:
\emph{(i)} Text2World shows rapid initial improvements. Notably, execution-based metrics improve substantially while similarity measures remain stable, suggesting that refinement enhances functional correctness rather than surface-level similarities. 
\emph{(ii)} CWMB demonstrates sustained improvement across iterations, reflecting the compound complexity of physics simulation where numerical accuracy and dynamics must be jointly optimized. 
\emph{(iii)} ByteSized32 exhibits the most dramatic gains, with several metrics showing step-function improvements that reflect the discrete nature of game logic debugging. 

\subsection{Manual Error Analysis}

% \begin{figure}[t]
%   \centering
%   \begin{minipage}[t]{0.57\linewidth}
%     \centering
%     \includegraphics[width=\linewidth]{images/sankey_diagram_cwmb.pdf}
%     \captionof{figure}{Error distribution on CWMB of evaluation-driven refinement. Due to page limit, the error distribution of other benchmarks is presented in Appendix~\ref{app:detailed_error_analysis}.}
%     \label{fig:sankey_diagram_cwmb}
%   \end{minipage}\hfill
%   \begin{minipage}[t]{0.41\linewidth}
%     \centering
%     \includegraphics[width=\linewidth]{images/specialization_efficiency.pdf}
%     \captionof{figure}{Performance comparisons and specialized efficiency of multi-agent and single-agent architectures.}
%     \label{fig:specialization_efficiency}
%   \end{minipage}
% \end{figure}

\begin{figure}[t]
  \centering
  \includegraphics[width=0.84\linewidth]{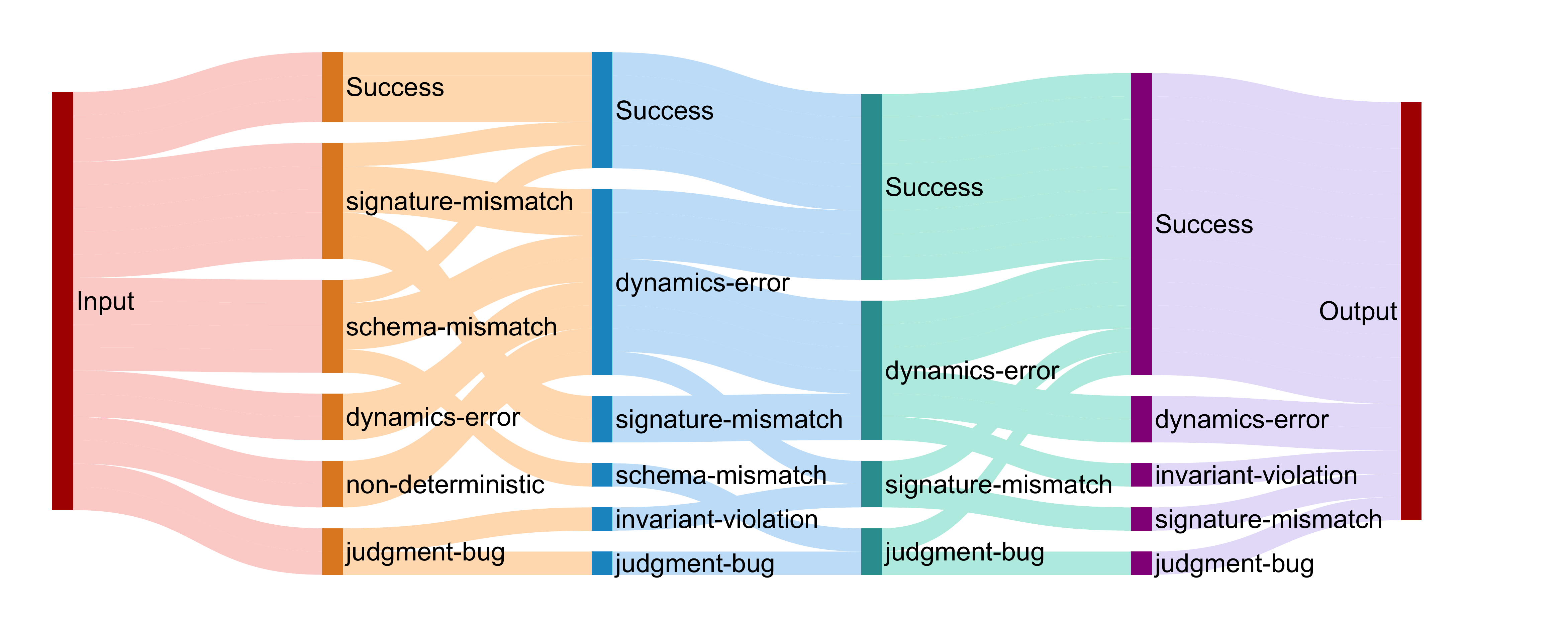}
  \caption{Error distribution on CWMB of evaluation-driven refinement. Due to page limit, the error distribution of other benchmarks is presented in Appendix~\ref{app:detailed_error_analysis}.}
  \label{fig:sankey_diagram_cwmb}
\end{figure}

We conducted a manual error analysis to examine the evolution of error patterns throughout the refinement process of \multi.
Taken CWMB in Figure~\ref{fig:sankey_diagram_cwmb} as an example, the initial turn predominantly exhibits superficial errors like \textit{signature-arity mismatches} and \textit{representation mismatches}, stemming from inadequate adherence to world model specifications.
Throughout the iterative refinement process, these surface-level inconsistencies are systematically eliminated, with the error landscape shifting toward more fundamental \textit{dynamics mismatches} in later iterations. 
This pattern demonstrates remarkable consistency across all benchmarks: refinement consistently shifts the error distribution from form-oriented problems (\textit{syntax}, \textit{arity}) to substance-oriented challenges (\textit{dynamics}, \textit{state transitions}) as shown in Figure~\ref{fig:sankey_diagram_text2world} and~\ref{fig:sankey_diagram_Bytesized32}. 
The systematic progression from surface to substance reflects the hierarchical nature of world model correctness and validates our multi-turn refinement architecture.
We also provide the detailed proportion of each error type in Appendix~\ref{app:detailed_error_analysis}.

% We visualize error/state flows over three refinement rounds for \textsc{Text2World}, \textsc{WMB}, and \textsc{Bytesized32}; link width encodes sample mass and “Success” expands across rounds.

% \textsc{Text2World}: superficial issues (parentheses/duplicates) diminish quickly, leaving a persistent tail of \emph{type/symbol} errors (\texttt{TypeMismatch}, \texttt{UndefinedConstant}).
% \textsc{WMB}: interface errors (\texttt{signature-arity}, \texttt{representation}) largely resolve, whereas \texttt{dynamics-mismatch} remains the dominant—and occasionally reappearing—failure after intermediate successes.
% \textsc{Bytesized32}: syntax and invalid-action shrink, but \texttt{state-bug} dominates the terminal failures, indicating logic over state transitions as the bottleneck.

\subsection{Multi-Agent vs. Single-Agent Architecture Analysis}

% To quantify the benefits of multi-agent specialization, we compare \multi against \single using each benchmark's primary metric as in Section~\ref{sec:pair_wise}, alongside specialization efficiency calculated as $\frac{\text{Performance Improvement (\%)}}{\text{Token Cost Increase (\%)}}$. 
% Figure~\ref{fig:specialization_efficiency} illustrates this \revinline{cost-performance trade-off}, showing specialization efficiencies of 4.6\% for Text2World, 10.5\% for CWMB, and 30.2\% for ByteSized32.
% % Figure~\ref{fig:specialization_efficiency} and the token cost experiments in Appendix~\ref{app:efficiency} analyzes the \textbf{cost-performance trade-off}. 
% \begin{revblock}
%     While \multi incurs a higher absolute token cost (as shown in Appendix~\ref{app:efficiency}) due to its proactive testing loop and additional knowledge synthesis stage, it delivers substantial performance gains. 
%     % Fix: Changed "performance improvement" to "specialization efficiency" to match the number's definition
%     Specifically, on token-intensive benchmarks like ByteSized32, the high specialization efficiency (30.2\%) justifies the additional computational investment. 
%     This indicates that for complex environments requiring high reliability, the multi-agent architecture offers a viable trade-off, investing upfront inference compute to ensure generation quality.
% \end{revblock}

\begin{revblock}
    To quantify the cost of multi-agent specialization, we analyze the token consumption reported in Appendix~\ref{app:efficiency}. 
Compared to \single, \multi incurs a higher computational cost during the generation phase. 
This increase stems from the proactive testing loop, where agents autonomously generate unit tests and simulation trajectories to diagnose errors.
However, this additional cost represents a finite \textbf{\textit{upfront investment}} rather than a recurring inefficiency, which is incurred only once during the synthesis of the world model. 
In exchange, the framework secures a permanent \textbf{\textit{performance gain}} in the quality of the generated artifact (e.g., raising the Normalized Return $\mathcal{R}$ from $0.3419$ to $0.4811$ on CWMB).
\end{revblock}

\section{Related Work}
\noindent \textbf{World Models.} 
World models are widely applied in reinforcement learning, robotics, and autonomous systems for planning, etc~\citep{rap,worldmodel}. 
Generally, there are two types of world models: 
(i) \emph{neural world models}, which employ neural networks to approximate dynamics~\citep{worldmodel,hafner2019dream}, 
and (ii) \emph{symbolic world models}, which are represented using formal languages such as the Planning Domain Definition Language (PDDL) or code-based implementations. 
In this paper, we investigate \emph{symbolic world-model generation}, which involves transforming world model descriptions into formal representations that can subsequently be utilized for applications such as model-based planning~\citep{guan2023leveraging,dainese2024cwmb}, dataset construction~\citep{hu2025agentgen}, and so on.
Most prior work follows a \emph{draft-repair} workflow where the model first proposes an initial implementation, then gradually refines under closed-loop diagnosis from some kind of feedback. 
The feedback mechanisms can vary significantly across different approaches. 
For instance, \citet{guan2023leveraging} employs human feedback to provide corrective signals and perform iterative modifications. 
Other works, such as \citet{hu2025text2world} and \citet{tang2024worldcoder}, utilize executors and validators to generate feedback. 
GIF-MCTS \citep{dainese2024cwmb} leverages gold experiences as feedback signals.
Compared to these scripted workflows, the \ours paradigm introduced in this paper can more flexibly adjust subsequent strategies based on feedback signals. 

\noindent \textbf{Large Language Model-based Agent.}
In recent years, benefiting from the rapid advancement of large language models (LLMs), LLM-based agents have emerged as particularly powerful systems that accept natural language user intentions as input and achieve goal states through planning and sequential decision-making \citep{hu2024hiagent,yao2022react,schick2023toolformer}.
These autonomous agents have demonstrated remarkable effectiveness across diverse applications, ranging from web navigation~\citep{webshop,webgpt,wang2025x} and software development~\citep{chatdev,metagpt} to scientific research~\citep{lu2024ai,chen2025ai4research} and robotic planning~\citep{huang2022language}.
Prominent examples of such systems include ReAct \citep{yao2022react}, which synergizes reasoning and acting in language models by interleaving thought, action, and observation steps; 
% WebGPT \citep{webgpt}, which enables agents to browse the web to answer questions; 
Existing research has explored how world models can assist LLM-based agents in planning, such as RAP~\citep{rap}, which uses Monte Carlo Tree Search with world models for improved reasoning, and~\citet{guan2023leveraging}, which leverages pre-trained LLMs to construct world models for model-based task planning.
These approaches primarily focus on \emph{utilizing} existing world models rather than \emph{generating} them. 
Similarly, recent work has investigated how world models can enhance training of LLM-based agents, as demonstrated by AgentGen \citep{hu2025agentgen} and Kimi-K2~\citep{team2025kimi}.
% However, to our best knowledge, no prior work has systematically explored how to harness the collaborative capabilities of LLM-based agents for world model construction itself. 
% While workflow-based approaches have been proposed for symbolic world model generation~\citep{hu2025text2world, guan2023leveraging}, the potential of autonomous agent remains largely unexplored in this domain. 
% Our work represents the first systematic investigation into using autonomous agents for comprehensive world model generation, bridging the gap between agent-based problem-solving and symbolic world modeling.
To our best knowledge, our work represents the first systematic investigation into using autonomous agents for world model generation, bridging the gap between agent-based problem solving and symbolic world modeling.

\section{Conclusion}

We introduced \multi, a unified multi-agent framework that employs autonomous LLM-based agents to generate symbolic world models across both PDDL and executable code representations. 
The framework operates through three specialized stages: knowledge synthesis via web search, world model development with iterative refinement, and evaluation-driven testing through unit tests and simulation. 
Experimental results demonstrate consistent state-of-the-art performance across three world-model generation benchmarks of different types.
% , achieving 75.4 macro F1 score on Text2World, 91.74\% accuracy on CWMB, and substantial improvements on ByteSized32. 
By enabling fully autonomous world model generation without human feedback or manual annotations, this work opens new possibilities for AI systems that can reliably understand and formalize complex environments from natural language.

\section*{Ethics Statement}
All authors have read and will adhere to the ICLR Code of Ethics. This work does not involve human subjects, personal data, demographic attributes, or user studies; IRB approval was therefore not required. 
Our experiments use public, non-sensitive benchmarks: \emph{Text2World}, \emph{Code World Models Benchmark (CWMB)}, and \emph{ByteSized32}. We complied with dataset licenses and did not attempt to deanonymize or enrich any data with personal information. 
Because \ours uses web search as a tool (\S~\ref{sec:ks}), we enforced safeguards to reduce legal and research-integrity risks: we retrieved only publicly accessible pages, implemented a denylist to avoid solution leakage from benchmark source repositories or discussion pages as discussed in Section~\ref{sec:implementation}; 
We have no conflicts of interest or undisclosed sponsorship related to this work.

\section*{Reproducibility Statement}
Implementation details needed to re-create agents, tools, and evaluation are specified in Section~\ref{sec:implementation} and Appendix~\ref{app:agent_tool_config}; 
algorithmic workflow and role specialization are detailed in \S\ref{sec:wmg}–\S\ref{sec:edr} (with pseudo code in Alg.~\ref{alg:method}); prompts are provided in Appendix~\ref{app:prompt_examples}. 
For datasets and metrics, benchmark compositions and metric definitions are summarized in \S\ref{sec:benchmark_detail} and Appendix~\ref{app:metric_explanation}; ablation settings and additional figures/tables appear in Appendix~\ref{app:ablation}. 
As discussed in Section~\ref{sec:implementation}, to facilitate exact runs, we fix decoding parameters (temperature $=0$, top\_p $=1$) and cap agent turns, specify external services (web search API) and denylisted domains to prevent leakage, and report hardware assumptions (CPU-only). 
We also provide the source code of our methods in supplementary materials.

\clearpage

\bibliography{iclr2026_conference}
\bibliographystyle{iclr2026_conference}

% \newpage
\appendix
% \section{Appendix}
{\textbf{\LARGE Appendix}}

\section{The Use of Large Language Models}

We used a large language model (LLM) strictly as a general-purpose writing assistant for surface-level editing, such as grammar correction, wording polish, and minor style consistency. The LLM was not used for conceptual ideation, literature review, algorithm or model design, data collection or labeling, experiment setup, result analysis, or drafting of technical content. 
% All scientific claims, methods, and conclusions are solely authored and verified by the paper's authors. Consistent with the ICLR 2026 author guidelines, we acknowledge this limited usage and take full responsibility for the contents of the paper.%

\section{More Details on Methodology}

\subsection{Per-agent Tool Configuration}
\label{app:agent_tool_config}

\begin{table}[h]
\centering
\caption{Per-agent configuration.}
\label{tab:agent_runtime_cfg_full}
\setlength{\tabcolsep}{6pt}
\renewcommand{\arraystretch}{1.15}
\resizebox{0.7\textwidth}{!}{%
\begin{tabular}{c >{\centering\arraybackslash}p{0.6\textwidth}}
\toprule
\textbf{Agent} & \textbf{Tools} \\
\midrule
Deep Researcher   & browser\_search; browser\_open \\
Model Developer   & file\_tool; sandbox; run\_code \\
Simulation Tester & play\_env; file\_tool \\
Unit Tester       & run\_code; run\_bash; file\_tool \\
\bottomrule
\end{tabular}%
}
\end{table}

Detailed per-agent tool configuration is presented in Table~\ref{tab:agent_runtime_cfg_full}.

% \subsection{Pseudo Code of \multi}
\subsection{\texorpdfstring{Pseudo Code of \multi}{Pseudo Code of Agent2World Multi}}

We formalize the process of \multi in Algorithm~\ref{alg:method}.

\SetAlgoNlRelativeSize{0}
\SetKwProg{Proc}{\textbf{Procedure}}{}{}
\SetKwFunction{Parse}{ParseTaskDescribe}
\SetKwFunction{Research}{WebSearch}
\SetKwFunction{FileTool}{FileTool}
\SetKwFunction{CodeTool}{CodeTool}
\SetKwFunction{PlayEnv}{PlayEnv}

\SetKwFunction{DeepResearch}{ResearchAgent}
\SetKwFunction{Develop}{DevelopAgent}
\SetKwFunction{Unit}{UnitTestAgent}
\SetKwFunction{Sim}{SimulationTestAgent}
\SetKwFunction{Merge}{MergeFeedback}

\begin{algorithm}[h]
\caption{The execution pipeline of \multi}
\label{alg:method}
\KwIn{$T, N$}
\KwOut{$e$}

$N_r \gets \text{predefined integers}$\;

$R \gets \emptyset$  \;
$E \gets \emptyset$  \;
$Q \gets \text{ExtractQuestions}(T)$\;

\For{$r \gets 1$ \KwTo $N_r$}{
    $q \gets \DeepResearch(\text{select},\{Q, E, R\})$\;
    \If{$q = \emptyset$}{\textbf{break}}
    $L \gets $ \Research($q$)\;
    $E \gets E \cup \DeepResearch(\text{summarize},\{L\})$\;
    $R \gets \DeepResearch(\text{update},\{T,E,R\})$\;
}
$F_t \gets R$\;
$C_{\text{last}} \gets \emptyset$\;
\For{$n \gets 1$ \KwTo $N$}{
  $C_d \gets \Develop(T,F_t)$\;
  
  \If{$C_d \neq \emptyset $}{
  $C_{\text{last}} \gets C_d$\;
  $p_{\text{code}} \gets \FileTool(\text{save}, C_d)$\;
  }
  \Else{
    \textbf{continue}
  }
  
  $C_t \gets \Unit(C_d, T, R)$\;
  $p_{\text{test}} \gets \FileTool(\text{save}, C_t)$\;
  $U_t \gets \CodeTool(\text{run\_tests}, \{p_{\text{code}}, p_{\text{test}}\})$\;
  
  $S_t^\star \gets \PlayEnv(C_d)$\;
  $S_t \gets \Sim(S_t^\star, T)$\;
  
  \If{$U_t.\text{pass} \land S_t.\text{pass}$}{
    $e \gets C_{\text{last}}$\;
    \KwRet{$e$};
  }
  $F_t \gets $ \Merge{$U_t, S_t$}\;
}
$e \gets C_{\text{last}}$\;
\KwRet{$e$}
\end{algorithm}

\section{More Details on Benchmarks}
\label{sec:benchmark_detail}

\subsection{Side-by-side Comparison}

\begin{table}[h]
  \centering
  \caption{Overview of Text2World~\citep{hu2025text2world}, Code World Models Benchmark (CWMB)~\citep{dainese2024cwmb}, and ByteSized32~\citep{bytesized32}. ``Type'' denotes the target representation (PDDL vs.\ executable code). Metrics are shown at the family level. A detailed explanation of each metric is presented in Appendix~\ref{app:metric_explanation}.}
  \label{tab:benchmark_overview}
  \renewcommand{\arraystretch}{1.12}
  \setlength{\tabcolsep}{6pt}
  \resizebox{\textwidth}{!}{%
  \begin{tabular}{lccc}
    \toprule
    \textbf{Benchmark} & \textbf{\#Environments} & \textbf{Type} & \textbf{Metrics (core)} \\
    \midrule
    Text2World
      & 103
      & PDDL
      & Executability;\; Domain similarity;\;F1 scores \\
    % \midrule
    CWMB
      & 18
      & Code (Python)
      & Accuracy; Normalized return $\mathcal{R}$ (discrete/continuous) \\
    % \midrule
    ByteSized32
      & 32
      & Code (Python)
      & \makecell[c]{Technical validity;\;
        Specification compliance;\; \\
        Winnability;\;
        Physical reality alignment} \\
    \bottomrule
  \end{tabular}
  }
\end{table}

A side-by-side comparison of the evaluated benchmarks in this paper is presented in Table~\ref{tab:benchmark_overview}

\subsection{Metric Explanation}
\label{app:metric_explanation}

%================= Text2World =================
\paragraph{Text2World}
\begin{description}[leftmargin=1.5em,labelsep=0.5em]

  \item[\textbf{Executability}.]
  \emph{Name:} \textit{Exec}. \emph{Range:} $[0,1]$ (higher is better).
  Whether the generated \{domain, problem\} can be successfully parsed and validated by standard PDDL validators; reported as the fraction (percentage) over all test cases.
  Fine-grained metrics below are computed only when \textit{Exec}$=1$.

  \item[\textbf{Domain similarity}.]
  \emph{Name:} \textit{Sim}. \emph{Range:} $[0,1]$ (higher is better).
  Textual/structural similarity between the generated and gold PDDL measured by a \emph{normalized Levenshtein ratio}.
  
  Let $X$ and $Y$ be the character sequences of the two files with lengths $|X|$ and $|Y|$, and let $\mathrm{Lev}(X,Y)$ denote their Levenshtein distance, then
  \begin{equation}\label{eq:sim-def}
    \mathrm{Sim}(X,Y) \;=\; 1 - \frac{\mathrm{Lev}(X,Y)}{\max\{|X|,\,|Y|\}} \;\in\; [0,1].
  \end{equation}
  
  \item[\textbf{F1 scores}.]
  \emph{Range:} $[0,1]$ (higher is better).
  When \textit{Exec}$=1$, we parse both generated and gold PDDL into structured representations and report \emph{macro-averaged} F1 for the following components:
  \textbf{Predicates} ($\mathrm{F1}_{\text{PRED}}$),
  \textbf{Parameters} ($\mathrm{F1}_{\text{PARAM}}$),
  \textbf{Preconditions} ($\mathrm{F1}_{\text{PRECOND}}$), and
  \textbf{Effects} ($\mathrm{F1}_{\text{EFF}}$).We use the standard definition of $F_1$, where $\mathrm{P}$ and $\mathrm{R}$ denote precision and recall, respectively:
\[
  F_1 \;=\; \frac{2\,\mathrm{P}\,\mathrm{R}}{\mathrm{P}+\mathrm{R}}
\]
\end{description}

%================= CWMB=================
\paragraph{CWMB}
\begin{description}[leftmargin=1.5em,labelsep=0.5em]

  \item[\textbf{Prediction Accuracy}.]
  \emph{Symbol:} $\mathrm{Acc}_{\text{pred}}$. \emph{Range:} $[0,1]$ (higher is better).
  \emph{Definition:} We use the same accuracy metric as in the evaluation phase of GIF--MCTS (Sec.~4).
  Given a validation set $D=\{(s_i,a_i,r_i,s'_i,d_i)\}_{i=1}^N$ and CWM predictions
  $(\hat s'_i,\hat r_i,\hat d_i)=\texttt{CWM.step}(s_i,a_i)$, the accuracy uniformly weights next state,
  reward, and termination:
  \begin{equation}
    \mathrm{Acc}_{\text{pred}}
    = \frac{1}{N}\sum_{i=1}^{N}\!\left[
      \tfrac{1}{3}\,\mathbf{1}(\hat s'_i = s'_i)
      + \tfrac{1}{3}\,\mathbf{1}(\hat r_i = r_i)
      + \tfrac{1}{3}\,\mathbf{1}(\hat d_i = d_i)
    \right].
  \end{equation}

  \item[\textbf{Normalized Return}.]
  \emph{Symbol:} $\mathcal{R}$. \emph{Range:} unbounded (higher is better; $\mathcal{R}\!>\!0$ means better than random; $\mathcal{R}\!\to\!1$ approaches the oracle).
  \emph{Definition:}
  \begin{equation}\label{eq:norm-ret-cwmb}
    \mathcal{R}
    = \frac{R(\pi_{\mathrm{CWM}}) - R(\pi_{\mathrm{rand}})}
           {R(\pi_{\mathrm{true}}) - R(\pi_{\mathrm{rand}})} ,
  \end{equation}
  where $R(\pi)$ denotes the return.
  \emph{Protocol:} as in the original setup, we use vanilla MCTS for discrete action spaces and CEM for continuous action spaces; $R(\cdot)$ is averaged across a fixed number of episodes per environment (10 in the original), and $R(\pi_{\mathrm{rand}})$ uses the environment’s random policy baseline.

\end{description}

%================= ByteSized32 =================
\paragraph{ByteSized32}
\begin{description}[leftmargin=1.5em,labelsep=0.5em]

  \item[\textbf{Technical Validity}.]
  Range: $[0,1]$. Measured in the order of API calls, such that failure of an earlier function implies failure of subsequent tests. \texttt{Game initialization} is evaluated once at the beginning of the game, whereas \texttt{GENERATEPOSSIBLEACTIONS()} and \texttt{STEP()} are evaluated at \emph{every step}. We check:
  \begin{itemize}
    \item \textit{Game initialization}: the game/world initializes without errors;
    \item \textit{Valid actions generation}: the routine that enumerates valid actions for the current state returns without errors (verified via a bounded path crawl);
    \item \textit{Runnable game}: a bounded-depth crawl of trajectories executes without errors.
  \end{itemize}

  \item[\textbf{Specification Compliance}.]
  Range: $[0,1]$. An LLM acts as the judge for \emph{true/false} compliance against the task specification. The prompt provides the task spec \texttt{\{GAME\_SPEC\}}, the game code \texttt{\{GAME\_CODE\}}, and an evaluation question \texttt{\{EVAL\_QUESTION\}}; the LLM is instructed to first output \texttt{Yes/No} and then a brief rationale. To reduce variance, we use a fixed prompt template and perform multiple independent runs with majority vote/mean aggregation. We report three submeasures: \textit{Task-critical objects}, \textit{Task-critical actions}, and \textit{Distractors}.

  \item[\textbf{Physical Reality Alignment}.]
  Range: $[0,1]$. Automatic evaluation proceeds in two stages:\\
  \emph{(1) Trajectory generation:} perform a breadth-first crawl using the action strings returned by \texttt{GENERATEPOSSIBLEACTIONS()} at each step; actions are grouped by verb (first token) and expanded in a bounded manner. If an error occurs, the error message is recorded as the observation, and the search continues.\\
  \emph{(2) Sampling and judgment:} group paths by the last action verb, draw a fixed-size subsample approximately balanced across groups, and submit each path—together with the task description \texttt{\{GAME\_TASK\}}—to an LLM for a binary judgment (\texttt{yes/no}; errors are treated as failures). The final score is the fraction judged aligned.

  \item[\textbf{Winnability}.]
  Range: $[0,1]$. A text-game agent (LLM agent) attempts to reach a terminal \texttt{win} within horizon $H$; we report the fraction of tasks deemed winnable. Given the limited agreement between automatic and human assessments for this metric, we prioritize human evaluation in the main results and use the automatic estimate asan  auxiliary reference.
\end{description}

% Optional breakdown: we additionally report submetrics such as Game Init., Possible Actions, Runnable Game, Critical Objects/Actions, and Distractors (see Table~X).

% original cwmb
% In this setting, we are interested in both the accuracy of the generated CWM, as
% well as its performance when actually employed by a planning algorithm. We use as accuracy the
% same metric used in the evaluation phase of GIF-MCTS (Section 4). To measure the performance of
% planning with the CWM, we define the normalized return R of a CWM as:
% R(CWM) = R(πCWM) − R(πrand)
% R(πtrue) − R(πrand)
% , (3)
% where R(πCWM) represents the return obtained when using the CWM as the internal model for the
% planner, R(πtrue) is the return gathered with the true environment as the model while using the same
% planner (oracle planner), and R(πrand) is the return from a random policy. This metric is positive
% when the performance of the CWM planner is above that of a random policy and reaches one when
% the return approaches the value from the oracle planner. We report results for the CWMB in Table 2.
% As the planner, we use a vanilla MCTS implementation for the environments with discrete actions
% and a Cross Entropy Method (CEM) planner for the ones with continuous action
% spaces (full details of the two planning algorithms are reported in Appendix L).

\section{More Details on Ablation Study}
\label{app:ablation}

% \begin{table}[!htbp]
% \centering
% \caption{Ablation Study of \ours on CWMB~\citep{dainese2024cwmb}.}
% \label{tab:gif_mcts_ablation}
% \resizebox{\textwidth}{!}{%
% \begin{tabular}{l cc cc cc}
% \toprule
%     \multirow{2}{*}{\textbf{Method}} &
%     \multicolumn{2}{c}{\textbf{Discrete Action Space}} &
%     \multicolumn{2}{c}{\textbf{Continuous Action Space}} &
%     \multicolumn{2}{c}{\textbf{Overall}} \\
%     \cmidrule(lr){2-3}\cmidrule(lr){4-5}\cmidrule(lr){6-7}
%      & \multicolumn{1}{c}{Accuracy (\(\uparrow\))} &
%      \multicolumn{1}{c}{\(\mathcal{R}\) (\(\uparrow\))} &
%      \multicolumn{1}{c}{Accuracy (\(\uparrow\))} &
%      \multicolumn{1}{c}{\(\mathcal{R}\) (\(\uparrow\))} &
%      \multicolumn{1}{c}{Accuracy (\(\uparrow\))} &
%      \multicolumn{1}{c}{\(\mathcal{R}\) (\(\uparrow\))} \\
% \midrule
% \ours & \textbf{0.9174} & \textbf{0.8333} & \textbf{0.3575} & \textbf{0.3050} & \textbf{0.5441} & \textbf{0.4811}\\
% \midrule
% No \emph{Deep Researcher} & 0.8794 & 0.4407 & 0.3404 & 0.2201 & 0.5201 & 0.2936 \\
% No \emph{Simulation Tester} & 0.8920 & 0.5718 & 0.3288 & 0.1577 & 0.5275 & 0.3039 \\
% No \emph{Unit Tester} & 0.6166 & 0.3863 & 0.3025 & 0.1704 & 0.4072 & 0.2423 \\
% \bottomrule
% \end{tabular}%
% }
% \end{table}

\begin{table}[!htbp]
\centering
\caption{Ablation Study of \ours on CWMB~\citep{dainese2024cwmb}.}
\label{tab:gif_mcts_ablation}
\resizebox{\textwidth}{!}{%
\begin{tabular}{l cc cc cc}
\toprule
    \multirow{2}{*}{\textbf{Method}} &
    \multicolumn{2}{c}{\textbf{Discrete Action Space}} &
    \multicolumn{2}{c}{\textbf{Continuous Action Space}} &
    \multicolumn{2}{c}{\textbf{Overall}} \\
    \cmidrule(lr){2-3}\cmidrule(lr){4-5}\cmidrule(lr){6-7}
     & \multicolumn{1}{c}{Accuracy (\(\uparrow\))} &
     \multicolumn{1}{c}{\(\mathcal{R}\) (\(\uparrow\))} &
     \multicolumn{1}{c}{Accuracy (\(\uparrow\))} &
     \multicolumn{1}{c}{\(\mathcal{R}\) (\(\uparrow\))} &
     \multicolumn{1}{c}{Accuracy (\(\uparrow\))} &
     \multicolumn{1}{c}{\(\mathcal{R}\) (\(\uparrow\))} \\
\midrule
\ours & \textbf{0.9174} & \textbf{0.8333} & \textbf{0.3575} & \textbf{0.3050} & \textbf{0.5441} & \textbf{0.4811}\\
\midrule
No \emph{Deep Researcher} & 0.8794\footnotesize{$_{\color{red}-0.0380}$} & 0.4407\footnotesize{$_{\color{red}-0.3926}$} & 0.3404\footnotesize{$_{\color{red}-0.0171}$} & 0.2201\footnotesize{$_{\color{red}-0.0849}$} & 0.5201\footnotesize{$_{\color{red}-0.0240}$} & 0.2936\footnotesize{$_{\color{red}-0.1875}$} \\
No \emph{Simulation Tester} & 0.8920\footnotesize{$_{\color{red}-0.0254}$} & 0.5718\footnotesize{$_{\color{red}-0.2615}$} & 0.3288\footnotesize{$_{\color{red}-0.0287}$} & 0.1577\footnotesize{$_{\color{red}-0.1473}$} & 0.5275\footnotesize{$_{\color{red}-0.0166}$} & 0.3039\footnotesize{$_{\color{red}-0.1772}$} \\
No \emph{Unit Tester} & 0.6166\footnotesize{$_{\color{red}-0.3008}$} & 0.3863\footnotesize{$_{\color{red}-0.4470}$} & 0.3025\footnotesize{$_{\color{red}-0.0550}$} & 0.1704\footnotesize{$_{\color{red}-0.1346}$} & 0.4072\footnotesize{$_{\color{red}-0.1369}$} & 0.2423\footnotesize{$_{\color{red}-0.2388}$} \\
\bottomrule
\end{tabular}%
}
\end{table}

Detailed experimental results of the ablation study are presented in Table~\ref{tab:gif_mcts_ablation}.

\section{More Details on Data Construction}
\label{app:train_data_construction}

\begin{table}[htbp]
\centering
\footnotesize
\caption{Train Data Before and After Filtering}
\label{tab:train_data}
\setlength{\tabcolsep}{6pt}
\renewcommand{\arraystretch}{1.15}
\begin{revblock}
\begin{tabular}{l|cccc}
\toprule
\textbf{Status} & \textbf{CWMB} & \textbf{ByteSized32} & \textbf{PDDL} & \textbf{MCP} \\
\midrule
Unfiltered & 500 & 500 & 600 & 800 \\
Filtered   & 266 & 277 & 483 & 500 \\
\bottomrule
\end{tabular}
\end{revblock}
\end{table}

\section{More Details on Error Analysis}
\label{app:detailed_error_analysis}

\subsection{Error Analysis on Text2World and ByteSized32}

\begin{figure}[!htbp]
\centering
\includegraphics[width=0.99\textwidth]{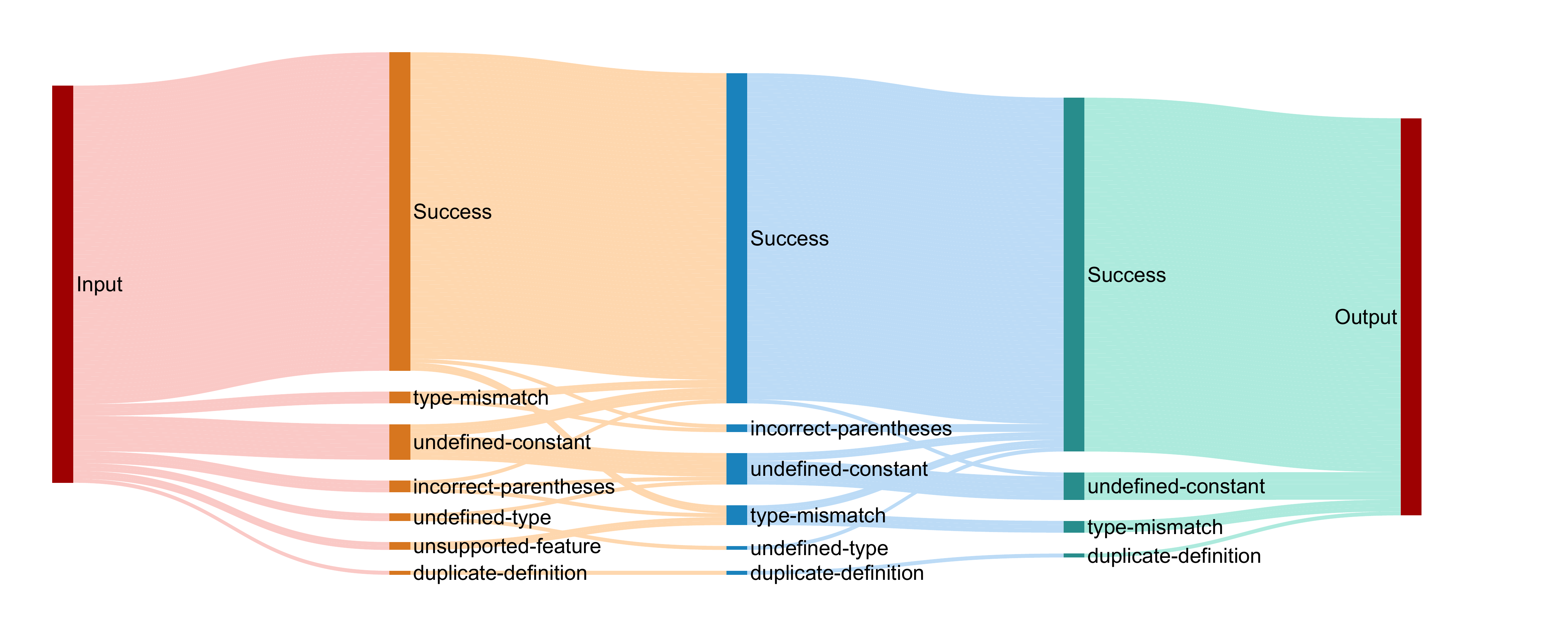}
\caption{Error distribution of \multi on Text2World.}
\label{fig:sankey_diagram_text2world}
\end{figure}

\begin{figure}[!htbp]
\centering
\includegraphics[width=0.99\textwidth]{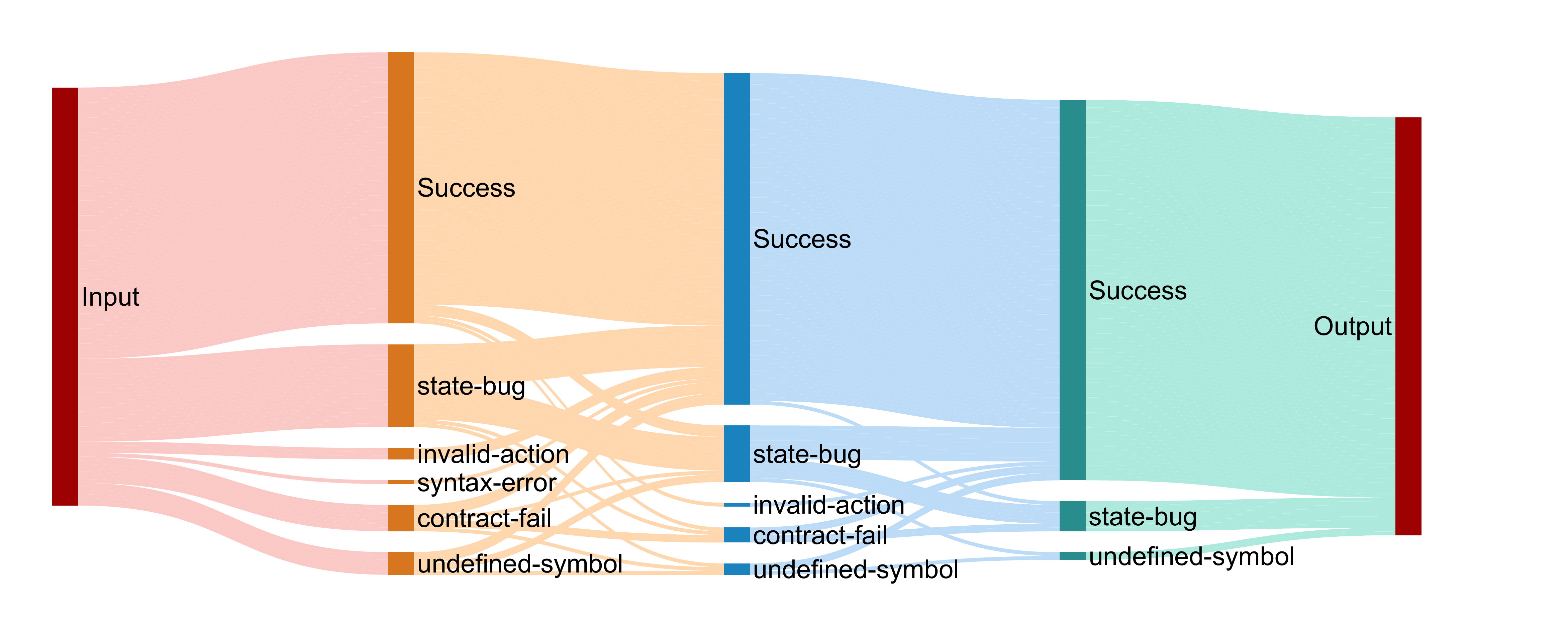}
\caption{Error distribution of \multi on ByteSized32.}
\label{fig:sankey_diagram_Bytesized32}
\end{figure}

We visualize the error patterns during evaluation-driven refinement on Text2World and ByteSized32 in Figure~\ref{fig:sankey_diagram_text2world} and Figure~\ref{fig:sankey_diagram_Bytesized32}.

\subsection{Distribution of Error Types}

A detailed proportion of error types on Text2World, CWMB, ByteSized32 are presented in Table~\ref{tab:text2world_error_by_turn}, Table~\ref{tab:cwmb_error_by_turn} and Table~\ref{tab:bytesized32_error_by_turn}, respectively.

\begin{table}[htbp]
\centering
\footnotesize
\caption{Distribution of Syntax Errors in Text2World Across Turns}
\label{tab:text2world_error_by_turn}
\setlength{\tabcolsep}{3pt}       
\renewcommand{\arraystretch}{1.1} 
\begin{tabularx}{\textwidth}{@{}l|X|c|c|c@{}}
\toprule
\textbf{Error Type} & \textbf{Explanation} &
\makecell{\textbf{Turn 0}\\(\%)} &
\makecell{\textbf{Turn 1}\\(\%)} &
\makecell{\textbf{Turn 2}\\(\%)} \\
\midrule
undefined-constant     & Reference to undeclared constants in predicates or actions. & 8.91 & 7.92 & 6.93 \\
type-mismatch          & Parameter type conflict with declared type constraints.     & 2.97 & 4.95 & 2.97 \\
incorrect-parentheses  & Invalid or mismatched parentheses.                          & 2.97 & 1.98 & 0.00 \\
undefined-type         & Undeclared parent type in hierarchical type definitions.    & 1.98 & 0.99 & 0.00 \\
unsupported-feature    & Parser-incompatible features (e.g.,either types).           & 1.98 & 0.00 & 0.00 \\
duplicate-definition   & Multiple declarations of identical domain elements.         & 0.99 & 0.99 & 0.99 \\
\bottomrule
\end{tabularx}
\end{table}

\begin{table}[htbp]
\centering
\footnotesize
\caption{Distribution of Syntax Errors in CWMB Across Turns}
\label{tab:cwmb_error_by_turn}
\setlength{\tabcolsep}{3pt}        
\renewcommand{\arraystretch}{1.1}  
\begin{tabularx}{\textwidth}{@{}l|X|c|c|c|c@{}} 
\toprule
\textbf{Error Type} & \textbf{Explanation} &
\makecell{\textbf{Turn 0}\\(\%)} &
\makecell{\textbf{Turn 1}\\(\%)} &
\makecell{\textbf{Turn 2}\\(\%)} &
\makecell{\textbf{Turn 3}\\(\%)} \\
\midrule
signature-mismatch   & Arity/types do not match the declared signature. & 27.78 & 11.11 & 11.11 & 5.56 \\
schema-mismatch      & Value type/shape/dtype violates the expected schema.                      & 22.22 & 5.56  & 0.00  & 0.00 \\
dynamics-error       & State or reward deviates from expected dynamics.                 & 11.11 & 44.44 & 33.33 & 11.11 \\
non-deterministic    & Results are inconsistent under fixed conditions.                  & 11.11 & 0.00  & 0.00  & 0.00 \\
judgment-bug         & Environment setup inconsistent with the description.              & 11.11 & 5.56  & 11.11 & 5.56 \\
invariant-violation  & Internal invariants are broken (e.g., illegal config). & 0.00  & 5.56  & 0.00  & 5.56 \\

\bottomrule
\end{tabularx}
\end{table}

\begin{table}[htbp]
\centering
\footnotesize
\caption{Distribution of Syntax Errors in ByteSized32 Across Turns}
\label{tab:bytesized32_error_by_turn}
\setlength{\tabcolsep}{3pt}        
\renewcommand{\arraystretch}{1.1}  
\begin{tabularx}{\textwidth}{@{}l|X|c|c|c@{}} 
\toprule
\textbf{Error Type} & \textbf{Explanation} &
\makecell{\textbf{Turn 0}\\(\%)} &
\makecell{\textbf{Turn 1}\\(\%)} &
\makecell{\textbf{Turn 2}\\(\%)} \\
\midrule
state-bug        & State inconsistency across steps.                                     & 19.82 & 13.51 & 7.21 \\
contract-fail    & Task/API contract not satisfied.                                      & 6.31  & 3.60  & 0.00 \\
undefined-symbol & Reference to undeclared name/type/domain/constant.                    & 5.41  & 2.70  & 1.80 \\
invalid-action   & Unknown or unsupported action not safely handled.                     & 2.70  & 0.90  & 0.00 \\
syntax-error     & Load/parse failure (e.g., null bytes, syntax/indentation errors).     & 0.90  & 0.00  & 0.00 \\

\bottomrule
\end{tabularx}
\end{table}

\subsection{\revinline{Error Distribution of Baselines and Comparison}}

\begin{table}[htbp]
\centering
\footnotesize
\caption{\revinline{Distribution of Syntax Errors in CWMB Across Methods and Turns}}
\label{tab:cwmb_error_by_method}
\setlength{\tabcolsep}{3pt}
\renewcommand{\arraystretch}{1.1}
\begin{revblock}
\begin{tabular}{l|*{6}{c}}
\toprule
\textbf{Method} &
\makecell{\textbf{signature-}\\\textbf{mismatch}} &
\makecell{\textbf{schema-}\\\textbf{mismatch}} &
\makecell{\textbf{dynamics-}\\\textbf{error}} &
\makecell{\textbf{non-}\\\textbf{deterministic}} &
\makecell{\textbf{judgment-}\\\textbf{bug}} &
\makecell{\textbf{invariant-}\\\textbf{violation}} \\
\midrule
WorldCoder (Last Turn)      & 0 & 0 & 3 & 0 & 2 & 2 \\
GIF-MCTS (Last Turn)        & 0 & 0 & 2 & 0 & 3 & 2 \\
\multi (Turn 1) & 2 & 1 & 8 & 0 & 1 & 1 \\
\multi (Turn 2) & 2 & 0 & 6 & 0 & 2 & 0 \\
\multi (Turn 3) & 1 & 0 & 2 & 0 & 1 & 1 \\
\bottomrule
\end{tabular}
\end{revblock}
\end{table}

\begin{revblock}
As shown in Table~\ref{tab:cwmb_error_by_method},  both WorldCoder and GIF-MCTS still exhibit several
residual dynamics-related failures in their final outputs: although they no
longer trigger signature- or schema-mismatch errors, they retain multiple
\textit{dynamics-error} and \textit{judgment-bug} cases, together with
non-trivial \textit{invariant-violation} counts. This suggests that their
search procedures are effective at ironing out interface- and type-level
inconsistencies, but are less successful at fully aligning the learned
dynamics with the intended environment specification.

In contrast, \multi starts from a broader
spectrum of error types at turn~1 (including signature/schema mismatches and
a larger number of dynamics errors), and progressively suppresses them over
subsequent refinement rounds. By turn~3, schema mismatches and
nNon-deterministicbehaviors have disappeared, signature mismatches are
reduced to a single case, and dynamics errors, judgment bugs, and invariant
violations are all substantially lower than those of the baselines. This
trajectory is consistent with the qualitative analysis in Section~5.4: the
Testing Team first drives the Model Developer to resolve structural and
specification-level issues, and then focuses on subtle dynamics and invariant
failures, ultimately producing world models that are both syntactically robust
and semantically faithful to the target environments.
\end{revblock}

\section{Prompt Examples}
\label{app:prompt_examples}

\definecolor{PromptGrayBg}{HTML}{F9F9F9} 
\definecolor{PromptGrayFrame}{HTML}{404040} 
\definecolor{PromptText}{HTML}{111111}

\newtcolorbox{promptbox}[1][]{
  enhanced,
  breakable,
  drop shadow,
  width=\linewidth,
  boxrule=0.8pt,
  colback=PromptGrayBg,         % 背景（浅灰）
  colframe=PromptGrayFrame,     % 边框（深灰）
  colbacktitle=PromptGrayFrame, % 标题栏底色（深灰）
  coltitle=white,               % 标题文字：白色
  coltext=PromptText,           % 正文深灰
  left=8pt, right=8pt, top=4pt, bottom=4pt,
  fonttitle=\small,
  title={\bfseries #1},
}
%================= Deep Researcher =================
\begin{promptbox}[Deep Researcher]
You are a world-class Systems Analyst and Technical Specification Writer, specializing in creating reinforcement learning environments compliant with the Gymnasium API. Your mission is to transform an ambiguous task description into a precise, actionable, and verifiable technical specification.

<Environment Name> \\
\textbf{CliffWalking-v0} \\
</Environment Name>

\textbf{<TASK DESCRIPTION>} \\
Cliff walking involves crossing a gridworld from start to goal while avoiding falling off a cliff.

\#\# Description
...

\textbf{</TASK DESCRIPTION>}

\textbf{<Workflow>} \\
Please strictly follow the following six-step process:
\begin{itemize}
  \item \textbf{Deconstruction and Analysis (Use Version Locking)}
  \begin{itemize}
    \item Identify all ambiguities, gaps, and conflicts in the task description.
    \item Lock the exact environment version and all key library versions (record name, version, and source link).
    \item Categorize gaps by type: missing value/unit/range boundary/time-sensitive/ambiguous reference/unclosed list/conflict/no provenance.
  \end{itemize}

  \item \textbf{Planning and Investigation (Authoritative Search + Evidence Log)}
  \begin{itemize}
    \item For each high/mid-level project, write 1--2 focused queries that include: synonyms/abbreviations, site filters, authoritative domains (e.g., site:numpy.org, site:mujoco.readthedocs.io, site:doi.org), and recency windows (e.g., after 2024-01-01 or ``last 2 years'').
    \item Execute the query using browser\_search and open >= 2 trusted results with browser\_open.
    \item If the top source disagrees, open >= 1 additional authoritative sources and triangulate.
    \item Create an evidence log entry for each opened page: Title | Organization/Author | Version/Submission | URL (+ archived URL) | Publication Date | Access Date (Asia/Singapore) | 3 Key Facts | Confidence (High/Medium/Low).
  \end{itemize}

  \item \textbf{Synthesis and Citation (Conflict Resolution)}
  \begin{itemize}
    \item Integrate the findings into a concise evidence summary with citations.
    \item When sources conflict, explain the differences and justify the chosen resolution (related to version locking).
  \end{itemize}

  \item \textbf{Refinement and Improvement (Specification Patch)}
  \begin{itemize}
    \item Generate a structured ``diff'': action/observation space; rewards; termination/truncation; timing (dt/frame\_skip); seeding and certainty; numerical tolerances; dependencies; interface flags.
  \end{itemize}

  \item \textbf{Formalization and Finalization (Ready-to-Use Specification)}
  \begin{itemize}
    \item Write the final specification according to the <Output Format> , including the public API, core logic, usage scenarios, and a verification plan aligned with metrics and statistical validation.
  \end{itemize}

  \item \textbf{Review and Self-Correction (Compliance Check)}
  \begin{itemize}
    \item Verify conformance to the <Output Constraints> (OUTPUT\_CONSTRAINTS>), version consistency, 
    SI units, ISO dates, and the inclusion of any code.
  \end{itemize}
\end{itemize}
\textbf{</Workflow>}

\textbf{<OUTPUT\_CONSTRAINTS>}
\begin{itemize}
  \item Strictly adhere to the structure defined in \texttt{<PLANNING\_STRUCTURE>}.
  \item Do \textbf{NOT} output runnable code definitions (classes, functions). Only may include short illustrative snippets or pseudo-code.
  \item All claims about industry standards or common practices MUST be supported by citations.
  \item Use ISO-8601 dates (e.g., 2025-09-02).
  \item Use SI units for physical and mathematical quantities.
  \item Data-leakage rule: Do not access, copy, quote, or derive from raw source code in the OpenAI/Gym/Gymnasium repositories or similar code repositories. Do not include any repository code in the output. Prefer official documentation, standards, papers, or reputable secondary sources. If the only available evidence is a code repository, summarize behavior without copying code and mark it as an inference with risks.
\end{itemize}

\textbf{<PLANNING\_STRUCTURE>}
\begin{itemize}
  \item Your output must begin with this planning and analysis section.
  \item \textbf{Ambiguity Analysis}
    \begin{itemize}
      \item List each ambiguity/vagueness/conflict and mark Impact: High / Medium / Low.
      \item Cover at least: missing numeric value, missing unit, missing boundary/range, time-sensitive items, unclear references, open lists (``etc.''/``e.g.''), conflicts, and missing citation.
    \end{itemize}
  \item \textbf{Investigation Plan}
    \begin{itemize}
      \item For each High/Medium item, provide one atomic question.
      \item For each question, provide 1--2 executable queries including: synonyms/abbreviations, a site filter to authoritative domains, and a time window (e.g., \texttt{after:2024-01-01} or ``past 2 years'').
      \item State the minimum evidence policy: High/Medium $\rightarrow$ $\ge{}2$ credible sources; if disagreement $\rightarrow$ add $\ge{}1$ more for triangulation.
    \end{itemize}
\end{itemize}

\textbf{<Formula requirements>}
\begin{itemize}
  \item For any formula, define all symbols, units, and applicability constraints.
  \item Cite the source of the formula immediately after its definition.
  \item Provide the complete formula rather than a descriptive explanation.
\end{itemize}

\textbf{<OUTPUT\_FORMAT>}
\begin{itemize}
  \item Please provide the final specification document structured as follows. This is the primary deliverable. Do \textbf{NOT} include code.
  \item <Version \& Provenance>
  \item <Evidence Summary>
  \item <Spec Patch>
  \item <Theoretical Foundations>
  \item <Final Specification>
  \item <Assumptions \& Risks>
  \item <Third-Party Library Usage>
\end{itemize}

\end{promptbox}

%================= Model Developer =================

\begin{promptbox}[Model Developer]

\textbf{<TASK DESCRIPTION>} \\
Cliff walking involves crossing a gridworld from start to goal while avoiding falling off a cliff.

\#\# Description
...

\textbf{</TASK DESCRIPTION>}

\textbf{<Research Report>} \\
\# CliffWalking-v0 Environment Specification

\noindent \textbf{Version \& Provenance}
\begin{itemize}
  \item Environment: CliffWalking-v0
  \item Gymnasium version: v0.26.3 (2022-09-15) and latest documentation snapshot (2025-01-01)
  \item Scope: This specification normalizes the CliffWalking environment as implemented in Gymnasium v0.26.3, aligning with the classic gridworld cliff walking task.
  \item Accessed date: 2024-06-01 (Asia/Singapore)
\end{itemize}

\noindent \textbf{Evidence Summary}
\begin{itemize}
  \item The environment is a 4x12 grid with 48 discrete states indexed by row-major flattening: state = row * 12 + col.
  \item Valid states exclude cliff cells ([3,1..10]) and the goal ([3,11]). The player can occupy all cells in the first 3 rows plus the bottom-left cell [3,0].
  \item The start state is 36 (row=3, col=0), and the goal state is 47 (row=3, col=11).
  \item The action space is Discrete(4) with actions: 0 (up), 1 (right), 2 (down), 3 (left).
  \item Each step yields a reward of -1. Stepping into the cliff yields a reward of -100 and resets the player to the start state; the episode continues.
  \item The episode terminates only when the player reaches the goal state.
  \item The set\_state method is not explicitly documented but is inferred to accept any valid non-terminal, non-cliff state and sets the environment to not done.
  \item No explicit error handling for invalid actions is documented; valid inputs are assumed.
  \item The environment is deterministic.
\end{itemize}

\noindent \textbf{Spec Patch}
\begin{itemize}
  \item action.space = Discrete(4), values \{0,1,2,3\}, shape (1,)
  \item observation.space = Discrete(48), integer in [0,47], representing flattened grid index: row * 12 + col
  \item valid states = all grid cells except cliff cells ([3,1..10]) and goal ([3,11])
  \item start\_state = 36 (row=3, col=0)
  \item goal\_state = 47 (row=3, col=11)
  \item reward.step = -1 per step
  \item reward.cliff = -100 on stepping into cliff
  \item episode termination = True if state == goal\_state; else False
  \item stepping into cliff resets player to start\_state, episode continues
  \item set\_state(state: int) sets environment state to given valid state, done = False
  \item no maximum episode length or truncation condition
  \item no explicit error handling for invalid actions; assume valid input
  \item environment deterministic transitions
\end{itemize}

\noindent \textbf{Theoretical Foundations}
\begin{itemize}
  \item \textbf{MDP Definition:}
    \begin{itemize}
      \item State space $S = \{0,\ldots,47\} \setminus \{\text{cliff states}, \text{goal state}\}$.
      \item Action space $A = \{0,1,2,3\}$.
      \item Transition function $T(s,a,s')$ (deterministic):
        \begin{itemize}
          \item If $s'$ is a cliff cell, next state $=$ start\_state.
          \item Else next state $= s + \delta(a)$ (with grid boundary checks).
        \end{itemize}
      \item Reward function $R(s,a,s')$:
        \begin{itemize}
          \item $R = -100$ if $s'$ is a cliff cell.
          \item $R = -1$ otherwise.
        \end{itemize}
      \item Episode ends when $s' =$ goal\_state.
    \end{itemize}
  \item \textbf{Symbol Table:}
    \begin{itemize}
      \item $s$: state (int), flattened grid index, $0 \le s \le 47$, $s$ not in cliff or goal.
      \item $a$: action (int), in $\{0:\ \text{up},\ 1:\ \text{right},\ 2:\ \text{down},\ 3:\ \text{left}\}$.
      \item $r$: reward (float), $-100$ or $-1$.
      \item done: boolean, True if $s =$ goal\_state.
      \item start\_state: 36 (int).
      \item goal\_state: 47 (int).
      \item cliff\_cells: set of ints corresponding to grid positions $[3,1..10]$.
      \item nrows: 4 (int).
      \item ncols: 12 (int).
    \end{itemize}
  \item \textbf{Assumptions:}
    \begin{itemize}
      \item No shaping rewards.
      \item Deterministic environment.
      \item No truncation or max step limit.
    \end{itemize}
\end{itemize}

\noindent \textbf{Final Specification}
\begin{itemize}
  \item \textbf{Environment Name}
    \begin{itemize}
      \item CliffWalking-v0
    \end{itemize}
  \item \textbf{Public API}
    \begin{itemize}
      \item \_\_init\_\_()
      \item set\_state(state: int)
      \item step(action: int) $\rightarrow$ (observation: int, reward: float, done: bool)
    \end{itemize}
  \item \textbf{Core Logic Description}
    \begin{itemize}
      \item \_\_init\_\_:
        \begin{itemize}
          \item Initialize grid size (4 rows $\times$ 12 columns).
          \item Define start\_state = 36 (row=3, col=0).
          \item Define goal\_state = 47 (row=3, col=11).
          \item Define cliff cells as positions $[3,1..10]$.
          \item Initialize current state to start\_state.
          \item Initialize done flag to False.
        \end{itemize}
      \item set\_state(state: int):
        \begin{itemize}
          \item Validate that state is a valid observation (not cliff or goal).
          \item Set current state to state.
          \item Set done flag to False.
        \end{itemize}
      \item step(action: int) $\rightarrow$ (obs, reward, done):
        \begin{itemize}
          \item Validate action is in $\{0,1,2,3\}$.
          \item Compute next position based on action with boundary checks:
            \begin{itemize}
              \item 0: move up (row $-1$)
              \item 1: move right (col $+1$)
              \item 2: move down (row $+1$)
              \item 3: move left (col $-1$)
            \end{itemize}
          \item If next position is outside grid, remain in current position.
          \item If next position is a cliff cell:
            \begin{itemize}
              \item reward $= -100$
              \item next state $=$ start\_state
              \item done $=$ False
            \end{itemize}
          \item Else if next position is goal\_state:
            \begin{itemize}
              \item reward $= -1$
              \item done $=$ True
              \item next state $=$ goal\_state
            \end{itemize}
          \item Else:
            \begin{itemize}
              \item reward $= -1$
              \item done $=$ False
              \item next state $=$ computed position
            \end{itemize}
          \item Update current state to next state and return (next state, reward, done).
        \end{itemize}
    \end{itemize}
  \item \textbf{Usage Scenarios}
    \begin{itemize}
      \item \textbf{Scenario 1: Normal move into cliff}
        \begin{itemize}
          \item Initial state: 36 ([3,0])
          \item Action: 1 (right)
          \item Next state: 37 ([3,1]) is cliff $\rightarrow$ reset to 36
          \item Reward: $-100$
          \item done: False
        \end{itemize}
      \item \textbf{Scenario 2: Move to goal}
        \begin{itemize}
          \item Current state: 46 ([3,10])
          \item Action: 1 (right)
          \item Next state: 47 ([3,11]) goal
          \item Reward: $-1$
          \item done: True
        \end{itemize}
      \item \textbf{Scenario 3: Move up from start}
        \begin{itemize}
          \item Current state: 36 ([3,0])
          \item Action: 0 (up)
          \item Next state: 24 ([2,0])
          \item Reward: $-1$
          \item done: False
        \end{itemize}
    \end{itemize}
\end{itemize}

\noindent \textbf{Assumptions \& Risks}
\begin{itemize}
  \item No truncation or maximum episode length specified; risk of infinite episodes if goal not reached.
  \item set\_state accepts only valid non-terminal, non-cliff states; no explicit error handling for invalid states.
  \item No explicit error handling for invalid actions; assumed caller responsibility.
  \item Reward and reset behavior on cliff stepping inferred from documentation; no episode termination on cliff.
  \item Observation space includes all grid cells but player never occupies cliff or goal states.
  \item Potential future Gymnasium version changes may alter behavior.
\end{itemize}

\noindent \textbf{Third-Party Library Usage}
\begin{itemize}
  \item Gymnasium API conventions inform environment interface: discrete action and observation spaces, step and reset semantics.
  \item No new runtime dependencies introduced.
  \item RNG seeding not specified; environment deterministic.
  \item No source code copied; specification based on official Gymnasium documentation and Sutton \& Barto RL book.
  \item Numpy indexing conventions used for grid flattening.
\end{itemize}

\textbf{</Research Report>} \\

\textbf{<Workflow>} \\
1. \textbf{Deconstruct Specification:} Carefully review the <Task Describe>,<Research Report>{} to fully understand the environment's specification, including state/action spaces, dynamics, reward function, and termination conditions. \\
2. \textbf{Physics Engine Selection:} Evaluate if the task requires physics simulation. If so, choose an appropriate physics engine for the specific task requirements. \\
3. \textbf{Model Design:} If using a physics engine, design the model structure and embed it as needed in the Python file. \\
4. \textbf{Plan Class Structure:} Outline the `Environment` class, including its internal state variables, helper methods, and the public interface (`\_\_init\_\_`, `reset`, `set\_state`, `step`). \\ 
5. \textbf{Implement Complete Code:} Write the full implementation of the `Environment` class.
6. \textbf{Self-Correction Review:} Meticulously check that the generated code fully complies with the <TASK DESCRIPTION>{}, the <Research Report>{}, and all <ImplementationRequirements>{}. \\
7. \textbf{Finalize Output:} Present the complete, reviewed, and runnable single-file code in the specified final format. \\
\textbf{</Workflow>} \\

\noindent \textbf{<ImplementationRequirements>}

\begin{enumerate}
  \item \textbf{Interface (single file):}
    \begin{itemize}
      \item Implement a complete, self-contained Python class Environment with:
        \begin{itemize}
          \item \_\_init\_\_(self, seed: int $\mid$ None = None)
          \item reset(self, seed: int $\mid$ None = None) $\rightarrow$ ndarray \ (reinitialize the episode and return the initial observation in canonical shape)
          \item set\_state(self, state) \ (must accept ndarray \textit{or} list/tuple in canonical shape)
          \item step(self, action) $\rightarrow$ tuple[ndarray, float, bool] \ (returns: observation, reward, done)
        \end{itemize}
      \item Requirements:
        \begin{itemize}
          \item Single-file constraint: all code, including any model definitions, must be contained in one Python file.
          \item For physics-based environments, embed model definitions as string constants within the class.
          \item Explicitly define state, action, and observation spaces (types, shapes, ranges, formats).
          \item Provide reproducibility (seeding) via the constructor and/or a seed(int) method.
          \item Be robust to common representations:
            \begin{itemize}
              \item set\_state: accept list/tuple/ndarray of the same logical content.
              \item step: accept int / NumPy integer scalar / 0-D or 1-D len-1 ndarray (convert to canonical form; raise clear TypeError/ValueError on invalid inputs).
            \end{itemize}
          \item No dependence on external RL frameworks; no Gym inheritance.
          \item No external file dependencies (model definitions must be embedded).
          \item Maintain internal state consistency; allow reconstruction from observations where applicable.
          \item Clean, readable code suitable for RL experimentation.
        \end{itemize}
    \end{itemize}

  \item \textbf{Determinism \& validation:}
    \begin{itemize}
      \item Provide reproducibility via seed (constructor and/or seed(int) method).
      \item Normalize inputs: accept equivalent representations (e.g., NumPy scalar/int/len-1 array) and convert to a canonical form.
      \item Validate inputs; raise clear one-line errors (ValueError/TypeError) on invalid shapes or ranges.
    \end{itemize}

  \item \textbf{Dynamics (MCTS/control oriented):}
    \begin{itemize}
      \item For physics-based tasks, prefer suitable physics simulation methods with embedded model definitions over custom physics implementations.
      \item Choose and document an integration scheme (e.g., implicit integrator, explicit Euler) consistent with the research report.
      \item Use a stable time step $dt$; clamp to safety bounds; keep all values finite (no NaN/Inf).
      \item Keep per-step computation efficient and allocation-light.
    \end{itemize}

  \item \textbf{Dependencies \& style:}
    \begin{itemize}
      \item No Gym inheritance or external RL frameworks unless explicitly allowed.
      \item Allowed: third-party libraries as needed (e.g., NumPy, physics engines, SciPy, Numba, JAX, PyTorch, etc.).
      \item For robotics/physics tasks, physics engines with embedded model definitions are recommended over custom implementations.
      \item Clean, readable code suitable for RL experimentation.
      \item All dependencies must be importable standard libraries or commonly available packages.
    \end{itemize}
\end{enumerate}

\noindent \textbf{</ImplementationRequirements>}

\textbf{<Output Format>}

\textbf{<final>}
\textbf{<code\_file\_path>}
The entrypoint file path of the generated code.
\textbf{</code\_file\_path>}
\textbf{<entrypoint\_code>}
```python
\# Your complete, runnable single-file implementation here.
```
\textbf{</entrypoint\_code>}
\textbf{</final>}

\textbf{</Output Format>}

\end{promptbox}

%================= Unit Tester =================

\begin{promptbox}[Unit Tester]

\textbf{<TASK DESCRIPTION>}
Cliff walking involves crossing a gridworld from start to goal while avoiding falling off a cliff.

\#\# Description
...

\textbf{</TASK DESCRIPTION>}

\textbf{<CodeArtifact path="environment.py">}
\texttt{\{code\}}
\textbf{</CodeArtifact>}

\textbf{<ExecutionPolicy>}
\begin{itemize}
  \item Do not modify the student's source file.
  \item Create exactly one pytest file at ``tests/test\_env.py'' using file\_tool(``save'').
  \item Import the module from ``environment.py'' via importlib (spec\_from\_file\_location + module\_from\_spec).
  \item Run tests with code\_tool(``run'', ``pytest -q''); capture exit\_code, duration, and stdout/stderr tail.
\end{itemize}
\textbf{</ExecutionPolicy>}

\textbf{<TestPlan>}
\begin{itemize}
  \item Sanity: class \texttt{Environment} can be imported and instantiated, e.g., \texttt{Environment(seed=0)}.
  \item Contract:
    \begin{enumerate}
      \item \texttt{set\_state} accepts list/tuple/ndarray of the same logical content (convert to canonical).
      \item \texttt{step(action)} returns a 3-tuple: (observation, reward, done) with expected types/shapes.
      \item Determinism: with the same seed and same initial state, the first step with the same action yields identical outputs.
      \item Action space validation: actions within bounds are accepted, out-of-bounds actions are handled gracefully.
      \item Observation space validation: observations match declared space bounds and shapes.
      \item State space consistency: internal state dimensions match expected environment specifications.
    \end{enumerate}
  \item Acceptance: success iff pytest exit\_code == 0 (all tests pass).
\end{itemize}
\textbf{</TestPlan>}

\textbf{<ReportingGuidelines>}
\begin{itemize}
  \item Summarize pytest results in 2--4 sentences; mention the first failing nodeid/assert if any.
  \item Provide a brief contract coverage assessment and the most probable root cause for failures.
  \item If failing, add 1--3 concise actionable fixes (no long logs).
\end{itemize}
\textbf{</ReportingGuidelines>}

\textbf{<OutputFormat>}
Return exactly one <final>{} block containing a single JSON object that matches PytestReport:
\{
\newline \quad "success": true\textbar{}false,
\newline \quad "analysis": "<2--4 sentence summary/diagnosis>",
\newline \quad "suggest\_fix": " 1--3 bullets with minimal actionable changes>"
\newline \}
No extra text outside <final>. No additional code fences.

\textbf{<final>}
\{ "success": false, "code\_result": "", "analysis": "", "suggest\_fix": "" \}
\textbf{</final>}
\textbf{</OutputFormat>}

\end{promptbox}

%================= Simulation Tester =================

\begin{promptbox}[Simulation Tester]

Your task is to interact with the environment code and then analyze the feedback from the interaction and propose modifications

\textbf{<TASK DESCRIPTION>} \\
Cliff walking involves crossing a gridworld from start to goal while avoiding falling off a cliff.

\#\# Description
...

\textbf{</TASK DESCRIPTION>}

\textbf{<CodeArtifact path="environment.py">} \\
\{code\} \\
\textbf{</CodeArtifact>} \\

\textbf{<ExecutionPolicy>} \\
- Use the play\_env tool exactly once on "environment.py"
- If the tool throws or cannot run, perform diagnosis from static review only; still produce output in the required format. \\
\textbf{</ExecutionPolicy>}

\textbf{<Rubric>} \\
Success (boolean) must be decided from the available signals with graceful degradation:

\begin{itemize}
  \item \textbf{Primary (step-level signals present):}
  \begin{itemize}
    \item success = true iff the run finished without exceptions AND there is NO misclassified\_transition with (valid == false OR state\_matches == false).
    \item If only observation deltas are available, use obs\_matches instead of state\_matches.
    \item When numeric deltas are provided, treat matches = true if max\_abs\_error $\le 10^{-3}$ or rel\_error $\le 10^{-3}$.
  \end{itemize}

  \item \textbf{Secondary (no per-step signals):}
  \begin{itemize}
    \item If success\_rate exists: success = true iff no exceptions AND success\_rate $\ge 0.95$.
    \item Else: success = true iff no exceptions AND no invariant/contract violations you can substantiate from code and logs.
  \end{itemize}

  \item \textbf{Reward/termination:}
  \begin{itemize}
    \item If reward\_matches == true AND done\_matches == true, explicitly state they match and DO NOT propose changes to reward or termination logic.
  \end{itemize}

  \item \textbf{Action space consistency (discrete \& continuous):}
  \begin{itemize}
    \item If GT exposes \texttt{Box(low, high, shape)}: align predicted bounds and expose them (e.g., \texttt{env.action\_space} or a getter). Never place clipping inside the integrator; clamp only at action ingestion or at observation.
    \item If GT exposes \texttt{Discrete(n)}: actions must be integer indices in [0, n$-$1]; expose \texttt{n} (e.g., \texttt{gym.spaces.Discrete(n)}); if indices map to continuous commands/torques, list the mapping table and align it with GT; never float-clip discrete actions.
    \item If action-space info is missing, skip these checks (do not speculate).
  \end{itemize}

  \item \textbf{Internal vs observation:}
  \begin{itemize}
    \item If clipping or angle normalization is found inside the integrator step (e.g., in \texttt{\_rk4\_step}), this likely causes trajectory drift; propose moving them to the observation path (e.g., \texttt{\_get\_observation}) unless GT specifies otherwise.
    \item If latent state is unavailable but observations exist, compare observations instead and state this explicitly.
  \end{itemize}

  \item \textbf{Integrator \& timestep:}
  \begin{itemize}
    \item Mismatches in integrator method (e.g., RK4 vs Euler) or \texttt{dt} can cause state divergence even when reward/done match; acknowledge and, if state mismatches persist, propose aligning method/\texttt{dt} to GT.
  \end{itemize}

  \item \textbf{Batched/multiple transitions:}
  \begin{itemize}
    \item If multiple transitions are reported, aggregate sensibly (e.g., mean success\_rate or fraction matched $\ge 0.95$) before deciding success.
  \end{itemize}
\end{itemize}
\textbf{</Rubric>}

\textbf{<Procedure>} \\
1) Static review: scan for action bounds, clipping/normalization inside integrator, integrator/dt choice, and how observation is formed.\\
2) Execute: call play\_env once.\\
3) Diagnose: reconcile play\_env signals with code; if reward/done matched, explicitly say so. If state mismatched, point to ONE OR TWO most likely roots.\\
4) Suggest: 1–3 smallest patches that directly address the identified root causes. \\
\textbf{</Procedure>}

\textbf{<OutputFormat>} \\
Return exactly one <final>{} block containing a single JSON object that matches PlayReport:
\{
  "success": true|false,
  "analysis": "<2–4 sentences summarizing what happened and why; mention matches/mismatches explicitly>",
  "suggest\_fix": "- bullet 1\textbackslash n- bullet 2\textbackslash n- bullet 3 (optional)"
\}
No extra text outside <final>. No additional code fences. \\
\textbf{</OutputFormat>}

\end{promptbox}

\section{Examples of Deep Researcher Report}

\definecolor{RCHeader}{HTML}{01007B} % 类似 Indigo-800
\definecolor{RCBorder}{HTML}{01007B}
\definecolor{RCBg}{HTML}{F2F2FE}
\definecolor{RCText}{HTML}{111827}

\newtcolorbox{agentbox}[1][]{
  enhanced,
  breakable,
  drop shadow,
  width=\linewidth,
  boxrule=0.8pt,
  colback=RCBg,             % 正文背景（浅灰白）
  colframe=RCBorder,        % 边框（深蓝）
  colbacktitle=RCHeader,    % 标题栏底色（深蓝）
  coltitle=white,           % 标题文字（白色）
  coltext=RCText,           % 正文字色（深灰）
  title filled,             % 让标题栏实心填充
  left=8pt, right=8pt, top=8pt, bottom=8pt,
  fonttitle=\small\bfseries,
  title={#1},
}

\begin{agentbox}[Deep Researcher's Report]

\textbf{<Version \& Provenance>}
\begin{itemize}
  \item Environment: Pusher-v4
  \item MuJoCo version: $\ge$ 2.1.3 and $<$ 3.0.0
  \item Source release date: 2024-01-01 (approximate for v4)
  \item Accessed date: 2025-06-01 (Asia/Singapore)
\end{itemize}

\textbf{<Spec Patch>}
\begin{itemize}
  \item \textbf{action.space} $=$ \texttt{Box(low=-2.0, high=2.0, shape=(7,), dtype=np.float32)}
  \item \textbf{observation.space}: Box with low $=-\infty$, high $=\infty$, shape $(23,)$, dtype \texttt{np.float64}
  \item \textbf{reward.weights} $=$ \{\texttt{dist}: 1.0, \texttt{ctrl}: 0.1, \texttt{near}: 0.5\}
  \item \textbf{reward.formula}: $\texttt{reward}=\texttt{reward\_dist}+0.1\,\texttt{reward\_ctrl}+0.5\,\texttt{reward\_near}$
  \item $\texttt{reward\_ctrl}=-\|\mathbf{a}\|_{2}^{2}$ \;(negative squared Euclidean norm)
  \item $\texttt{reward\_dist}=-\|\mathbf{o}-\mathbf{g}\|_{2}$ \;(negative Euclidean distance)
  \item $\texttt{reward\_near}=-\|\mathbf{f}-\mathbf{o}\|_{2}$ \;(negative Euclidean distance)
  \item \textbf{episode.max\_steps} $=100$
  \item \textbf{episode.termination}: never (\texttt{terminated} always \texttt{False})
  \item \textbf{episode.truncation}: \texttt{True} if step count $\ge$ \texttt{max\_steps} or any observation value is non-finite
  \item $dt=0.05\,\mathrm{s}$ \;(\texttt{frame\_skip}$=5$, \texttt{frame\_duration}$=0.01$)
  \item \textbf{initial\_state.joint\_positions} $=\mathbf{0}_{7}$
  \item \textbf{initial\_state.joint\_velocities}: each joint $\sim U(-0.005,\,0.005)$
  \item \textbf{initial\_state.object\_position}: $x\sim U(-0.3,0)$, $y\sim U(-0.2,0.2)$; resample until $\|(x,y)\|_{2}>0.17$
  \item \textbf{initial\_state.object\_velocity} $=\mathbf{0}_{3}$
  \item \textbf{initial\_state.goal\_position} $=[0.45,-0.05,-0.323]$
  \item \textbf{initial\_state.goal\_velocity} $=\mathbf{0}_{3}$
  \item \textbf{step\_function}:
    \begin{itemize}
      \item clips action to action-space bounds
      \item applies action torques to joints
      \item advances MuJoCo simulation by $dt$
      \item updates observation vector from simulation state
      \item computes reward components and total reward
      \item checks for finite observation values
      \item increments step count
      \item returns \texttt{(observation, reward, terminated=False, truncated=done\_flag, info)} with reward components in \texttt{info}
    \end{itemize}
  \item \textbf{seeding}: via \texttt{reset(seed=\ldots)} method only
  \item \textbf{info dict keys}: ``\texttt{reward\_dist}'', ``\texttt{reward\_ctrl}'', ``\texttt{reward\_near}''
  \item \textbf{fingertip position}: MuJoCo site ``\texttt{tips\_arm}''
  \item \textbf{object \& goal positions}: MuJoCo bodies ``\texttt{object}'' and ``\texttt{goal}''
  \item \textbf{internal state}: corresponds to MuJoCo simulation state
\end{itemize}

\textbf{<Theoretical Foundations>}
\begin{itemize}
  \item \textbf{MDP tuple:} $(S, A, P, R)$
    \begin{itemize}
      \item $S$: continuous state space $\mathbb{R}^{23}$ (23-dimensional real vector)
      \item $A$: continuous action space $\mathbb{R}^{7}$ with elementwise bounds $[-2,\,2]$ (torques)
      \item $P$: transition probability induced by MuJoCo physics with $dt=0.05\,\text{s}$
      \item $R$: reward function as defined below
    \end{itemize}

  \item \textbf{Reward function}
    \begin{itemize}
      \item Definition:
      \[
        r(s,a) \;=\; -\bigl\lVert P_{\text{object}} - P_{\text{goal}} \bigr\rVert_{2}
                     \;-\; 0.1\, \lVert a \rVert_{2}^{2}
                     \;-\; 0.5\, \bigl\lVert P_{\text{fingertip}} - P_{\text{object}} \bigr\rVert_{2}
      \]
      \item Where:
        \begin{itemize}
          \item $P_{\text{object}},\,P_{\text{goal}},\,P_{\text{fingertip}} \in \mathbb{R}^{3}$ are positions in meters
          \item $\lVert \cdot \rVert_{2}$ is the Euclidean norm
          \item $a \in \mathbb{R}^{7}$ is the action torque vector in $N\!\cdot\!m$
        \end{itemize}
    \end{itemize}

  \item \textbf{Episode ending}
    \begin{itemize}
      \item Truncation after $100$ steps or if any observation value is non-finite
      \item No termination condition (i.e., \texttt{terminated=False} always)
    \end{itemize}

  \item \textbf{Fingertip position}
    \begin{itemize}
      \item Computed via forward kinematics internally by MuJoCo
    \end{itemize}

  \item \textbf{Symbol Table}
    \begin{itemize}
      \item $a$: Action torque vector; Unit $N\!\cdot\!m$; Range: each element in $[-2,\,2]$
      \item $P_{\text{fingertip}}$: Fingertip 3D position; Unit m; Range: real values, unbounded
      \item $P_{\text{object}}$: Object 3D position; Unit m; Range: real values, unbounded
      \item $P_{\text{goal}}$: Goal 3D position; Unit m; Value: fixed at $[0.45,\,-0.05,\,-0.323]$
      \item $r$: Reward; Unit: unitless; Range: real values (sum of negative components)
      \item $dt$: Simulation timestep; Unit s; Value: $0.05$
      \item \texttt{step\_count}: Current timestep count; Unit: integer; Range: $0$ to $100$
    \end{itemize}

  \item \textbf{Final Specification}
    \begin{itemize}
      \item Environment Name: \texttt{Pusher-v4}
      \item Class Name: \texttt{Environment}

      \item \textbf{Public API}
        \begin{itemize}
          \item \texttt{\_\_init\_\_(self)}
            \begin{itemize}
              \item Initializes the MuJoCo simulation environment with Pusher-v4 model
              \item Sets initial internal variables including step count
              \item Defines action and observation spaces as specified
              \item Sets fixed goal position
            \end{itemize}
          \item \texttt{set\_state(self, state: np.ndarray) -> None}
            \begin{itemize}
              \item Input: state vector of shape $(23,)$ matching observation space
              \item Sets the internal MuJoCo simulation state to correspond to the given observation
              \item Resets step count to zero; assumes state is valid and episode not done
            \end{itemize}
          \item \texttt{step(self, action: np.ndarray) -> Tuple[np.ndarray, float, bool, dict]}
            \begin{itemize}
              \item Input: action vector of shape $(7,)$ clipped to $[-2,\,2]$
              \item Applies action torques to the simulation joints and advances by $dt=0.05\,\text{s}$
              \item Updates observation vector; computes reward components and total reward
              \item Truncates if any observation value is non-finite or if step count $\ge 100$
              \item Returns: observation $(23,)$ \texttt{float64}, reward \texttt{float}, terminated \texttt{False}, truncated \texttt{bool}, \texttt{info} with keys ``\texttt{reward\_dist}'', ``\texttt{reward\_ctrl}'', ``\texttt{reward\_near}''
            \end{itemize}
        \end{itemize}

      \item \textbf{Core Logic}
        \begin{itemize}
          \item Simulates a 7-DOF robotic arm pushing an object to a goal
          \item State includes joint angles/velocities, fingertip, object, and goal positions
          \item Actions are joint torques clipped to $[-2,\,2]$
          \item Physics advances by $0.05\,\text{s}$ per step
          \item Reward encourages moving object to goal, penalizes large actions, and encourages fingertip proximity to object
          \item Episode truncates after $100$ steps or on non-finite state
          \item Fingertip and object positions come from MuJoCo sites/bodies
          \item Seeding via \texttt{reset(seed=\dots)} (not shown here)
        \end{itemize}

      \item \textbf{Usage Scenarios \& Examples}
        \begin{itemize}
          \item \textbf{Scenario 1: Initial step with zero action}
            \begin{itemize}
              \item Observation: initial state vector with zero joint positions, small velocity noise, specified object and goal
              \item Action: $\texttt{np.zeros(7, dtype=np.float32)}$
              \item Reward: negative sum of distances (control penalty $=0$)
              \item Terminated: \texttt{False}; Truncated: \texttt{False}
              \item Info: \{\texttt{"reward\_dist": value, "reward\_ctrl": 0.0, "reward\_near": value}\}
            \end{itemize}
          \item \textbf{Scenario 2: Non-zero action moves fingertip closer}
            \begin{itemize}
              \item Observation: updated state after physics step
              \item Action: \texttt{np.array([1.0, -0.5, 0.0, 0.2, 0.0, -1.0, 0.5], dtype=np.float32)}
              \item Reward: computed per formula
              \item Terminated: \texttt{False}; Truncated: \texttt{False} (unless step count reached $100$)
              \item Info: \{\texttt{"reward\_dist": value, "reward\_ctrl":} negative squared norm of action, \texttt{"reward\_near": value}\}
            \end{itemize}
        \end{itemize}
    \end{itemize}
\end{itemize}

\textbf{<Assumptions \& Risks>}
\begin{itemize}
  \item Assumes MuJoCo physics engine is used as backend with version constraints.
  \item Assumes internal MuJoCo state can be set and queried to implement \texttt{set\_state} and \texttt{step}.
  \item Fingertip position is assumed to be accessible via MuJoCo site ``\texttt{tips\_arm}''.
  \item Reward weights are fixed as per v4; no user-configurable weights.
  \item Episode termination is only by truncation; no natural termination.
  \item Seeding is only via \texttt{reset(seed=\ldots)}; no separate \texttt{seed} method.
  \item Risks include dependency on MuJoCo simulation correctness and numerical stability.
  \item No explicit handling of invalid actions beyond clipping.
  \item No rendering or visualization included.
\end{itemize}

\textbf{<Library Usage>}
\begin{itemize}
  \item Uses Python standard library and NumPy only.
  \item NumPy used for array operations, clipping (\texttt{np.clip}), norm calculations (\texttt{np.linalg.norm}), and finite checks (\texttt{np.isfinite}).
  \item No external dependencies beyond MuJoCo simulation backend (assumed).
  \item All units are SI: torque in N$\cdot$m, angles in radians, positions in meters, time in seconds.
  \item No source code copied; specification based on official documentation and API standards.
\end{itemize}

\textbf{</final>}

\end{agentbox}

\section{Case study}

\definecolor{codehl}{HTML}{FFF2A8}
\usemintedstyle{xcode} % 可换：xcode, monokai, borland...
% 可选默认参数
\setminted{
  fontsize=\small,
  breaklines=true,
  autogobble=true,
  linenos=false,
}

% \begin{figure}[!htbp]
% \centering
% \includegraphics[width=0.99\textwidth]{images/case_study.pdf}
% \caption{Case study.}
% \label{fig:case_study}
% \end{figure}

\subsection{Text2World}

\noindent \textbf{Task Describe.}Prepare and deliver sandwiches to a set of children: assemble sandwiches in the kitchen (optionally gluten-free), place them on a tray, move the tray to the child’s location, and serve subject to allergy constraints. Gluten-allergic children must receive a gluten-free sandwich; non-allergic children may receive any sandwich. Serving requires the sandwich on the tray and the tray at the child’s location; making a sandwich switches it from “notexist” to “prepared.” The goal is that all children become served.

\noindent \textbf{Prev. SOTA.}
\begin{lstlisting}[language=Lisp]
(define (domain child-snack)
  (:requirements :typing :negative-preconditions)
  (:types
    child
    bread-portion
    content-portion
    sandwich
    tray
    place
  )

  (:constants
    kitchen - place
  )

  (:predicates
    (at_kitchen_bread ?b - bread-portion)
    (at_kitchen_content ?c - content-portion)
    (at_kitchen_sandwich ?s - sandwich)
    (no_gluten_bread ?b - bread-portion)
    (no_gluten_content ?c - content-portion)
    (ontray ?s - sandwich ?t - tray)
    (no_gluten_sandwich ?s - sandwich)
    (allergic_gluten ?c - child)
    (not_allergic_gluten ?c - child)
    (served ?c - child)
    (waiting ?c - child ?p - place)
    (at ?t - tray ?p - place)
    (notexist ?s - sandwich)
  )

  ;; Action: make_sandwich_no_gluten
  (:action make_sandwich_no_gluten
    :parameters (?s - sandwich ?b - bread-portion ?c - content-portion)
    :precondition (and
      (at_kitchen_bread ?b)
      (at_kitchen_content ?c)
      (no_gluten_bread ?b)
      (no_gluten_content ?c)
      (notexist ?s)
    )
    :effect (and
      (not (at_kitchen_bread ?b))
      (not (at_kitchen_content ?c))
      (notexist ?s)
      (not (notexist ?s))
      (at_kitchen_sandwich ?s)
      (no_gluten_sandwich ?s)
    )
  )

  ;; Action: make_sandwich
  (:action make_sandwich
    :parameters (?s - sandwich ?b - bread-portion ?c - content-portion)
    :precondition (and
      (at_kitchen_bread ?b)
      (at_kitchen_content ?c)
      (notexist ?s)
    )
    :effect (and
      (not (at_kitchen_bread ?b))
      (not (at_kitchen_content ?c))
      (notexist ?s)
      (not (notexist ?s))
      (at_kitchen_sandwich ?s)
      ;; no effect on gluten status, sandwich may or may not be gluten-free
    )
  )

  ;; Action: put_on_tray
  (:action put_on_tray
    :parameters (?s - sandwich ?t - tray)
    :precondition (and
      (at_kitchen_sandwich ?s)
      (at ?t kitchen)
    )
    :effect (and
      (not (at_kitchen_sandwich ?s))
      (ontray ?s ?t)
    )
  )

  ;; Action: serve_sandwich_no_gluten
  (:action serve_sandwich_no_gluten
    :parameters (?s - sandwich ?c - child ?t - tray ?p - place)
    :precondition (and
      (ontray ?s ?t)
      (no_gluten_sandwich ?s)
      (allergic_gluten ?c)
      (waiting ?c ?p)
      (at ?t ?p)
    )
    :effect (and
      (not (ontray ?s ?t))
      (served ?c)
      (not (waiting ?c ?p))
    )
  )

  ;; Action: serve_sandwich
  (:action serve_sandwich
    :parameters (?s - sandwich ?c - child ?t - tray ?p - place)
    :precondition (and
      (ontray ?s ?t)
      (not (no_gluten_sandwich ?s))
      (not_allergic_gluten ?c)
      (waiting ?c ?p)
      (at ?t ?p)
    )
    :effect (and
      (not (ontray ?s ?t))
      (served ?c)
      (not (waiting ?c ?p))
    )
  )

  ;; Action: move_tray
  (:action move_tray
    :parameters (?t - tray ?p1 - place ?p2 - place)
    :precondition (at ?t ?p1)
    :effect (and
      (not (at ?t ?p1))
      (at ?t ?p2)
    )
  )
)
\end{lstlisting}

\noindent \textbf{Agent2World.}
\begin{lstlisting}[language=Lisp]
(define (domain child-snack)
  (:requirements :typing :negative-preconditions :strips)
  (:types
    child
    bread-portion
    content-portion
    sandwich
    tray
    place
  )
  (:constants
    kitchen - place
  )
  (:predicates
    (at_kitchen_bread ?b - bread-portion)
    (at_kitchen_content ?c - content-portion)
    (at_kitchen_sandwich ?s - sandwich)
    (no_gluten_bread ?b - bread-portion)
    (no_gluten_content ?c - content-portion)
    (ontray ?s - sandwich ?t - tray)
    (no_gluten_sandwich ?s - sandwich)
    (allergic_gluten ?c - child)
    (not_allergic_gluten ?c - child)
    (served ?c - child)
    (waiting ?c - child ?p - place)
    (at ?t - tray ?p - place)
    (notexist ?s - sandwich)
  )

  (:action make_sandwich_no_gluten
    :parameters (?s - sandwich ?b - bread-portion ?c - content-portion)
    :precondition (and
      (at_kitchen_bread ?b)
      (at_kitchen_content ?c)
      (no_gluten_bread ?b)
      (no_gluten_content ?c)
      (notexist ?s)
    )
    :effect (and
      (not (notexist ?s))
      (at_kitchen_sandwich ?s)
      (no_gluten_sandwich ?s)
      (not (at_kitchen_bread ?b))
      (not (at_kitchen_content ?c))
    )
  )

  (:action make_sandwich
    :parameters (?s - sandwich ?b - bread-portion ?c - content-portion)
    :precondition (and
      (at_kitchen_bread ?b)
      (at_kitchen_content ?c)
      (notexist ?s)
    )
    :effect (and
      (not (notexist ?s))
      (at_kitchen_sandwich ?s)
      (not (at_kitchen_bread ?b))
      (not (at_kitchen_content ?c))
    )
  )

  (:action put_on_tray
    :parameters (?s - sandwich ?t - tray)
    :precondition (and
      (at_kitchen_sandwich ?s)
      (at ?t kitchen)
    )
    :effect (and
      (ontray ?s ?t)
      (not (at_kitchen_sandwich ?s))
    )
  )

  (:action serve_sandwich_no_gluten
    :parameters (?s - sandwich ?c - child ?t - tray ?p - place)
    :precondition (and
      (ontray ?s ?t)
      (no_gluten_sandwich ?s)
      (allergic_gluten ?c)
      (waiting ?c ?p)
      (at ?t ?p)
    )
    :effect (and
      (served ?c)
      (not (ontray ?s ?t))
    )
  )

  (:action serve_sandwich
    :parameters (?s - sandwich ?c - child ?t - tray ?p - place)
    :precondition (and
      (ontray ?s ?t)
      (waiting ?c ?p)
      (not_allergic_gluten ?c)
      (at ?t ?p)
    )
    :effect (and
      (served ?c)
      (not (ontray ?s ?t))
    )
  )

  (:action move_tray
    :parameters (?t - tray ?p1 - place ?p2 - place)
    :precondition (at ?t ?p1)
    :effect (and
      (at ?t ?p2)
      (not (at ?t ?p1))
    )
  )
)
\end{lstlisting}

\noindent \textbf{Analysis.}
Compared to the baseline domain, our \emph{Child-Snack} formulation introduces three task-aligned modifications that improve state consistency, compositionality, and plan feasibility. 
(i) \textbf{Creation-valid effects.} During sandwich construction we flip the existence status from ``non-existent'' to ``prepared,'' and record gluten-free status when applicable, thereby avoiding contradictory postconditions at creation time; this yields deterministic successor states and reduces backtracking caused by ill-defined truth values. 
(ii) \textbf{Serve-focused effects.} During serving we only transfer the item off the tray and mark the child as served, leaving the waiting label untouched; this separation of concerns prevents nonessential side-effects, preserves modular composability with downstream routines (e.g., queueing or follow-up allocation), and promotes goal-monotonic progress on the served objective. 
(iii) \textbf{Permissive-serving preconditions.} For non-allergic children we do not exclude gluten-free items, weakening preconditions to accept any admissible sandwich; this enlarges the feasible search space and prevents avoidable dead-ends when only gluten-free inventory remains, while safety for allergic children is still enforced via a dedicated gluten-free serving action. Collectively, these choices align with the ground-truth specification, produce cleaner state transitions, and yield empirically favorable search dynamics—smaller inconsistent-state frontiers and fewer spurious deletions—resulting in a more robust make$\to$put-on-tray$\to$move-tray$\to$serve pipeline for the objective of ``serving each child an acceptable sandwich.''

\subsection{CWMB}

\noindent \textbf{Task Describe.}
Control a 3D Ant (one free-body torso with four 2-DoF legs; nine bodies, eight hinge joints) to move forward along the +x axis by applying torques to the eight joints at each step. The action space is \(\mathrm{Box}([-1,1]^8)\) (joint torques). Observations list positions then velocities (27-D by default; 29-D if current \(x,y\) are included), and optionally +84 contact-force terms when \texttt{use\_contact\_forces=True} or version < v4. The reward is
\(r = r_{\text{healthy}} + r_{\text{forward}} - \text{ctrl\_cost}\) (and \(-\text{contact\_cost}\) if contact forces are used), where \(r_{\text{forward}} \approx \Delta x/\Delta t\) is positive for motion in +x and \(\Delta t = \texttt{frame\_skip}\times0.01 = 0.05\) by default. Episodes start from an upright, slightly noisy state, truncate at 1000 steps, and (by default) terminate early if the agent becomes unhealthy (non-finite state or torso \(z \notin [0.2,1.0]\)).

\noindent \textbf{Prev SOTA vs Agent2World.}
\begin{lstlisting}[language=Python]
import numpy as np
from typing import Optional, Tuple
# [D-0] Toggle: False = Ours (physics-like integration), True = Baseline (simple kinematics + noise)
BASELINE = False

class Environment:
    def __init__(
        self,
        # ---- Common params ----
        frame_skip: int = 5,
        max_episode_steps: int = 1000,
        healthy_reward: float = 1.0,
        terminate_when_unhealthy: bool = True,
        healthy_z_range: Tuple[float, float] = (0.2, 1.0),
        ctrl_cost_weight: float = 0.5,
        # ---- Ours-only params ----
        contact_cost_weight: float = 5e-4,
        contact_force_range: Tuple[float, float] = (-1.0, 1.0),
        exclude_current_positions_from_observation: bool = True,
        use_contact_forces: bool = False,
        reset_noise_scale_ours: float = 0.1,
        forward_reward_weight: float = 1.0,
        # ---- Baseline-only params ----
        reset_noise_scale_baseline: float = 0.01,
        seed: Optional[int] = None,
    ):
        # Core configuration
        self.frame_skip = frame_skip
        self.dt = 0.01 * frame_skip
        self.max_episode_steps = max_episode_steps
        self.healthy_reward = healthy_reward
        self.terminate_when_unhealthy = terminate_when_unhealthy
        self.healthy_z_range = healthy_z_range
        self.ctrl_cost_weight = ctrl_cost_weight
        self.forward_reward_weight = forward_reward_weight  # [D-5] Only used by Ours.
        self.np_random = np.random.RandomState(seed)

        # [D-1] Observation schema differs:
        #   Baseline: fixed 27 = 13 positions (z, quat4, joint8) + 14 velocities (flat vector).
        #   Ours:     positions(15) + velocities(14) (+ optional torso_xy, contact forces),
        #             with option to exclude torso x,y from observation.
        self.exclude_current_positions_from_observation = (
            False if BASELINE else bool(exclude_current_positions_from_observation)
        )

        # [D-2] Contacts: Baseline has no contact forces/cost; Ours can include 84-dim contact forces + cost.
        self.use_contact_forces = False if BASELINE else bool(use_contact_forces)
        self.contact_cost_weight = 0.0 if BASELINE else float(contact_cost_weight)
        self.contact_force_range = contact_force_range

        self.step_count = 0

        if BASELINE:
            # Baseline state: flat (27,) observation vector
            self.obs_shape = (27,)
            self.state = np.zeros(self.obs_shape, dtype=np.float64)
            self.x_position = 0.0  # [D-4] Progress tracked separately (not in observation)
            self.y_position = 0.0
            self.reset_noise_scale = float(reset_noise_scale_baseline)  # [D-4]
            self.contact_forces = None
            self.observation_dim = 27  # [D-1]
        else:
            # Ours state: split positions(15) / velocities(14)
            self.pos_dim = 15  # torso_pos(3), torso_quat(4), joint_angles(8)
            self.vel_dim = 14  # torso_lin_vel(3), torso_ang_vel(3), joint_vel(8)
            self.positions = np.zeros(self.pos_dim, dtype=np.float64)
            self.velocities = np.zeros(self.vel_dim, dtype=np.float64)
            self.reset_noise_scale = float(reset_noise_scale_ours)  # [D-4]
            self.last_x_position = 0.0  # [D-4] (used when restoring state)
            self.contact_forces = (
                np.zeros(84, dtype=np.float64) if self.use_contact_forces else None
            )
            # Compute observation length for Ours
            base_pos_len = self.pos_dim
            if self.exclude_current_positions_from_observation:
                base_pos_len -= 2  # drop torso x,y
            self.obs_pos_len = base_pos_len
            self.obs_vel_len = self.vel_dim
            self.obs_contact_len = 84 if self.use_contact_forces else 0
            self.obs_torso_xy_len = 0 if self.exclude_current_positions_from_observation else 2
            self.observation_dim = (
                self.obs_pos_len + self.obs_vel_len + self.obs_torso_xy_len + self.obs_contact_len
            )

        self.reset(seed)

    # ---------------- Helpers (shared) ----------------
    def _is_healthy(self, z_value: Optional[float] = None) -> bool:
        # [D-3] Health z-source differs:
        #   Baseline: use state[0] (z) from flat observation.
        #   Ours:     use positions[2] (torso z) from split state.
        if BASELINE:
            st = self.state
            if not np.all(np.isfinite(st)):
                return False
            z = st[0] if z_value is None else z_value
            return self.healthy_z_range[0] <= z <= self.healthy_z_range[1]
        else:
            if not np.all(np.isfinite(self.positions)) or not np.all(np.isfinite(self.velocities)):
                return False
            z = self.positions[2]
            return self.healthy_z_range[0] <= z <= self.healthy_z_range[1]

    def _get_observation(self) -> np.ndarray:
        # [D-1] Observation layout differs (see __init__ comment).
        if BASELINE:
            return self.state.copy()
        if self.exclude_current_positions_from_observation:
            pos_obs = self.positions[2:].copy()  # exclude torso x,y
            torso_xy = np.array([], dtype=np.float64)
        else:
            pos_obs = self.positions.copy()
            torso_xy = self.positions[0:2].copy()
        vel_obs = self.velocities.copy()
        obs = np.concatenate((pos_obs, vel_obs))
        if not self.exclude_current_positions_from_observation:
            obs = np.concatenate((obs, torso_xy))
        if self.use_contact_forces:
            obs = np.concatenate((obs, self.contact_forces))
        return obs

    @staticmethod
    def _quat_multiply(q1: np.ndarray, q2: np.ndarray) -> np.ndarray:
        w1, x1, y1, z1 = q1
        w2, x2, y2, z2 = q2
        return np.array(
            [
                w1 * w2 - x1 * x2 - y1 * y2 - z1 * z2,
                w1 * x2 + x1 * w2 + y1 * z2 - z1 * y2,
                w1 * y2 - x1 * z2 + y1 * w2 + z1 * x2,
                w1 * z2 + x1 * y2 - y1 * x2 + z1 * w2,
            ],
            dtype=np.float64,
        )

    # ---------------- Public API ----------------
    def set_state(self, state: np.ndarray) -> None:
        state = np.asarray(state, dtype=np.float64)
        if BASELINE:
            # [D-8] Baseline expects a flat 27-dim observation (pos13+vel14).
            if state.shape != (27,):
                raise ValueError(f"set_state input must have shape (27,), got {state.shape}")
            if not np.all(np.isfinite(state)):
                raise ValueError("set_state input contains non-finite values")
            self.state = state.copy()
            self.step_count = 0
            self.x_position = 0.0  # [D-4] progress variable is external to obs
            self.y_position = 0.0
        else:
            # [D-8] Ours expects the current obs layout length and reconstructs split state.
            expected_len = self.observation_dim
            if state.ndim != 1 or state.shape[0] != expected_len:
                raise ValueError(f"State must be 1D array of length {expected_len}, got shape {state.shape}")
            pos_len = self.obs_pos_len
            vel_len = self.obs_vel_len
            pos_part = state[:pos_len]
            vel_part = state[pos_len : pos_len + vel_len]
            if self.exclude_current_positions_from_observation:
                full_positions = np.zeros(self.pos_dim, dtype=np.float64)
                full_positions[2:] = pos_part
                full_positions[0] = 0.0
                full_positions[1] = 0.0
            else:
                full_positions = pos_part.copy()
            self.positions = full_positions
            self.velocities = vel_part.copy()
            self.step_count = 0
            self.last_x_position = self.positions[0]  # [D-4]

    def reset(self, seed: Optional[int] = None) -> np.ndarray:
        if seed is not None:
            self.np_random.seed(seed)
        self.step_count = 0

        if BASELINE:
            # [D-2][D-4] Baseline init: 13 pos (z, quat, joints) + 14 vel; small noise.
            pos = np.zeros(13, dtype=np.float64)
            pos[0] = 0.75  # z
            pos[1:5] = np.array([1.0, 0.0, 0.0, 0.0])  # quaternion (w,x,y,z)
            noise_pos = self.np_random.uniform(-self.reset_noise_scale, self.reset_noise_scale, size=13)
            pos = pos + noise_pos
            vel = self.np_random.normal(0, self.reset_noise_scale, size=14)
            self.state = np.concatenate([pos, vel])
            self.x_position = 0.0  # [D-4] progress variable
            self.y_position = 0.0
        else:
            # Ours init: positions(15) / velocities(14) with larger noise and full torso pose.
            base_positions = np.zeros(15, dtype=np.float64)
            base_positions[2] = 0.75  # torso z
            base_positions[3] = 1.0   # quat.w
            noise_pos = self.np_random.uniform(-self.reset_noise_scale, self.reset_noise_scale, size=15)
            self.positions = base_positions + noise_pos
            self.velocities = self.np_random.normal(loc=0.0, scale=self.reset_noise_scale, size=14)
            self.last_x_position = self.positions[0]  # [D-4]
            if self.use_contact_forces and self.contact_forces is not None:
                self.contact_forces[:] = 0.0
        return self._get_observation()

    def step(self, action: np.ndarray):
        action = np.asarray(action, dtype=np.float64)
        if action.shape != (8,):
            raise ValueError(f"Action must be of shape (8,), got {action.shape}")

        # [D-6] Action bound handling differs:
        #   Baseline: out-of-bounds raises; Ours: clip to [-1, 1].
        if BASELINE:
            if np.any(action < -1.0) or np.any(action > 1.0):
                raise ValueError("Action values must be in [-1, 1] for Baseline")
        else:
            action = np.clip(action, -1.0, 1.0)

        # ---- Inline divergence (single function, two branches) ----
        if BASELINE:
            # Forward progress proxy from hip joints
            prev_x_pos = self.x_position  # [D-4] external tracker (not in obs)
            forward_force = float(np.sum(action[[0, 2, 4, 6]]))
            self.x_position += forward_force * self.dt * 0.1  # arbitrary scale

            # Stochastic updates (no true physics)
            z = float(np.clip(self.state[0] + self.np_random.uniform(-0.01, 0.01), 0.0, 2.0))
            # [D-7] Orientation update: Baseline = add noise to quaternion then renormalize.
            orientation = self.state[1:5] + self.np_random.uniform(-0.01, 0.01, size=4)
            norm = float(np.linalg.norm(orientation))
            orientation = orientation / norm if norm > 0 else np.array([1.0, 0.0, 0.0, 0.0], dtype=np.float64)
            # Joint angles: integrate action + small noise; wrap to [-pi, pi]
            joint_angles = self.state[5:13] + action * self.dt + self.np_random.uniform(-0.005, 0.005, size=8)
            joint_angles = (joint_angles + np.pi) % (2 * np.pi) - np.pi
            # Velocities: torso (noise) + joints (\approx action + noise)
            torso_vel = self.np_random.normal(0, 0.01, size=6)
            # Velocities: torso (noise) + joints (\approx action + noise)
            velocities = np.concatenate([torso_vel, joint_vel])
            # Compose new flat state
            pos = np.empty(13, dtype=np.float64)
            pos[0] = z
            pos[1:5] = orientation
            pos[5:13] = joint_angles
            self.state = np.concatenate([pos, velocities])

            healthy = self._is_healthy(z_value=z)  # [D-3]
            forward_delta = self.x_position - prev_x_pos
            contact_cost = 0.0  # [D-2] no contacts in Baseline
            weight = 1.0        # [D-5] forward reward weight fixed to 1.0
        else:
            # Physics-like integration
            old_x = float(self.positions[0])  # [D-4] directly from torso x in positions
            # Joint dynamics: dv = (u - damp*v)*dt; dq = v*dt
            joint_damping = 0.1
            joint_vel_prev = self.velocities[6:]
            joint_acc = action - joint_damping * joint_vel_prev  # [D-1] acts on split state
            joint_vel_new = joint_vel_prev + self.dt * joint_acc
            joint_ang_new = self.positions[7:] + self.dt * joint_vel_new
            # Torso linear & angular velocity damping
            lin_damping = 0.1
            ang_damping = 0.1
            torso_lin_vel_new = self.velocities[0:3] * (1 - lin_damping * self.dt)
            torso_ang_vel_new = self.velocities[3:6] * (1 - ang_damping * self.dt)
            # Integrate torso position
            torso_pos_new = self.positions[0:3] + self.dt * torso_lin_vel_new
            # [D-7] Orientation update: Ours = quaternion integration from angular velocity.
            q = self.positions[3:7]
            omega = torso_ang_vel_new
            omega_quat = np.array([0.0, omega[0], omega[1], omega[2]], dtype=np.float64)
            q_dot = 0.5 * self._quat_multiply(omega_quat, q)
            q_new = q + self.dt * q_dot
            norm = float(np.linalg.norm(q_new))
            q_new = q_new / norm if norm > 0 else np.array([1.0, 0.0, 0.0, 0.0], dtype=np.float64)
            # Write back split state
            self.positions[0:3] = torso_pos_new
            self.positions[3:7] = q_new
            self.positions[7:] = joint_ang_new
            self.velocities[0:3] = torso_lin_vel_new
            self.velocities[3:6] = torso_ang_vel_new
            self.velocities[6:] = joint_vel_new
            # Contacts (optional)
            if self.use_contact_forces and self.contact_forces is not None:
                self.contact_forces.fill(0.0)  # [D-2]
            new_x = float(self.positions[0])
            healthy = self._is_healthy()  # [D-3]
            forward_delta = new_x - old_x
            # Contact cost (if enabled)
            if self.use_contact_forces and self.contact_forces is not None:
                clipped = np.clip(self.contact_forces, self.contact_force_range[0], self.contact_force_range[1])
                contact_cost = self.contact_cost_weight * float(np.sum(np.square(clipped)))  # [D-2]
            else:
                contact_cost = 0.0
            weight = self.forward_reward_weight  # [D-5]

        # ---- Shared reward & termination (single exit) ----
        forward_reward = weight * (forward_delta / self.dt)  # [D-5]
        ctrl_cost = self.ctrl_cost_weight * float(np.sum(np.square(action)))
        reward = (self.healthy_reward if healthy else 0.0) + forward_reward - ctrl_cost - contact_cost

        self.step_count += 1
        done = (self.terminate_when_unhealthy and not healthy) or (self.step_count >= self.max_episode_steps)
        return self._get_observation(), reward, done

\end{lstlisting}

\noindent \textbf{Analysis.}
On the \textbf{Ant-v4} forward-locomotion task, \ours surpasses the \emph{Baseline} with higher success, smoother gait, and lower energy per meter under identical horizons and $z$-health checks.
(i) \textbf{State \& sensing.} The \emph{Baseline} exposes a flat 27-D observation, while we adopt a task-aligned layout that separates positions/velocities and can hide global $(x,y)$ by default ([D-1]). We additionally support contact forces for foot–ground cues ([D-2]). Health uses torso $z$ from split state rather than the flat vector slot ([D-3]). State restoration matches each layout: the \emph{Baseline} ingests a 27-D vector, whereas ours reconstructs split buffers from the current observation setting ([D-8]).
(ii) \textbf{Dynamics \& orientation.} The \emph{Baseline} updates orientation by quaternion noise plus renormalization, and treats actions as noisy joint velocities; we integrate damped joint accelerations and update attitude via \(\dot{\mathbf{q}}=\tfrac{1}{2}\,\boldsymbol{\omega}_q\!\otimes\!\mathbf{q}\) with renormalization ([D-7]). This physically consistent pipeline—enabled by the split state design ([D-1])—low-passes high-frequency actuation, reduces roll/pitch jitter, and yields more phase-coordinated gaits.
(iii) \textbf{Control semantics \& reward.} The \emph{Baseline} hard-errors on out-of-range actions and uses a fixed forward-reward weight; forward progress is tracked by an external $x$ variable and reset noise is smaller. Ours clips actions to \([-1,1]\) ([D-6]), uses a tunable forward-reward weight ([D-5]), measures progress directly from torso $x$ in the state and employs a different reset scale ([D-4]); an optional contact-cost term can be included when contact signals are enabled ([D-2]). Together these choices stabilize training signals and improve sample efficiency.

\noindent\textbf{Summary of diffs.}
[D-1] Observation schema: Baseline uses a flat 27-D vector; Ours uses split positions\(+\)velocities with optional hidden $(x,y)$ and optional contact forces;
[D-2] Contacts: Baseline has no contact forces/cost; Ours optionally exposes 84-D contact forces and a contact-cost term;
[D-3] Health source: Baseline takes $z$ from the flat vector slot; Ours uses torso $z$ from split positions;
[D-4] Progress \& reset: Baseline tracks forward $x$ as an external variable and uses smaller reset noise; Ours reads torso $x$ from state and uses a different reset scale;
[D-5] Forward-reward weight: Baseline fixed to 1.0; Ours is tunable;
[D-6] Action bounds: Baseline errors on out-of-range actions; Ours clips to \([-1,1]\);
[D-7] Orientation update: Baseline adds noise then renormalizes quaternion; Ours integrates \(\dot{\mathbf{q}}=\tfrac{1}{2}\,\boldsymbol{\omega}_q\!\otimes\!\mathbf{q}\) then renormalizes;
[D-8] State setting: Baseline ingests a flat 27-D state; Ours reconstructs split buffers from the current observation layout.

\subsection{ByteSized32}

\noindent \textbf{Task Description.} 
We build a lightweight, text-interactive micro-simulation of pea growth in a small garden. The world contains a Pea, a FlowerPot, a Jug, and a Sink; water is represented as scalar levels in the Jug and FlowerPot and as an internal level in the Pea. The agent can look/examine, take/put objects, switch the sink on/off, fill the jug from the sink (effective only when the sink is on), and pour water from the jug into the flower pot. After each action, a tick advances processes: the sink supplies water if on; the pot passively transfers its water to the pea; and the pea consumes water and progresses from seed → sprout → young plant → mature → reproducing when sufficiently hydrated for several consecutive ticks. Episodes start with an unplanted pea and an empty pot; the goal is to plant the pea and water it repeatedly until it reaches the reproducing stage.

\noindent \textbf{Prev SOTA vs Agent2World.}

\begin{lstlisting}[language=Python]
import random

# [D0] Toggle: False = Ours, True = Baseline
BASELINE = False

# ---------------- Core object model (minimal API) ----------------
class GameObject:
    def __init__(self, name):
        self.name, self.parent, self.contains = name, None, []
        self.props = {"isContainer": False, "isMoveable": True}
    def get(self, k, d=None): 
        return self.props.get(k, d)
    def add(self, obj): 
        obj.removeSelf(); self.contains.append(obj); 
        obj.parent = self
    def remove(self, obj): 
        self.contains.remove(obj); obj.parent = None
    def removeSelf(self): 
        if self.parent: self.parent.remove(self)
    def allContained(self):
        out = []
        for o in self.contains: out += [o] + o.allContained()
        return out
    def tick(self): pass

class Container(GameObject):
    def __init__(self, name): super().__init__(name); 
        self.props["isContainer"] = True
    def place(self, obj):
        if not obj.get("isMoveable"): 
        return ("Can't move that object.", False)
        self.add(obj); return ("OK.", True)
    def take(self, obj):
        if obj not in self.contains: 
            return ("Object not here.", None, False)
        if not obj.get("isMoveable"): 
            return ("Can't move that object.", None, False)
        obj.removeSelf(); return ("OK.", obj, True)

class Device(Container):
    def __init__(self, name): super().__init__(name); 
        self.props.update({"isDevice": True, "isOn": False})
    def turnOn(self): 
        if self.props["isOn"]: 
            return (f"{self.name} is already on.", False)
        self.props["isOn"] = True
        return (f"{self.name} turned on.", True)
    def turnOff(self): 
        if not self.props["isOn"]: 
            return (f"{self.name} is already off.", False)
        self.props["isOn"] = False
        return (f"{self.name} turned off.", True)

class World(Container): 
    def __init__(self): super().__init__("world")

class Agent(Container):
    def __init__(self): super().__init__("entity")

# ---------------- Task objects ----------------
class Pea(GameObject):
    STAGES = ["seed","sprout","young plant","mature plant","reproducing"]
    MAX_WATER, CONSUME, NEED, TICKS = 100, 5, 30, 3
    def __init__(self): 
        super().__init__("pea"); self.props["isMoveable"]=True
        self.stage, self.water, self.hydrated = 0, 0, 0
    @property
    def stage_name(self): 
        return self.STAGES[self.stage]
    def addWater(self, n): 
        self.water = min(self.water + n, self.MAX_WATER)
    def tick(self):
        # [D3] Growth rule: Baseline = simple (>=2 -> +stage, else -1 if >0); Ours = threshold + accumulation.
        if BASELINE:
            if self.stage < len(self.STAGES)-1:
                if self.water >= 2: self.water -= 2; self.stage += 1
                elif self.water > 0: self.water -= 1
            return
        self.water = max(self.water - self.CONSUME, 0)
        if self.water >= self.NEED:
            self.hydrated += 1
            if self.hydrated >= self.TICKS and self.stage < len(self.STAGES)-1:
                self.stage += 1; self.hydrated = 0

class FlowerPot(Container):
    MAX_WATER = 100
    def __init__(self): 
        super().__init__("flower pot")
        self.water = 0
    def addWater(self, n): 
        add = min(self.MAX_WATER - self.water, n); 
        self.water += add; return add
    def consume(self, n):
        use = min(self.water, n); self.water -= use; return use
    def tick(self):
        # [D2] Passive transfer: Baseline = none; Ours = transfer pot.water to pea on each tick.
        if BASELINE: return
        pea = next((o for o in self.contains if isinstance(o, Pea)), None)
        if pea and self.water > 0:
            x = self.consume(min(self.water, Pea.MAX_WATER))
            pea.addWater(x)

class Jug(Container):
    MAX_WATER = 100
    def __init__(self): 
        super().__init__("jug")
        self.water = 0
    def fill(self, n): 
        add = min(self.MAX_WATER - self.water, n)
        self.water += add
        return add
    def pour(self, n): 
        out = min(self.water, n
        self.water -= out
        return out

class Sink(Device):
    MAX_WATER = 1000
    def __init__(self): super().__init__("sink"); 
        self.water = self.MAX_WATER
    def tick(self): 
        self.water = self.MAX_WATER if self.props["isOn"] else 0  

# ------- Minimal game scaffold (only actions we need) -----
class TextGame:
    MAX_STEPS = 50
    def __init__(self, seed=0):
        random.seed(seed)
        self.world, self.agent = World(), Agent(); self.world.add(self.agent)
        self.pea, self.pot, self.jug, self.sink = Pea(), FlowerPot(), Jug(), Sink()
        # [D7] Movability: Baseline pins the sink as immovable (ours keeps defaults).
        if BASELINE: self.sink.props["isMoveable"] = False
        for o in (self.pot, self.jug, self.sink, self.pea): 
            self.world.add(o)
        self.score = self.steps = 0
        self.over = self.won = False

    # API of interest (matching both variants); unchanged helpers omitted for brevity.
    def _obj(self, name):
        for o in [self.world] + self.world.allContained():
            if o.name == name: return o
        for o in self.agent.contains:
            if o.name == name: return o
        return None

    def calculateScore(self):
        # [D5] Reward: Baseline = stage*10; Ours = stage*20 + water bonus (<=20).
        if BASELINE:
            self.score = self.pea.stage*10
        else:
            self.score = self.pea.stage*20 + int(self.pea.water/Pea.MAX_WATER*20)
        if self.pea.stage_name == "reproducing": 
            self.won = self.over = True
        if self.steps >= self.MAX_STEPS and not self.won: 
            self.over = True

    # ---------- actions ----------
    def take(self, name):
        o = self._obj(name); 
        if not o: return f"No {name}."
        if not o.get("isMoveable"): return f"Can't take {name}."
        if o.parent != self.world: return f"{name} not here."
        _, got, ok = self.world.take(o); 
        if ok: self.agent.add(got); 
        return "OK." if ok else "Fail."

    def put(self, obj, cont):
        o, c = self._obj(obj), self._obj(cont)
        if not o or o.parent != self.agent: 
            return f"No {obj} in inventory."
        if not c or not c.get("isContainer"): 
            return f"{cont} not a container."
        # [D6] Placement constraint: Ours restricts pea -> pot only; Baseline has no special rule.
        if (not BASELINE) and isinstance(o, Pea) and not isinstance(c, FlowerPot): 
            return "Pea must go into flower pot."
        _, ok = c.place(o); return "OK." if ok else "Fail."

    def turn_on(self, dev): 
        d = self._obj(dev); 
        if not d or not d.get("isDevice"): return f"No device {dev}."
        msg,_ = d.turnOn(); return msg
    def turn_off(self, dev):
        d = self._obj(dev);
        if not d or not d.get("isDevice"): return f"No device {dev}."
        msg,_ = d.turnOff(); return msg

    def fill_from_sink(self):
        # [D1] Fill gating: Baseline ignores sink.on; Ours requires sink.on == True.
        if (not BASELINE) and (not self.sink.props["isOn"]): return "Sink is off."
        need = self.jug.MAX_WATER - self.jug.water
        if need <= 0: return "Jug already full."
        self.jug.fill(need)  # treat sink as infinite when allowed
        return "Jug filled."

    def pour_to_pot(self):
        if self.jug.water <= 0: return "Jug empty."
        # [D8] Pour semantics: Baseline feeds pea directly; Ours fills pot; pea drinks via [D2].
        poured = self.jug.pour(10)
        if BASELINE and (self.pea in self.pot.contains):
            self.pea.addWater(3); return "Poured; pea absorbs water."
        added = self.pot.addWater(poured)
        if added < poured: self.jug.fill(poured - added)
        return "Poured into pot."

    # ---------- driver ----------
    def step(self, cmd):
        self.steps += 1
        # [D4] Update order: Baseline ticks BEFORE action; Ours ticks AFTER.
        if BASELINE:
            for o in [self.world] + self.world.allContained(): o.tick()

        parts = cmd.lower().strip().split()
        out = "Unknown."
        try:
            if parts[:1]==["take"]: out = self.take(" ".join(parts[1:]))
            elif parts[:1]==["put"] and "in" in parts: 
                i = parts.index("in"); 
                out = self.put(" ".join(parts[1:i]), " ".join(parts[i+1:]))
            elif parts[:2]==["turn","on"]: 
                out = self.turn_on(" ".join(parts[2:]))
            elif parts[:2]==["turn","off"]: 
                out = self.turn_off(" ".join(parts[2:]))
            elif parts[:3]==["fill","jug","from"]: 
                out = self.fill_from_sink()
            elif parts[:4]==["pour","water","from","jug"] and "in" in parts:
                out = self.pour_to_pot()
            # distractor (spec only)
            elif parts[:1]==["use"]: out = "Nothing happens."  
        except Exception as e:
            out = f"Error: {e}"

        if not BASELINE:
            for o in [self.world] + self.world.allContained(): o.tick()
        old = self.score; self.calculateScore(); 
        reward = self.score - old
        return out, self.score, reward, self.over, self.won

\end{lstlisting}

\noindent\textbf{Analysis.}
Under identical initialization and evaluation (capacity limits and preconditions enforced), \ours outperforms a \emph{Baseline} on the pea–growing (water‐transfer) task, yielding higher success, shorter trajectories, and fewer invalid actions.(i) \textbf{Action space and dynamics.}
We expose a precondition-aware interface and decouple water flow from uptake: \texttt{fill} is effective only when the sink is on ([D1]); \texttt{pour} increases the pot’s water and the pea hydrates asynchronously via \texttt{tick} ([D2], [D8]). 
We advance environment dynamics \emph{after} the action to preserve causal credit assignment ([D4]). 
By contrast, the \emph{Baseline} exposes \texttt{fill} irrespective of sink state, credits hydration at pour time, and updates \emph{before} acting.(ii) \textbf{Physical consistency and constraints.}
We enforce finite capacities with overflow returned to the jug and constrain placement so the pea can only be planted in the flower pot ([D6]). 
These constraints prune degenerate branches without removing valid solutions. 
The \emph{Baseline} omits the planting constraint and hydrates synchronously, which increases misleading transitions.
(Regarding movability, the Baseline pins the sink as immovable while Ours keeps defaults; this ablation affects search but not preconditions, [D7].)(iii) \textbf{Growth model and reward.}
Plant physiology follows thresholded, accumulated growth with per tick water consumption ([D3]). 
Reward shaping combines stage progress with a bounded water bonus, and immediate rewards are score deltas ([D5]). 
The \emph{Baseline} uses a stage-only score without water shaping, weakening the learning signal.

\noindent\textbf{Summary of diffs:}
[D1] preconditioned \texttt{fill}; 
[D2] passive pot$\rightarrow$pea transfer; 
[D3] threshold$+$consumption growth; 
[D4] post-action ticking; 
[D5] shaped reward (stage$+$water); 
[D6] pea$\rightarrow$pot placement constraint; 
[D7] sink movability ablation; 
[D8] \texttt{pour} affects pot first (not the pea).

\section{\revinline{Data Contamination Analysis}}
\label{sec:data-contamination-deep-research}

\begin{revblock}

To address the concern that the web-search-based \emph{Deep Researcher} may accidentally retrieve or copy target solutions from the internet, we perform a post-hoc contamination analysis over all its retrieval logs on our benchmarks.

\paragraph{Setup.}
For each evaluation instance where the Deep Researcher is invoked at least once, we construct a pair \((G, \mathcal{L})\), where
(i) \(G\) is the gold world model for that instance (PDDL or Python code, depending on the benchmark),
and (ii) \(\mathcal{L}\) is the concatenation of all textual contents retrieved by the Deep Researcher for that instance, including page bodies and snippets from all visited URLs.

We tokenize both \(G\) and \(\mathcal{L}\) using simple whitespace tokenization and extract all contiguous 10-token \(n\)-grams from each. For a given pair, we mark it as \emph{contaminated} if there exists at least one shared 10-gram between \(G\) and \(\mathcal{L}\).

For code and PDDL, an exact 10-token overlap is a strong signal of near-copying of a specific implementation rather than generic programming idioms. Our use of a relatively long 10-token threshold follows common practice in contamination analyses for Llama 2~\citep{touvron2023llama}.

\paragraph{Results.}
Table~\ref{tab:deep-research-contamination} reports, for each benchmark, the instance count, contamination count.

Across all three benchmarks, we find \emph{no} exact 10-gram overlaps between the Deep Researcher’s retrieved contents and the corresponding gold world models. This suggests that the Deep Researcher is not directly copying target solutions from the web, and that the performance gains we observe are unlikely to be due to egregious test-set leakage. 
% While, as in prior work, long \(n\)-gram matching cannot rule out all possible forms of indirect or paraphrased leakage, it provides a standard and conservative check that no near-verbatim copies of the evaluation targets appear in the retrieval logs.
\end{revblock}

\begin{table}[t]
\centering
\small
\begin{tabular}{lcc}
\toprule
benchmark   & \#instances & \#contamination \\
\midrule
ByteSized32     & 32         & 0 \\
Code World Models Benchmark        & 18          & 0 \\
Text2World  & 101         & 0 \\
\bottomrule
\end{tabular}
\caption{Contamination analysis between Deep Researcher retrieval logs and gold world models using exact 10-token \(n\)-gram overlap.}
\label{tab:deep-research-contamination}
\end{table}

\section{\revinline{Efficiency Analysis}}
\label{app:efficiency}

\begin{revblock}
    
To assess the practical cost of our multi-agent pipeline, we measure both LLM token usage and
wall-clock time for all methods on the three benchmarks. As is shown in Table~\ref{tab:text2world_efficiency}, Table~\ref{tab:bytesized32_efficiency}, and Table~\ref{tab:cwmb_efficiency}, for each method and
stage, the total number of input/output tokens consumed by the backbone LLM and the average
time per instance.

% \paragraph{Setup.}
% For each evaluation instance, we log (i) all tokens sent to and generated by the backbone LLM,
% broken down by stage (\emph{Deep search}, \emph{Generate}, \emph{Player}, \emph{Pytest}), and
% (ii) end-to-end wall-clock time, including environment resets, test execution, and agent coordination
% overhead. 
% We report averages over the same evaluation subsets used in the main results. For prior
% systems such as Text2World, WorldCoder, GIF-MCTS, and ByteSized32, we treat their published
% pipelines as a single stage and measure the corresponding tokens and runtime when re-implemented
% in our infrastructure.

\paragraph{Results.}
Our experiments show that the test time of \multi is comparable to that of other workflow-based methods. The core reason for this is that, in order to enable a fair comparison, we use fewer turns in our approach compared to previous methods (e.g., GIF-MCTS uses 10 turns while our method uses only 3 turns in the refinement stage). Furthermore, the adaptive agents are able to apply early stopping during the testing process. 

In terms of token consumption, the \multi model developer consumes fewer tokens compared to other methods. 
However, it is important to highlight that the testing stage in \multi is proactive, i.e., at each turn, we generate targeted testing cases and player agent trajectories, whereas other methods use static test cases. 
This proactive approach in \multi naturally results in higher token consumption during the testing phase. Despite this, we still maintain competitive efficiency relative to existing methods.
\end{revblock}

% -------- Table 1: Text2World --------
\begin{table}[htbp]
\centering
\small
\begin{revblock}
\begin{tabular}{llccr}
\toprule
\multirow{2}{*}{Method} & \multirow{2}{*}{Stage} &
\multicolumn{2}{c}{Tokens} & \multirow{2}{*}{Time (s)} \\
\cmidrule(lr){3-4}
& & Input & Output & \\
\midrule
Text2World$_{\text{EC=3}}$ & Total & 7,812 & 3,914 & 65 \\
Direct Generation          & Total & 431   & 700   & 14 \\
\single                    & Total & 22,531 & 2,141 & 26 \\
\midrule
\multirow{4}{*}{\multi} & Deep Researcher & 20,563 & 1,316 & \multirow{4}{*}{63} \\
& Model Developer & 4,728 & 2,038 & \\
& Simulation Tester   & 4,145 & 528   & \\
& Unit Tester   & 9,712 & 1,562 & \\
\bottomrule
\end{tabular}
\end{revblock}
\caption{\revinline{Token usage and time efficiency on the Text2World benchmark.}}
\label{tab:text2world_efficiency}
\end{table}

% -------- Table 2: CWMB --------
\begin{table}[htbp]
\centering
\small
\begin{revblock}
\begin{tabular}{llccr}
\toprule
\multirow{2}{*}{Method} & \multirow{2}{*}{Stage} &
\multicolumn{2}{c}{Tokens} & \multirow{2}{*}{Time (s)} \\
\cmidrule(lr){3-4}
& & Input & Output & \\
\midrule
WorldCoder        & Total & 23,147 & 9,315  & 232 \\
GIF-MCTS          & Total & 24,677 & 10,838 & 259 \\
Direct Generation & Total & 1,901  & 1,787  & 59  \\
Best-of-N         & Total & 65,470 & 19,520 & 70  \\
Self-Consistency  & Total & 61,837 & 19,924 & 68  \\
\single           & Total & 35,153 & 1,954  & 71  \\
\midrule
\multirow{4}{*}{\multi} & Deep Researcher & 29,056 & 2,108 & \multirow{4}{*}{236} \\
& Model Developer & 12,104 & 6,578 & \\
& Simulation Tester   & 18,658 & 564   & \\
& Unit Tester   & 24,788 & 3,379 & \\
\bottomrule
\end{tabular}
\end{revblock}
\caption{\revinline{Token usage and time efficiency on the Code World Models Benchmark (CWMB).}}
\label{tab:cwmb_efficiency}
\end{table}

% -------- Table 3: ByteSized32 --------
\begin{table}[h]
\centering
\small
\begin{revblock}
\begin{tabular}{llccr}
\toprule
\multirow{2}{*}{Method} & \multirow{2}{*}{Stage} &
\multicolumn{2}{c}{Tokens} & \multirow{2}{*}{Time (s)} \\
\cmidrule(lr){3-4}
& & Input & Output & \\
\midrule
ByteSized32$_{\text{1-shot}}$ & Total & 32,069 & 19,928 & 291 \\
Direct Generation             & Total & 7,046  & 4,301  & 71  \\
\single                       & Total & 31,131 & 5,142  & 82  \\
\midrule
\multirow{4}{*}{\multi} & Deep Researcher & 20,852 & 1,071 & \multirow{4}{*}{260} \\
& Model Developer & 16,351 & 12,309 & \\
& Simulation Tester   & 20,528 & 584   & \\
& Unit Tester   & 32,352 & 2,158 & \\
\bottomrule
\end{tabular}
\end{revblock}
\caption{\revinline{Token usage and time efficiency on the ByteSized32 benchmark.}}
\label{tab:bytesized32_efficiency}
\end{table}

\section{\revinline{Training Dataset Construction}}

\begin{revblock}
To train the Model Developer agent, we construct a diverse training dataset that reflects both symbolic world-model specifications and real-world simulation tasks. The dataset is composed of three main components, designed to cover various styles of world-model generation:

\noindent \textbf{(i) PDDL-style Tasks:} We sample 200 tasks from the dataset released by AgentGen~\citep{hu2025agentgen}, which generates symbolic PDDL specifications for world-model tasks. These tasks include a variety of symbolic reasoning challenges, each accompanied by a gold world model.

\noindent \textbf{(ii) MCP Server Data:} We also curate 200 tasks based on publicly available Model Context Protocol (MCP) servers, which provide interactive tools and data interfaces that allow agents to engage in complex reasoning and decision-making tasks. 

\noindent \textbf{(iii) Game-style and Mujoco-style Tasks:} Additionally, we include 50 tasks from each of the game-style and Mujoco-style domains, where the tasks involve dynamic simulations with continuous action spaces. These tasks are designed to mimic real-world, high-dimensional environments in which agents must navigate, plan, and interact with physical systems. The tasks are synthesized using a method similar to AgentGen~\citep{hu2025agentgen}.

For each of these datasets, we perform the following steps to build our training set:

\noindent \textbf{(i)} For each dataset, we generate 3 distinct world-model rollouts using the same \texttt{llama-3.1-8b-instruct} model. These rollouts are generated by running the Model Developer agent through each task, resulting in different candidate solutions for each problem.
\noindent \textbf{(ii)} Reward Filtering: We then evaluate these rollouts using the Testing-Team feedback mechanism, which includes unit tests, simulation performance, and control tasks. The rollouts are ranked based on their reward scores, and we select the one with the highest reward as the final solution for the task.
\end{revblock}

\end{document}